\documentclass{article}
\usepackage{graphicx} 
\usepackage{xcolor}
\usepackage{soul}
\usepackage{comment}
\usepackage{authblk}
\usepackage[letterpaper,top=2cm,bottom=2cm,left=2cm,right=2cm,marginparwidth=1.75cm]{geometry}
\usepackage{amsthm}
\usepackage{amsmath}
\usepackage{amssymb}
\usepackage{algorithmic}
\usepackage{algorithm}
\title{\textbf{Hypersonic Flow Control: Generalized Deep Reinforcement Learning for Hypersonic Intake Unstart Control under Uncertainty}}
\author[]{Trishit Mondal}
\author[]{Ameya D. Jagtap\thanks{Corresponding author: Ameya D. Jagtap (ajagtap@wpi.edu, ameyadjagtap@gmail.com)}}

\affil[]{\textit{\small{Aerospace Engineering Department, Worcester Polytechnic Institute, Worcester, MA 01609, USA.}}}
\date{}

\begin{document}

\maketitle

\begin{abstract}
The hypersonic unstart phenomenon poses a major challenge to reliable air-breathing propulsion at Mach 5 and above, where strong shock--boundary-layer interactions and rapid pressure fluctuations can destabilize inlet operation. Here, we demonstrate a deep reinforcement learning (DRL)- based active flow control strategy to control unstart in a canonical two-dimensional hypersonic inlet at Mach 5 and Reynolds number $5\times 10^6$. The in-house CFD solver enables high-fidelity simulations with adaptive mesh refinement, resolving key flow features, including shock motion, boundary-layer dynamics, and flow separation, that are essential for learning physically consistent control policies suitable for real-time deployment. The DRL controller robustly stabilizes the inlet over a wide range of back pressures representative of varying combustion chamber conditions. It further generalizes to previously unseen scenarios, including different back-pressure levels, Reynolds numbers, and sensor configurations, while operating with noisy measurements, thereby demonstrating strong zero-shot generalization. Control remains robust in the presence of noisy sensor measurements, and a minimal, optimally selected sensor set achieves comparable performance, enabling practical implementation. These results establish a data-driven approach for real-time hypersonic flow control under realistic operational uncertainties.
\end{abstract}

\vspace{0.2cm}

 \begin{small}Keywords: \textit{Hypersonic Flow Control}; \textit{Deep Reinforcement Learning}; \textit{Zero-Shot Generalization}; \textit{Hypersonic Unstart Control}.
\end{small}

\section{Introduction}
Hypersonic air-breathing propulsion systems are attracting increasing attention as nations and private entities compete in the global race for faster, more efficient aerospace technologies. Their potential to enable rapid global transportation, responsive access to space, and advanced high-speed flight capabilities has made them a focal point of research and development, with implications for both commercial and strategic applications. In such systems, including scramjets, the inlet plays a central role in establishing the high-pressure conditions required for efficient combustion by compressing the incoming hypersonic airstream through a sequence of carefully arranged oblique and terminal shocks. Under ideal operation, these shocks remain stably anchored within the intake, ensuring a continuous and controlled delivery of compressed air to the combustor. However, at certain flight conditions or during transient engine-airframe interactions, this delicate balance can break down. A rapid rise in internal pressure, arising from combustor back pressure, boundary-layer growth, shock–boundary-layer interaction, or inadequate starting Mach number, can destabilize the internal shock system of a hypersonic inlet and trigger the \textit{unstart phenomenon}, in which the shock system convects upstream toward the intake entrance. When the imposed back pressure exceeds the inlet’s maximum mass-flow capacity, the flow at the throat becomes choked and can no longer adjust to further pressure increase. This condition defines the \textit{Kantrowitz limit} \cite{Kantrowitz1945}, beyond which a normal shock is expelled from the inlet, producing an abrupt transition from started to unstarted operation. In this regime, unstart is not driven by shock oscillations or inlet buzz but instead represents a quasi-steady, unavoidable limit governed by one-dimensional mass, momentum, and energy conservation. The resulting inlet spillage, loss of captured mass flow, and large pressure excursions severely degrade thrust and impose substantial aerodynamic and structural loads, thereby setting a fundamental upper bound on the tolerable combustor pressure rise in hypersonic air-breathing propulsion systems.

Using active flow control (AFC) to mitigate the hypersonic unstart phenomenon represents a modern and adaptive approach to stabilizing shock systems in high-Mach inlets, where traditional passive methods or fixed-geometry solutions often fail under transient or off-design conditions. In hypersonic intakes, unstart is typically triggered by adverse shock–boundary-layer interactions, separation, and internal over-pressurization; AFC techniques, such as synthetic jets, pulsed blowing, plasma actuators, or micro-vortex generators, enable direct, localized manipulation of these critical flow structures. By injecting momentum, altering near-wall vorticity, or modifying effective boundary-layer thickness, AFC provides a way to dynamically reshape the internal shock pattern and maintain anchoring even as the external environment or combustor back-pressure fluctuates. When integrated with deep reinforcement learning (DRL), AFC becomes a powerful closed-loop system: pressure sensors, heat-flux gauges, or optical diagnostics supply real-time indicators of shock movement and incipient separation, while the DRL agent learns how to command the actuators, modulating injection frequency, amplitude, phase, or duty cycle, to preemptively suppress the flow instabilities that precede unstart. Through training in high-fidelity CFD or reduced-order surrogate models, the controller discovers complex, non-intuitive actuation strategies that classical methods cannot easily derive, such as phase-synchronized pulsed blowing timed to weaken separation bubbles or targeted momentum bursts to push shocks downstream.

\subsection{Related Work}

The concept of synthetic jets, also known as zero-net-mass-flux jets, was firmly established in the early 2000s as a viable actuation mechanism for AFC. A synthetic jet actuator typically consists of a cavity with an oscillating diaphragm that alternately ingests and expels fluid through an orifice. This cycle produces a train of vortices without net mass addition to the flow, relying instead on momentum transfer to manipulate the surrounding fluid. Foundational work by Holman et al. \cite{holman2005formation} clarified the physical mechanisms governing the formation of these coherent vortical structures. Subsequent research has extensively characterized how geometric parameters such as cavity shape, orifice width, and jet orientation, along with actuation waveform properties, influence vortex generation and their subsequent interaction with external boundary layers. These insights established synthetic jets as effective tools for delaying flow separation on airfoils and bluff bodies when appropriately tuned.

The integration of DRL into AFC has evolved significantly over the last two decades, transitioning from fixed harmonic forcing to self-learning control systems. Early reviews by Choi et al. \cite{choi2008control} emphasized wake manipulation strategies for bluff bodies, while Pivot et al. \cite{pivot2017continuous} demonstrated the potential of continuous RL for closed-loop drag reduction using sparse sensor data. A key study in this field is conducted by Rabault et al. \cite{rabault2019artificial}, who successfully integrated modern DRL with traditional AFC, replacing open-loop fixed control with autonomous closed-loop policies, using a second-order accurate Finite Element Method framework to resolve the laminar flow field. To refine these control strategies, Qin et al. \cite{qin2021application} introduced data-driven reward functions based on dynamic mode decomposition, while Zheng et al. \cite{zheng2021active} applied off-policy-based Soft Actor-Critic algorithm \cite{haarnoja2018soft} over active learning for vortex suppression. Experimental success followed, where Shimomura et al. \cite{shimomura2020closed} applied Deep Q-Networks to plasma actuators on a NACA0015 airfoil, and Fan et al. \cite{fan2020reinforcement} validated RL-based cylinder rotation for drag reduction. Researchers also expanded into complex multi-actuator environments. Dobakhti et al. \cite{dobakhti2023active} combined cylinder rotation with synthetic jets to achieve up to 50\% drag reduction. Concurrently, Jia et al. \cite{jia2024effect} highlighted the importance of physical design in these setups, demonstrating the need to co-optimize actuator geometry, such as jet width and placement, alongside the learned policy. Paris et al. \cite{paris2023reinforcement} introduced frameworks for adaptive actuator selection. Suarez et al. \cite{suarez2025flow} advanced 3-D control by implementing a multi-agent framework for spanwise-distributed jets to achieve coordinated strategies with reduced mass flux. Efforts have focused on training efficiency and transferability. He et al. \cite{he2023policy} and Yan et al. \cite{yan2025deep} demonstrated the transfer of control policies from 2-D to 3-D flows, while Font et al. \cite{font2025deep} showed that policies could generalize from coarse to fine CFD grids, which drastically reduced computational costs. In recent years, DRL has demonstrated remarkable robustness in fully turbulent flows at high Reynolds numbers.
Chen et al. \cite{chen2025active} achieved significant drag reduction in turbulent bluff-body flows at $Re=2.74\times10^5$. Similarly, Zhou et al. \cite{zhou2025reinforcement} and Guastoni et al. \cite{guastoni2023deep} successfully applied DRL to turbulent channel flows to suppress near-wall streaks. Further improving on the efficiency, Wang et al. \cite{wang2024dynamic} enabled effective control from sparse wall-pressure data using dynamic feature extraction, and Montala et al. \cite{montala2025deep} utilized PPO agents to autonomously modulate leading-edge vortices for lift enhancement on wings.

While the vast majority of DRL-based AFC research has focused on incompressible regimes, recent studies have increasingly demonstrated its effectiveness in compressible flow environments, where shock waves, thermodynamics, and fluid-structure interactions (FSI) present unique control challenges. Effective control in these high-speed regimes requires robust sensing capabilities, often limited by computationally expensive environment. Poulinakis et al. \cite{poulinakis2023deep} addressed this by applying Long Short-Term Memory networks to supersonic shock-boundary layer interactions (SBLI). They demonstrated that deep learning could accurately reconstruct near-wall pressure fluctuations from sparse, under-sampled datasets, enabling the prediction of complex aerodynamic loads without computationally expensive full-domain simulations. In the transonic regime, research has focused heavily on suppressing aeroelastic instabilities such as buffet and flutter. Ren et al. \cite{ren2024adaptive} applied DRL to suppress both transonic buffet (aerodynamic instability) and buffeting (structural vibration). Unlike traditional methods requiring accurate dynamic models, their model-free agent autonomously mitigated unsteady loads in coupled FSI systems. Similarly, Gong et al. \cite{gong2024active} utilized PPO agents to control transonic airfoil flutter using synthetic jets. Their second-order spatially accurate framework treated the FSI problem as a learning environment, successfully reducing pitching and plunging oscillations across a wide range of Mach numbers and extending the flutter boundary beyond the training conditions. There are DRL approaches, which stand in contrast to concurrent model-based strategies. For example, Deng et al. \cite{deng2023closed} proposed a classical closed-loop control algorithm using active shock control bumps to reduce buffet on supercritical airfoils, relying on lift-coefficient feedback and optimized gain/delay parameters. In parallel, Gao et al. \cite{gao2025transonic} developed a time-variant reduced-order model for adaptive control. Recently, Mondal et al. \cite{mondal2025shocks} addressed the control of transonic shock-boundary layer interactions on an RAE2822 airfoil using a fifth-order spectral Discontinuous Galerkin framework. By using a fifth-order spectral DG scheme to resolve shock oscillations and Kutta waves, they showed that DRL can reduce drag both with and without lift penalties.

As flight speeds accelerate into the hypersonic regime, the primary control challenges shift from optimizing aerodynamic efficiency to ensuring propulsion stability. Before active control can be applied, rapid detection of instability is required. Min et al. \cite{min2024machine} addressed the challenge of inlet unstart in combined-cycle engines across a wide speed range. They developed a hybrid machine learning framework using Support Vector Machines to define the unstart boundary and 1-D Convolutional Neural Networks to predict backpressure ratios. By utilizing only 2-4 surface pressure sensors, their model achieved prediction times under 2 ms with high accuracy. Downstream of the inlet, the \textit{isolator} must contain a series of shock waves known as a \textit{shock train}, that decelerate the flow for combustion. Sethuraman et al. \cite{sethuraman2021control} investigated the use of boundary layer suction to manage self-excited shock trains. Their numerical study demonstrated that suction not only reduced the physical length of the shock train but also dampened its oscillation amplitude. Moving to active electric control, Hahn et al. \cite{hahn2025control} utilized quasi-DC filamentary electrical discharges to control Mach 2 shock trains. Experiments showed that the discharge effectively anchored the leading edge of the shock wave to the electrodes, which stabilizes the pressure gradient along the isolator wall. In a model-based RL setup, Tao et al. \cite{tao2025control} applied DRL to control separation on a hypersonic compression ramp using microjets. To overcome the prohibitive cost of hypersonic CFD, they used a Long Short-Term Memory network as a surrogate environment. This approach reduced training time, allowing the agent to learn policies quickly. But for generating CFD-based data they used a second-order finite volume scheme. Finally, control extends into the combustion chamber itself. Liang et al. \cite{liang2025research} developed a robust control strategy for scramjet fuel pulse injection. By combining Active Disturbance Rejection Control (ADRC) with a Twin Delayed Deep Deterministic Policy Gradient (TD3) \cite{fujimoto2018addressing} agent, they created a real-time adjustment method (RL-ADRC). This system dynamically optimized injection parameters to satisfy thrust demands and prevent unstart.

\subsection{Contribution of this work}
In this work, the primary objective is to employ a DRL-based control framework to actively suppress unstart in an air-breathing hypersonic intake operating at Mach 5. The control strategy is trained to remain robust to variations in back pressure and Reynolds number, both with and without measurement noise, thereby improving intake stability and overall operational reliability. The main contributions of this work are summarized as follows:

\begin{enumerate}
    \item A high-fidelity CFD framework is employed, featuring a fifth-order spectral Discontinuous Galerkin (DG) discretization in space and a strong-stability-preserving Runge--Kutta (SSP-RK) scheme of order (5,4) for temporal integration. The in-house solver further incorporates conservative adaptive mesh refinement (AMR), enabling enhanced resolution of shocks and other fine-scale flow features critical to unstart dynamics.
    
    \item A DRL-based controller is trained across multiple throttling ratios (TRs), corresponding to varying back-pressure conditions, thereby ensuring robustness over a wide range of operating regimes.
    
    \item Zero-shot generalization capability is demonstrated by deploying a controller trained exclusively at TR40 to previously unseen operating conditions, namely TR30 and TR50. This result highlights the ability of the learned policy to generalize beyond its training envelope, indicating strong potential for real-time implementation without retraining.
    
    \item The performance and robustness of the proposed controller under zero-shot generalization are also evaluated using noisy sensor measurements, demonstrating resilience to measurement uncertainty and practical sensing limitations.
    
    \item A systematic methodology is presented to identify the optimal sensor placement and to assess controller performance under zero-shot generalization. By relying on a minimal sensor set, the approach improves practical implementability while maintaining effective unstart control.

\item We further demonstrate the zero-shot generalization of the controller trained on the TR40 case at a Reynolds number of $5 \times 10^6$ to two previously unseen Reynolds numbers, $10 \times 10^6$ and $15 \times 10^6$, in the TR50 configuration with 10\% noise, highlighting its capacity to maintain stable and consistent control performance without any additional training.

\end{enumerate}
The remainder of the paper is organized as follows. Section 2 describes the governing compressible Navier–Stokes equations and the problem formulation. Section 3 gives details of high-fidelity CFD solver along with $hp$-convergence analysis. Section 4 introduces the DRL setup, including the reward design, learning objective, and off-policy learning algorithms. Section 5 presents the results, beginning with training performance at different throttle ratios, followed by an evaluation of zero-shot generalization. The robustness of the controller to noisy sensor measurements and its effectiveness using an optimally selected minimal sensor set are also assessed. Section 6 concludes with a summary of the main findings.

\section{Governing Equations and Problem Statement}
The flow within a hypersonic inlet is governed by the two-dimensional compressible Navier--Stokes equations, and all simulations are performed using the \textit{full-order model}.
In conservative form, these equations are written as:
\begin{equation}\label{NSE}
    \frac{\partial \mathbf{U}}{\partial t} + \frac{\partial \mathbf{F}_{\text{inv}}}{\partial x} + \frac{\partial \mathbf{G}_{\text{inv}}}{\partial y} = \frac{\partial \mathbf{F}_v}{\partial x} + \frac{\partial \mathbf{G}_v}{\partial y},  ~~ t \in \mathbb{R}^+ ~\text{and}~ (x,y) \in \Omega \subset \mathbb{R}^2,
\end{equation}
with appropriate initial and boundary conditions.
Here, $\mathbf{U}$ denotes the vector of conservative variables, while $\mathbf{F}_{\mathrm{inv}}$ and $\mathbf{G}_{\mathrm{inv}}$ represent the inviscid convective flux vectors in the $x$- and $y$-directions, respectively. These quantities are defined as follows:
\begin{equation}
    \mathbf{U} = \begin{bmatrix}
        \rho \\
        \rho u \\
        \rho v \\
        \rho E
    \end{bmatrix}, \quad
    \mathbf{F}_{\text{inv}} = \begin{bmatrix}
        \rho u \\
        \rho u^2 + p \\
        \rho uv \\
        u(\rho E + p)
    \end{bmatrix}, \quad \text{and} \ \ 
    \mathbf{G}_{\text{inv}} = \begin{bmatrix}
        \rho v \\
        \rho uv \\
        \rho v^2 + p \\
        v(\rho E + p)
    \end{bmatrix},
\end{equation}
where $\rho$ denotes the fluid density, $\mathbf{u} = [u,\, v]^{T}$ is the velocity vector, and $p$ represents the static pressure. The total energy per unit mass, $E$, is related to the pressure and kinetic energy as:
\begin{equation}
    E = \frac{p}{\rho(\gamma - 1)} + \frac{1}{2} \|\mathbf{u}\|_2^2.
\end{equation}
The diffusive effects due to viscosity and thermal conduction are captured by the viscous flux vectors, $\mathbf{F}_v$ and $\mathbf{G}_v$, which are given by
\begin{equation}
    \mathbf{F}_v = \begin{bmatrix}
        0 \\
        \tau_{xx} \\
        \tau_{xy} \\
        u\tau_{xx} + v\tau_{xy} - q_x
    \end{bmatrix}, \quad
    \mathbf{G}_v = \begin{bmatrix}
        0 \\
        \tau_{xy} \\
        \tau_{yy} \\
        u\tau_{xy} + v\tau_{yy} - q_y
    \end{bmatrix}.
\end{equation}
The system is closed using the ideal gas law, $p = \rho R T$, where $R$ is the specific gas constant and $T$ denotes the temperature. The ratio of specific heats is defined as $\gamma = c_p / c_v$. The components of the viscous stress tensor, $\boldsymbol{\tau}$, are given by:
\begin{align*}
\tau_{xx} &= \mu\left(2\frac{\partial u}{\partial x} - \frac{2}{3}\nabla \cdot \mathbf{u}\right), \quad
\tau_{yy} = \mu\left(2\frac{\partial v}{\partial y} - \frac{2}{3}\nabla \cdot \mathbf{u}\right) \\
\tau_{xy} &= \mu\left(\frac{\partial u}{\partial y} + \frac{\partial v}{\partial x}\right).
\end{align*}
where $\mu$ is the dynamic viscosity. The heat flux with thermal conductivity k is described as:
\begin{equation}
    q_x = -k \frac{\partial T}{\partial x}, \quad q_y = -k \frac{\partial T}{\partial y}.
\end{equation}

\subsection{Computational Domain and Unstart Dynamics}
\begin{figure}[htpb]
\centering
\includegraphics[scale=1.2, clip=true]{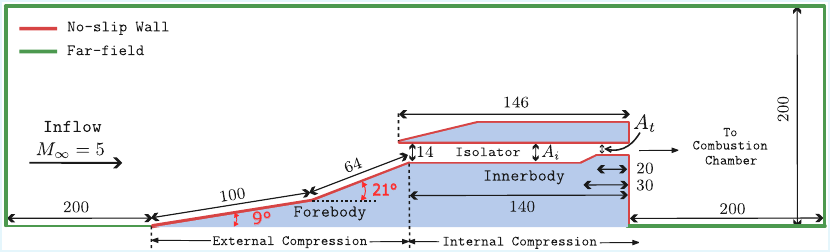}
\caption{Computational domain $(\Omega)$ with inflow Mach 5 and boundary conditions. All dimensions are in 'mm'.}
\label{fig:comp_dom}
\end{figure}

Figure~\ref{fig:comp_dom} illustrates the computational domain of the two-dimensional hypersonic intake. All simulations are performed at a freestream Mach number of $M_{\infty}=5.0$ and a Reynolds number of $Re=5.0\times10^{6}$ per unit length. The freestream conditions correspond to a static pressure of $p_{\infty}=900.0$~Pa and a static temperature of $T_{\infty}=101.0$~K. The freestream density is obtained from the ideal gas law,
$\rho_{\infty}=p_{\infty}/(R T_{\infty})\approx0.031$~kg~m$^{-3}$,
using a specific gas constant of $R=287.87$~J~kg$^{-1}$~K$^{-1}$. The Prandtl number is fixed at $Pr=0.72$. External compression is achieved using two ramps inclined at $9^{\circ}$ and $21^{\circ}$ to the horizontal. Upon entering the isolator, the flow undergoes internal compression through a series of shock trains. Different back-pressure conditions are imposed via \textit{Throttling}.

Throttling in hypersonic intakes can induce an unstart by increasing the back pressure inside the inlet. As the engine reduces mass flow to throttle down, the pressure downstream of the inlet rises. When this back pressure exceeds the inlet’s ability to compress the incoming supersonic flow, the shock system inside the intake is forced upstream, leading to boundary-layer separation and expulsion of the shock. 
The Throttling Ratio (TR) quantifies the magnitude of the downstream blockage applied to the hypersonic intake to generate back-pressure. It is defined based on the relationship between the constant cross-sectional area of the isolator, $A_i$, and the effective exit area, $A_t$ (as shown in Figure \ref{fig:comp_dom}), as follows:
\begin{equation}
    \text{TR} = \left( 1 - \frac{A_t}{A_i} \right) \times 100\%.
\end{equation}
A higher TR value corresponds to a smaller exit area $A_t$ relative to $A_i$, which induces higher back pressure within the isolator and pushes the shock train upstream, eventually triggering an unstart.

\section{High-fidelity CFD Solver with \textit{hp}-Convergence Analysis}
\begin{figure}[htpb]
\centering
\includegraphics[scale=0.8, clip=true]{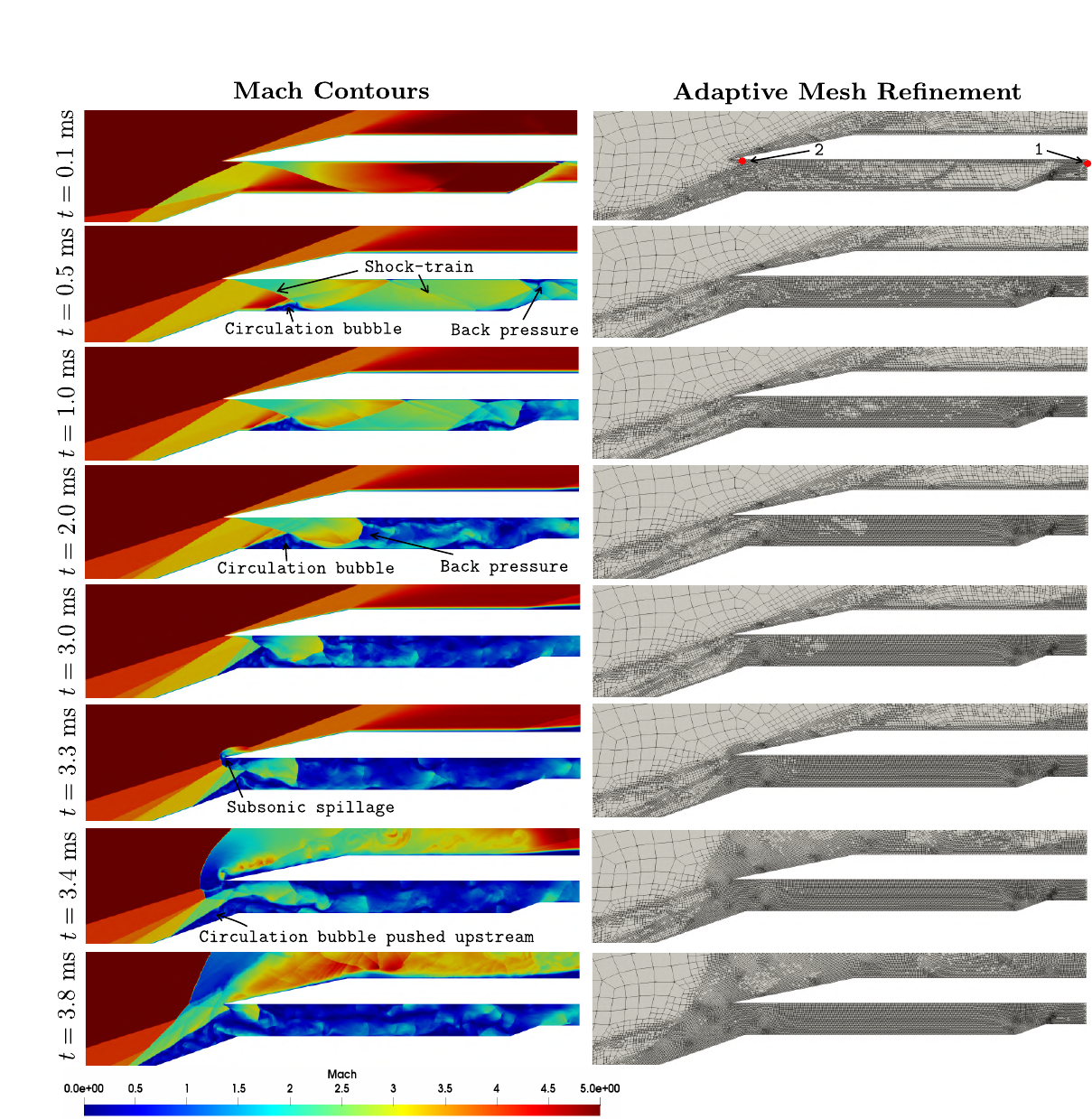}
\caption{TR34: Mach number contours at successive time steps illustrating the gradual buildup of back pressure, followed by subsonic spillage and the onset of inlet unstart at Mach~5 and $R_e = 5 \times 10^6$. From top to bottom, the rows correspond to times $t = 0.1, 0.5, 1.0, 2.0, 3.0, 3.3, 3.4,$ and $3.8~\mathrm{ms}$. The first column shows the Mach number contours, while the second column shows the corresponding AMR distribution. Points~1 and~2 (top-right) shows the location where temporal evolution of pressure is measured.
}
\label{fig:mach_contour}
\end{figure}
\begin{figure}
\centering
\includegraphics[scale=0.34, clip=true]{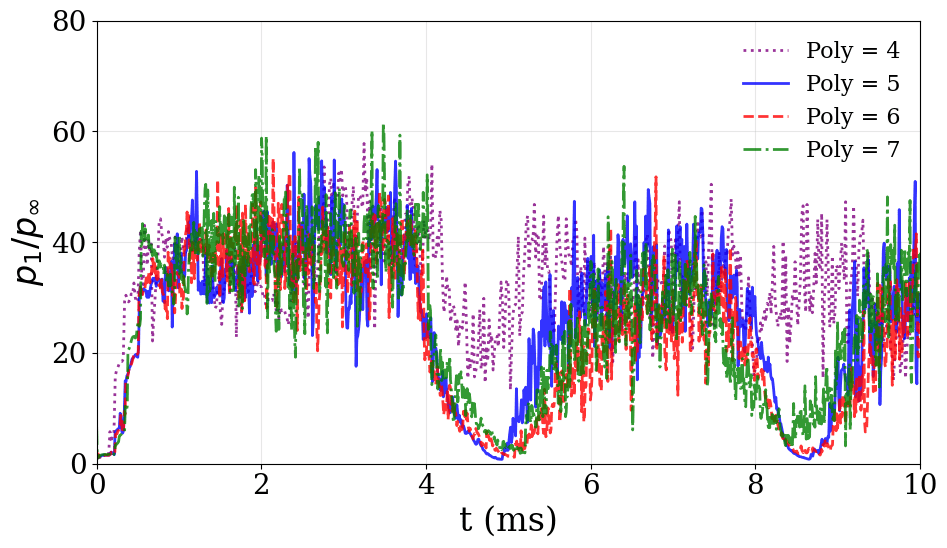}
\includegraphics[scale=0.34, clip=true]{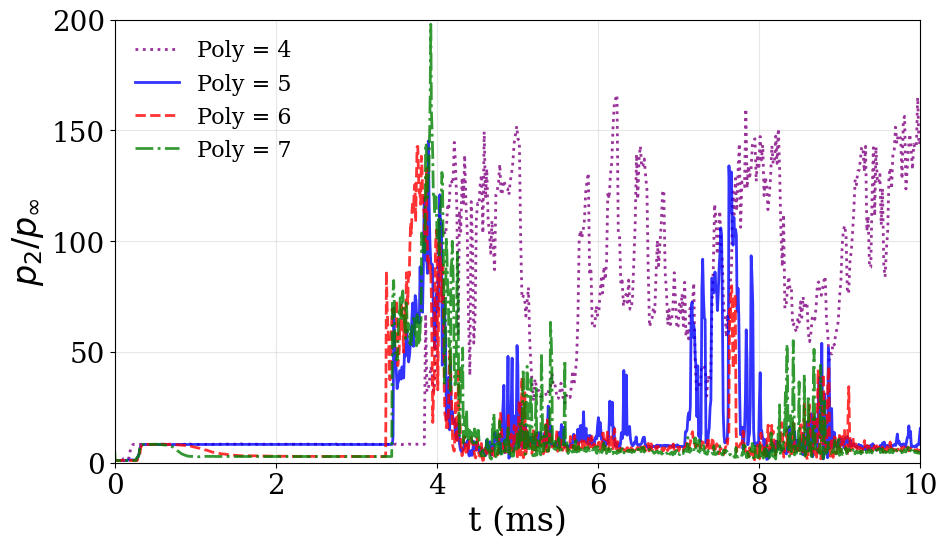}
\caption{TR34: hp-convergence study of the pressure ratios $p_1/p_{\infty}$ and $p_2/p_{\infty}$ for polynomial orders $5$, $6$, and $7$. The results demonstrate convergence of the solution with increasing polynomial order as well as through mesh refinement using AMR. The approximate number of elements for 5th, 6th, and 7th polynomial orders is 40k, 50k, and 60k, respectively.}
\label{fig:Fig1}
\end{figure}
\begin{figure}
\centering
\includegraphics[scale=0.34, clip=true]{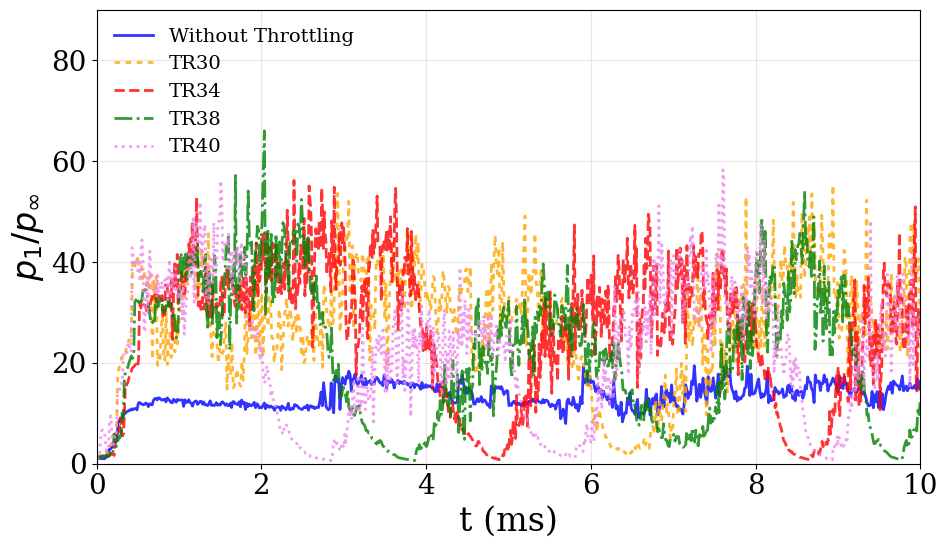}
\includegraphics[scale=0.34, clip=true]{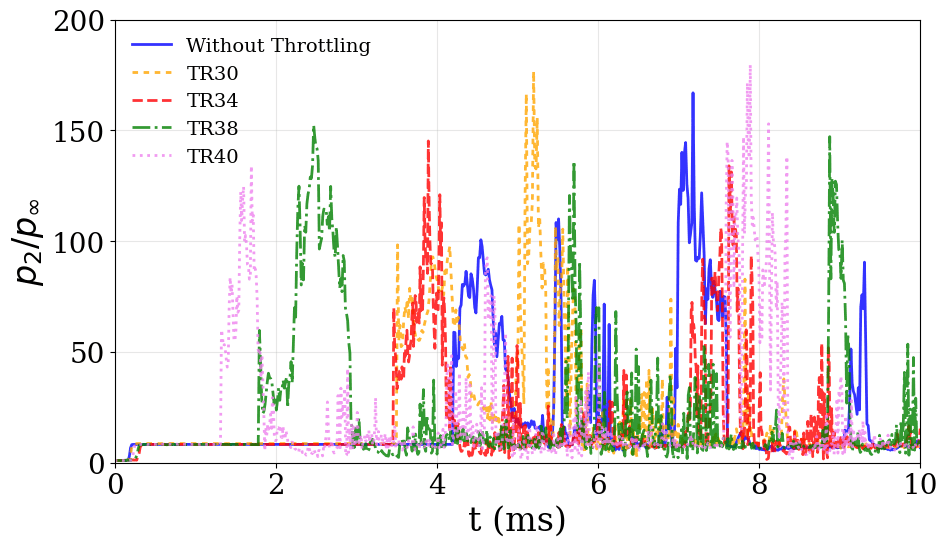}
\caption{Effect of throttling on the pressure ratios $p_1/p_{\infty}$ and $p_2/p_{\infty}$ for TR30, TR34, TR38, and TR40, including the unthrottled condition. 
The variation of $p_2/p_{\infty}$ clearly indicates the onset of the upstart phenomenon, characterized by a sudden rise in pressure. 
Furthermore, the onset of upstart occurs earlier in time as the TR increases, which is attributed to the corresponding increase in back pressure.}
\label{fig:FigTHRO}
\end{figure}
Most existing studies on DRL-based active flow control rely on low-order numerical schemes, typically second/third-order accurate, for the discretization of spatial and temporal derivatives. However, hypersonic unstart is a highly critical phenomenon that involves a wide range of complex, multi-scale flow features, including strong shock--boundary-layer interactions, flow separation, and large unsteady pressure gradients. Accurately resolving these phenomena requires the use of high-fidelity CFD solvers. In the present work, a high-order spectral DG method is employed for spatial discretization (see the weak formulation in Appendix \ref{appA}), while time integration is performed using high-order strong-stability-preserving Runge--Kutta (SSP-RK) schemes (see Appendix \ref{appB}, for more details). To resolve flow features across multiple scales, including shocks, boundary layers, and wakes, we employ adaptive mesh refinement (AMR) facilitated by the \texttt{P4estMesh} library \cite{burstedde2011p4est}.  
The computational domain is organized as a parallel forest-of-quadtrees, allowing each quadrilateral element to be recursively subdivided across several refinement levels.  
Refinement is driven by a L\"ohner indicator \cite{Lohner1987} based on density gradients,  
\[
\eta_i = \frac{|\nabla^2 \rho_i|}{|\nabla \rho_i| + \epsilon},
\]  
which effectively identifies shocks and regions with steep gradients, ensuring that critical flow features are captured with high fidelity. We employed a fifth-order DG scheme coupled with the SSPRK(5,4) time integrator.  
Figure~\ref{fig:mach_contour} presents Mach number contours at a throttle ratio of 34\% (TR34), illustrating the temporal evolution of the flow within the hypersonic intake.
 As time progresses, a buildup of back pressure is observed, which triggers subsonic spillage at the inlet lip. This flow disturbance subsequently leads to the initial onset of unstart, characterized by upstream movement of the shock system and loss of stable supersonic flow through the intake at Mach 5. The contours effectively illustrate how throttling-induced back pressure can destabilize the inlet and initiate unstart phenomena.

To further assess the adequacy of numerical resolution, an \textit{hp-convergence} study is conducted by varying both the polynomial order and the number of mesh elements. 
The results for the TR34 case are presented in Fig. \ref{fig:Fig1}, showing the time evolution of normalized pressure $p_1/p_{\infty}$ and $p_2/p_{\infty}$ (at points~1 and~2, indicated in the top-right panel of Fig.~\ref{fig:mach_contour}). The comparison highlights the insufficiency of fourth-order polynomial (Poly = 4) discretizations, the solution shows excessive non-physical oscillations and significant deviation from the higher-order solutions, particularly evident in the $p_2$ probe data where it fails to capture the correct unstart timing. In contrast, grid convergence is observed for polynomial orders of five and higher ($Poly =5, 6, 7$). These solutions shows strong agreement in capturing the transient pressure dynamics, despite the variation in mesh density (approximate element counts of 40k, 50k, and 60k for 5th, 6th, and 7th orders, respectively). This confirms that high-order polynomial is strictly necessary for capturing the essential dynamics of hypersonic unstart. Therefore, the use of high-fidelity CFD solvers is essential in DRL-based active flow control frameworks, as inadequate numerical accuracy may compromise the fidelity of the resolved flow physics and, in turn, lead to suboptimal or misleading control strategies. Figure~\ref{fig:FigTHRO} shows the temporal evolution of pressure for the unthrottled case and for TR of 30\%, 34\%, 38\%, and 40\%. At point~2, the onset of unstart dynamics is clearly identified by a sharp rise in pressure associated with the accumulation of back pressure. As the throttle ratio increases, the elevated back pressure drives the shock train rapidly upstream within the isolator, leading to a sudden increase in pressure at point~2.

\section{Methodology: The DRL Set-up}
Synthetic jets offer efficient flow control, but their effectiveness is fundamentally limited at high Reynolds or Mach numbers. The momentum they can inject is capped by actuator mechanics, and as flow speeds increase, the momentum required to influence the base flow quickly exceeds this limit. At high Reynolds or compressible conditions, the vortical structures that drive synthetic-jet performance break down rapidly due to turbulence and, at finite Mach numbers, shock interactions, reducing their entrainment capability. Furthermore, their efficiency depends on careful frequency tuning, which becomes impractical as freestream velocity rises. In contrast, continuous or pulsed jets can supply larger momentum through mass injection, though at higher energetic cost. In the hypersonic flow regime, microjets can serve as powerful tools for momentum injection, shock-boundary layer interaction, or localized heating. Compared to classical synthetic jets, they are significantly more aggressive and can potentially damage surfaces if applied near delicate airfoils. DRL offers a promising approach to optimize their usage, allowing control over parameters such as jet angle and their magnitude to effectively manipulate flow dynamics at such a high-speed. 
\begin{figure}[htpb]
\centering
\includegraphics[scale=0.9, clip=true]{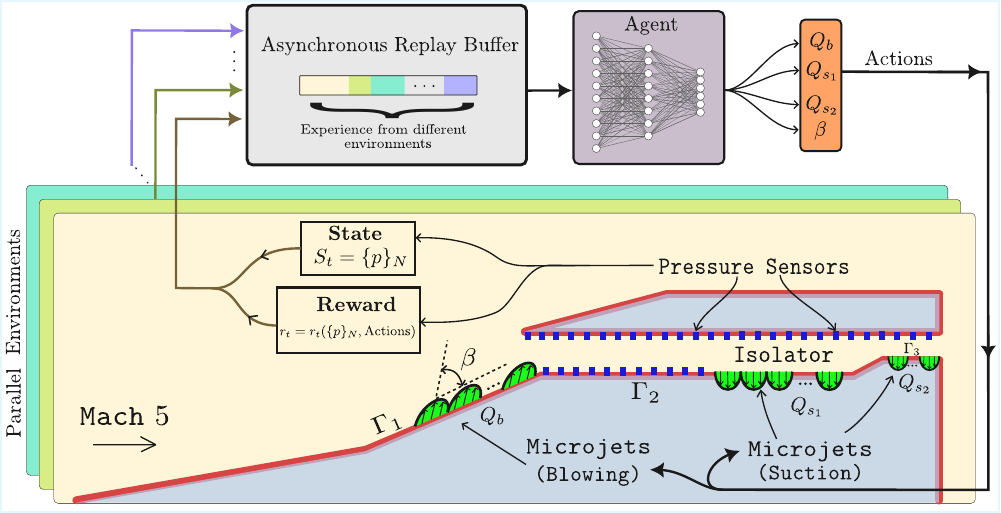}
\caption{Schematic of microjet placement along the wall: blowing microjets with learnable jet angle ($\beta$) are applied on $\Gamma_1$, whereas suction microjets are applied on $\Gamma_2$ and $\Gamma_3$. Pressure sensor points are indicated by blue rectangles located within the isolator.} 
\label{fig:Fig085}
\end{figure}
Figure \ref{fig:Fig085} shows the schematic of microjet placement along the wall. Velocity profiles of the microjets and the mass-balance equation are provided in Appendix \ref{appC}. Blowing microjets with a learnable jet angle~($\beta$) are applied on~$\Gamma_1$, whereas suction microjets are applied on~$\Gamma_2$ and~$\Gamma_3$. Pressure sensor locations are indicated by blue rectangles within the isolator. The state $s_t$ and the corresponding reward $r_t$ are provided to the distributed DRL environments, which map the state vector $s_t$ to actions $a_t$ that are subsequently applied to the microjets.

Suction jets implemented within the isolator of a hypersonic intake do \emph{not} increase the mass flow rate at the isolator exit; rather, they deliberately reduce the core-flow mass flux to enhance inlet operability and delay unstart. From a control-volume perspective, the steady mass conservation relation for an isolator with suction is $\dot{m}_{\text{in}} = \dot{m}_{\text{exit}} + \dot{m}_{\text{suction}},$
where $\dot{m}_{\text{suction}} > 0$ represents the mass extracted through the suction jets. Consequently, the mass flow delivered to the combustor, $\dot{m}_{\text{exit}}$, is lower than the mass captured at the inlet. The purpose of suction is therefore not to augment mass flow, but to modify the internal flow structure such that the remaining core flow can be processed without choking. Physically, suction alters the isolator flow by weakening the shock train, reducing boundary-layer growth and blockage, and lowering the effective static pressure transmitted upstream from the combustor. These effects decrease the likelihood of the flow at the effective throat reaching sonic conditions, allowing the inlet to remain started at back pressures exceeding the classical Kantrowitz limit for the unbled configuration. In this sense, suction increases the \emph{effective pressure-rise tolerance} of the inlet–isolator system, even though the absolute mass flow reaching the isolator exit is reduced. The apparent improvement in operability arises because the choking condition, $M=1$, is postponed to higher downstream pressures for a smaller core mass flow, rather than because the inlet can pass more mass. Accordingly, isolator suction embodies a trade-off between stability and performance: it sacrifices delivered mass flow and thrust potential in exchange for delayed Kantrowitz unstarting and enhanced robustness under high or unsteady combustor back pressure, as encountered in dual-mode scramjets or pressure-gain combustors. The fundamental Kantrowitz constraint remains satisfied, but for a modified control volume with reduced throughflow, highlighting that suction enhances operability by relieving the mass-flow limitation rather than by increasing the isolator exit mass flow.

\subsection{DRL State-Action Space}
We formulate the hypersonic unstart control problem as a Markov Decision Process (MDP) and employ an off-policy algorithm to identify optimal jet actuation strategies.  
In this framework, the flow control problem is represented by the tuple $(\mathcal{S}, \mathcal{A}, \mathcal{R}, \gamma)$, where $\mathcal{S}$ and $\mathcal{A}$ denote the state and action spaces, $\mathcal{R}$ is the reward function, and $\gamma = 0.99$ is the discount factor.  
This MDP formulation naturally defines the interaction between the DRL agent and the hypersonic environment: the agent observes the system through a continuous state space and selects actions within a continuous action space, designed to provide sufficient observability of shock dynamics inside the isolator while enabling precise control via the microjet actuators.

\vspace{0.2cm}
\noindent \textbf{State Space:} The state vector $s_t \in \mathcal{S}$ is constructed from instantaneous wall-pressure measurements, which serve as the primary observable for detecting shock position and boundary-layer separation. Pressure is sampled at $N=100$ discrete sensor locations uniformly distributed along the isolator walls, as illustrated in Fig.~\ref{fig:Fig085} (blue rectangular markers). To facilitate efficient neural-network training, the raw pressure signals are normalized by the freestream static pressure $p_{\infty}$. The state at time step $t$ is thus defined as
\[
s_t =
\left[
\frac{p_1}{p_{\infty}},
\frac{p_2}{p_{\infty}},
\ldots,
\frac{p_N}{p_{\infty}}
\right]^{\mathrm{T}}
\in \mathcal{S} \subseteq \mathbb{R}^N .
\]
This high-dimensional observation enables the agent to capture the spatial pressure distribution and the evolution of the shock train along the isolator.

\vspace{0.1cm}
\noindent \textbf{Action Space:}
Based on the observed state $s_t$, the agent selects a continuous control action $a_t\in \mathcal{A}$ to regulate the supersonic microjet actuation. The action space is defined as
\[
a_t =
\left[
Q_b,
Q_{s1},
Q_{s2},
\beta
\right]
\in \mathcal{A} \subseteq \mathbb{R}^4 ,
\]
where $Q_b$ denotes the mass flux of the blowing microjets located on the compression ramp ($\Gamma_1$), $Q_{s1}$ and $Q_{s2}$ represent the mass fluxes of the suction microjets on the isolator floor ($\Gamma_2$) and the step top ($\Gamma_3$), respectively, and $\beta$ is the injection angle of the blowing jets measured relative to the wall normal. By manipulating these four continuous control variables, the policy dynamically adjusts both the magnitude and direction of actuation to stabilize the shock train and suppress flow separation.

\subsection{Reward Design and Learning Objective}
The reward function $r_t = \mathcal{R}(s_t, a_t)$ is designed to minimize the deviation from the baseline pressure profile while penalizing excessive control usage and rapid actuation. It is defined as:

\begin{equation}
r_t = - \left\| \frac{\mathbf{p}/p_{\infty} - \mathbf{p}_{\text{baseline}}/p_{\infty}}{\mathbf{p}_{\text{baseline}}/p_{\infty}} \right\|_2^2 - w_P \left( \frac{P_t}{\max(P)} \right)^2- w_R  \left(\frac{|Q^t - Q^{t-1}|}{\max(Q)\Delta t} \right)\end{equation}

where $\mathbf{p}_t \in \mathbb{R}^{100}$ and $\mathbf{p}_{\text{baseline}} \in \mathbb{R}^{100}$ denote the vectors of instantaneous and baseline pressures, respectively, and $p_{\infty}$ denotes the freestream pressure constant of 900 Pa used for normalization. The baseline pressure values were obtained from the no-throttling case, in which the flow within the isolator sustains a quasi-steady shock train. The term $\|\cdot\|_2$ denotes the negative squared $\ell_2$ norm of the element-wise relative pressure error. The mass flux is represented by $Q = \sum_{\forall i
} Q_i$, while the kinetic power of the jets is formulated as:
\begin{equation}
P = \sum_{\forall i
}\frac{1}{2} \frac{Q_i^3}{\rho_i^2 A_i^2}
\end{equation}
where the index $i$ denotes the three distinct jet groups configured in the environment: blowing jets along the outer ramp ($\Gamma_1$ wall boundary), suction jets along the isolator floor ($\Gamma_2$ wall boundary), and suction jets on the step top ($\Gamma_3$ wall boundary) (Fig.~\ref{fig:Fig085}). 
The weights for the penalty terms were set to $w_P = 0.005$ and $w_R = 0.05$.
The following observations are made:
\begin{itemize}
    \item The reward function directly promotes shock-train stabilization by minimizing the squared $\ell_2$ norm of the relative wall-pressure deviation from a quasi-steady baseline profile, providing dense and physically meaningful feedback.
    \item Normalization by the baseline and freestream pressure renders the reward scale-invariant, improving training robustness across varying flow conditions.
    \item The jet power penalty discourages energetically inefficient control strategies, encouraging minimal actuation while maintaining flow stability.
    \item The action-rate penalty (last term) suppresses chattering and enforces smooth temporal control, enhancing numerical stability and physical realizability.
    \item The reward is negative-definite and bounded above, attaining its maximum value $r_t \approx 0$ when the pressure field matches the baseline exactly.
\end{itemize}
The goal of the DRL algorithm is to learn a policy $\pi_{\theta}$ that
maximizes the expected cumulative discounted return,
\begin{equation}
\max_{\theta} \; J(\theta)
= \mathbb{E}_{\tau \sim \pi_\theta}
\left[
\sum_{t=0}^{\infty} \gamma^t r_t
\right],
\end{equation}
where $\theta$ denotes the parameters of the policy and $\tau = (s_0, a_0, r_0, s_1, a_1, r_1, \dots)$
represents a trajectory generated by interacting with the environment under policy
$\pi_\theta$. The expectation $\mathbb{E}_{\tau \sim \pi_\theta}[\cdot]$ is taken over all
trajectories induced by the policy and the environment dynamics.

\subsection{Off-policy Algorithms}
In this work, we employ off-policy DRL algorithms to learn the optimal control strategy. Fundamentally, off-policy learning is defined by the decoupling of the behavior policy (used to explore the environment and collect data) from the target policy (the optimal policy being learned). This allows the agent to update and improve its policy using experience transitions $(s_t, a_t, r_t, s_{t+1})$ that were generated by earlier, different versions of the policy, rather than requiring data only from the current policy iteration. The primary reason behind selecting off-policy algorithms over on-policy methods (such as Proximal Policy Optimization \cite{schulman2017proximal}) is sample efficiency. In our framework, the environment is a high-fidelity, fifth-order CFD solver, where every step is computationally very expensive. On-policy algorithms are inherently sample-inefficient because they require new data for every gradient update and typically discard experiences immediately after use. In contrast, off-policy methods utilize a Replay Buffer to store past transitions, which allows the agent to reuse the same data multiple times for training. This significantly reduces the total number of CFD simulations required to converge to an optimal policy, making the training process computationally feasible. In Replay Buffer, a data structure that stores experience tuples $(s_t, a_t, r_t, s_{t+1})$ collected during the agent's interaction with the environment. In our framework, this collection process is parallelized to maximize throughput. As shown in Figure \ref{fig:Fig085}, multiple independent CFD environments run simultaneously, executing actions and computing states in parallel. These parallel environments asynchronously push their transitions into a shared, global replay buffer. The central agent then samples random mini-batches from this dataset to update the neural network weights.

This approach provides an important advantage, which is a drastic improvement of sample efficiency by allowing the agent to learn from sparse, high-value experiences multiple times. However, the use of a large or unbounded replay buffer might cause \textit{distributional shift}. If the buffer retains extremely old experiences generated by a very old or random policy, the agent may sample wrong data that no longer reflects the current policy. In many control tasks, training on such outdated experiences can mislead the value function estimation and delay or fail convergence.  In our specific case, using older data is not a problem because the state representation of the flow does not change significantly till it encounters unstart. Since the state dynamics do not show long-term consequences until the inlet has spillage or unstarts, a transition recorded early in training is just as relevant as one recorded later. 
We specifically investigate two distinct off-policy algorithms: Twin Delayed Deep Deterministic Policy Gradient (TD3) \cite{fujimoto2018addressing} and Soft Actor-Critic (SAC) \cite{haarnoja2018soft}. 
\begin{itemize}
\item \textbf{TD3:} This is a deterministic algorithm designed to address the overestimation bias inherent in standard actor-critic methods. It uses clipped double Q-learning (using the minimum of two critic estimates) and delayed policy updates (updating the actor less frequently than the critic) to stabilize the learning of the value function.
\item \textbf{SAC:} Conversely, SAC operates on the maximum entropy framework, learning a stochastic policy. In addition to maximizing the expected cumulative reward, SAC maximizes the entropy of the policy, which serves as a regularizer to encourage robust exploration. This stochasticity is particularly valuable for preventing the agent from prematurely converging to suboptimal control strategies.
\end{itemize} 
More discussion and pseudocode for both TD3 and SAC are provided in Appendix \ref{appD} and Appendix \ref{appE}, respectively.

\section{Results}
In this section, we present a systematic comparison of the TD3 and SAC controllers across different TRs. We further evaluate zero-shot generalization performance with and without sensor noise. In addition, an optimal reduced sensor set is identified, and zero-shot generalization is re-evaluated under noisy measurements across unseen back-pressure conditions and Reynolds numbers.

\subsection{Results for varying TRs}
\begin{figure}[htpb]
\centering
\includegraphics[scale=0.20, clip=true]{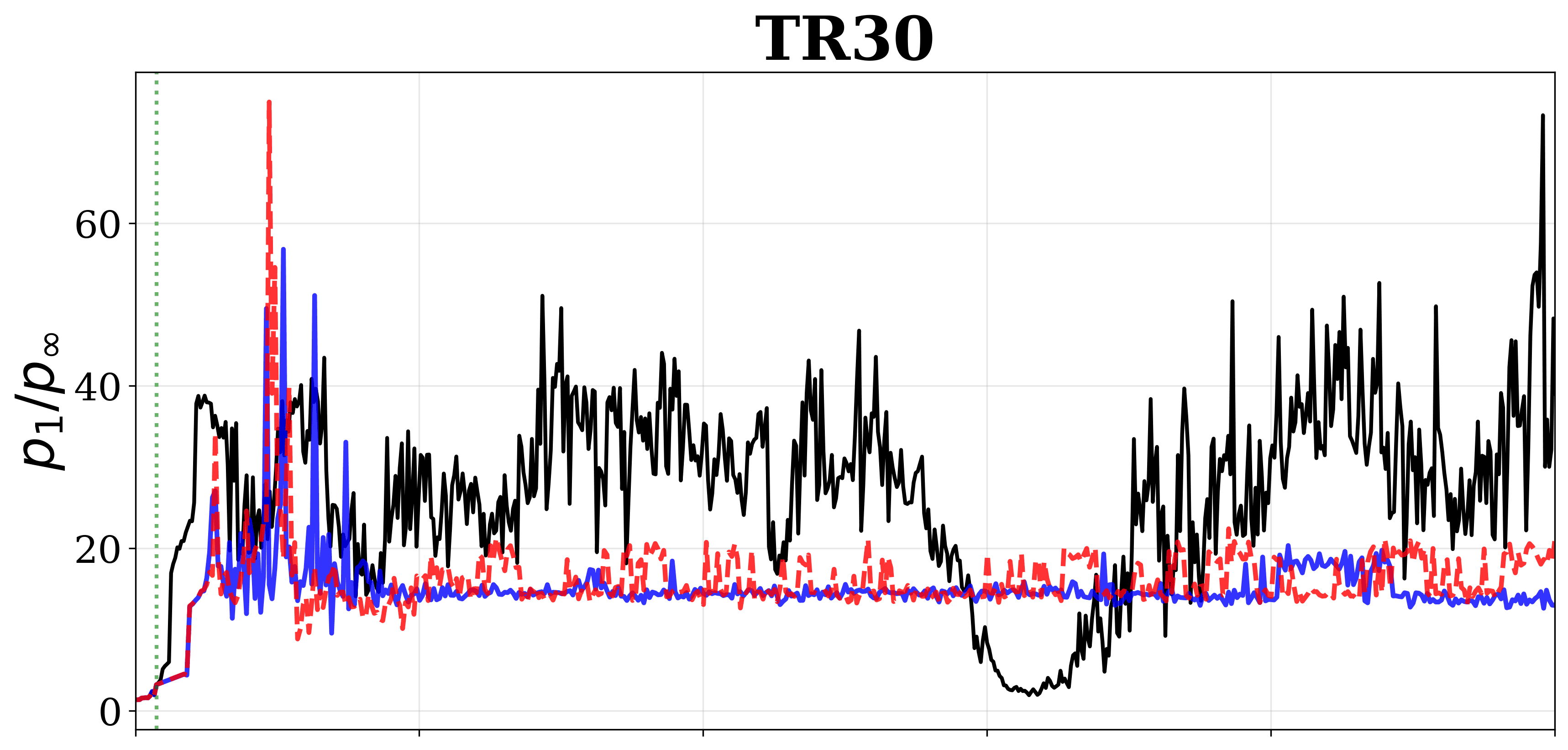}
\includegraphics[scale=0.19, clip=true]{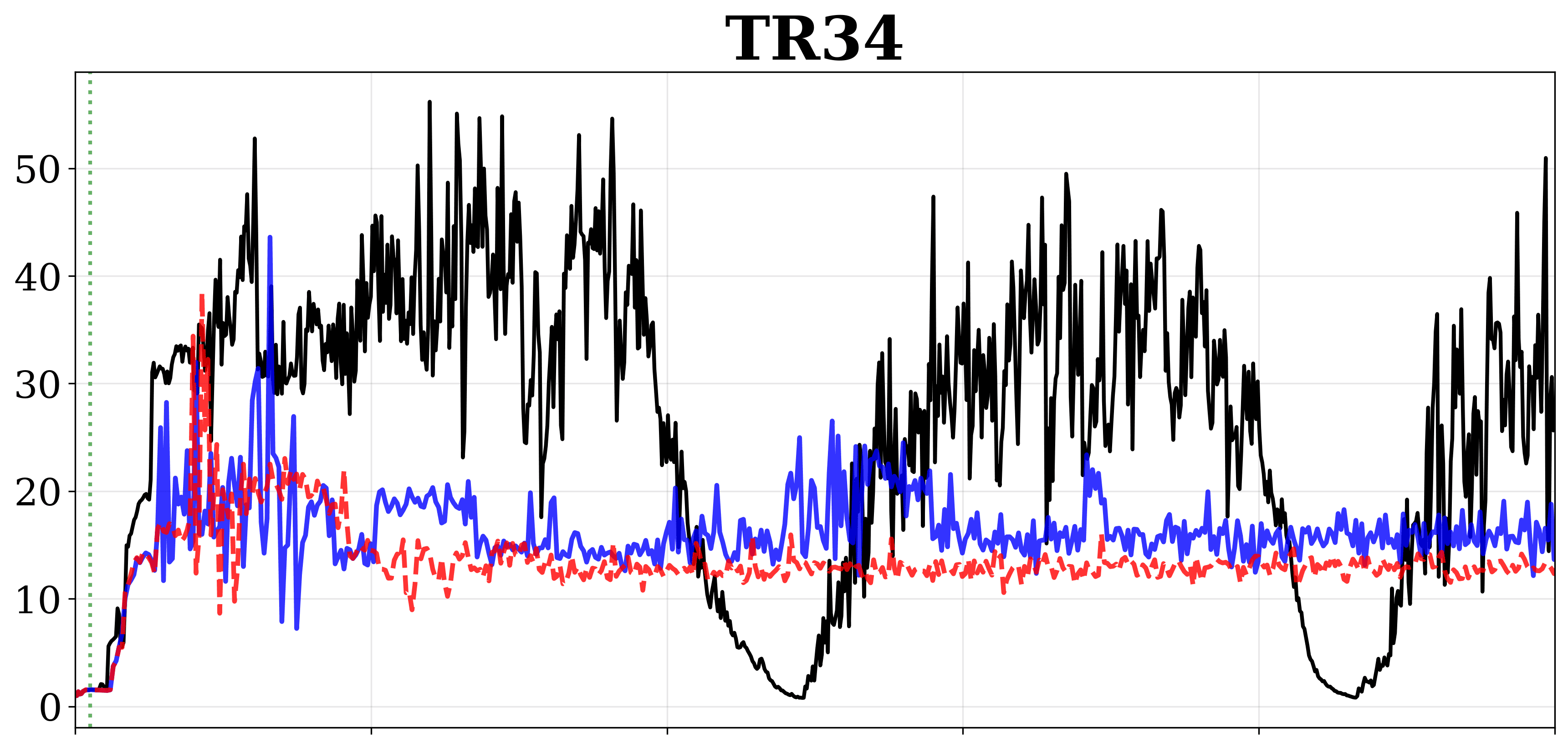}
\includegraphics[scale=0.19, clip=true]{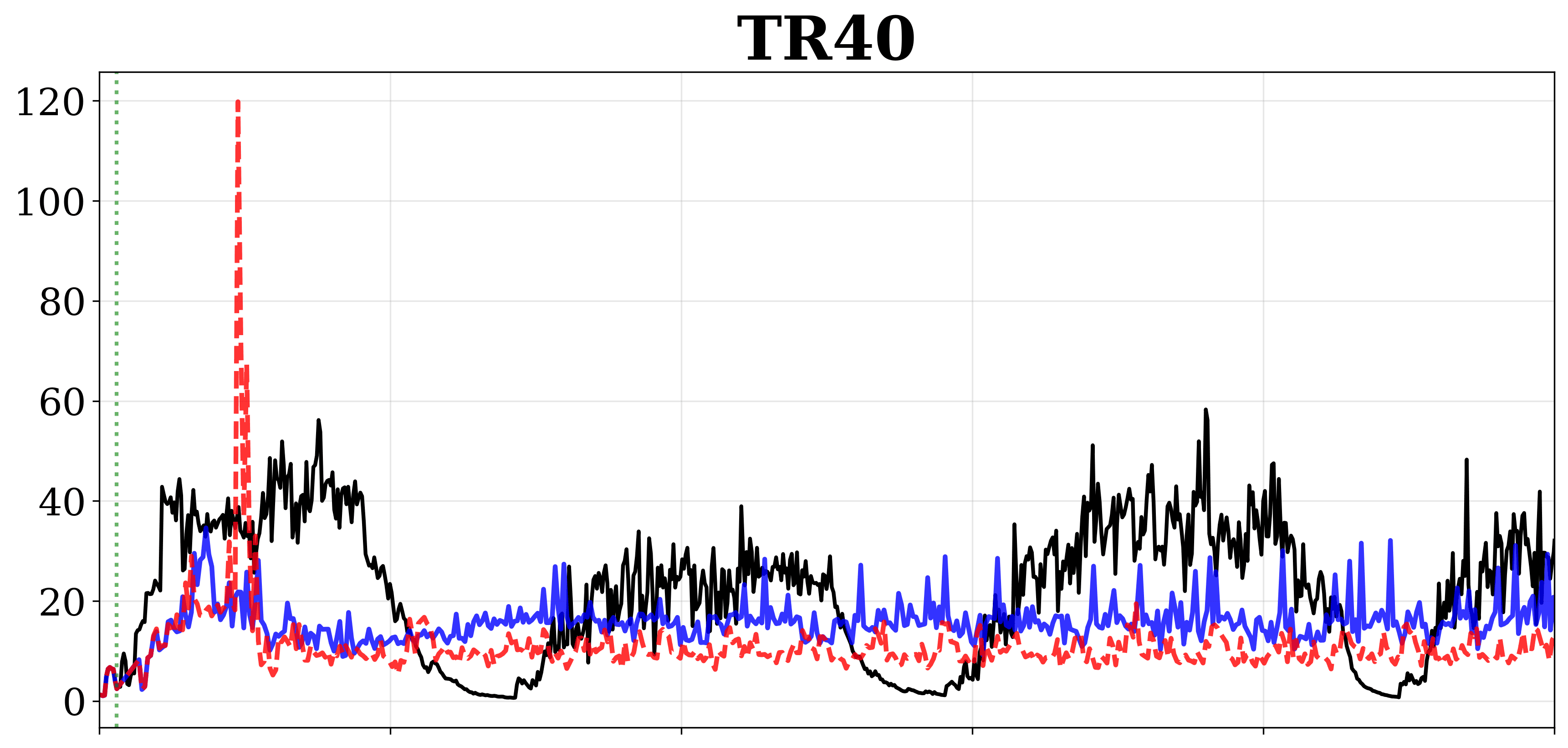}
\includegraphics[scale=0.20, clip=true]{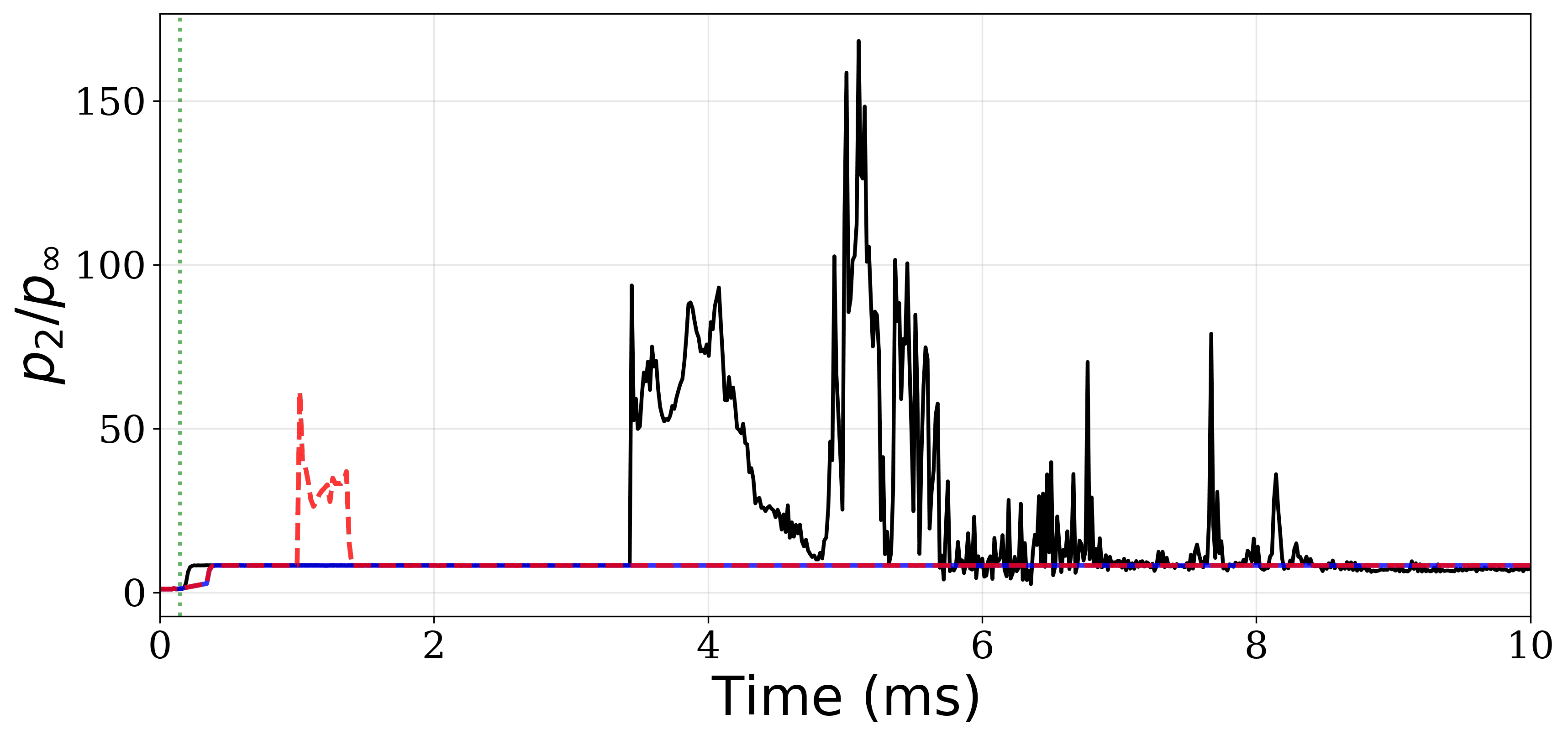}
\includegraphics[scale=0.19, clip=true]{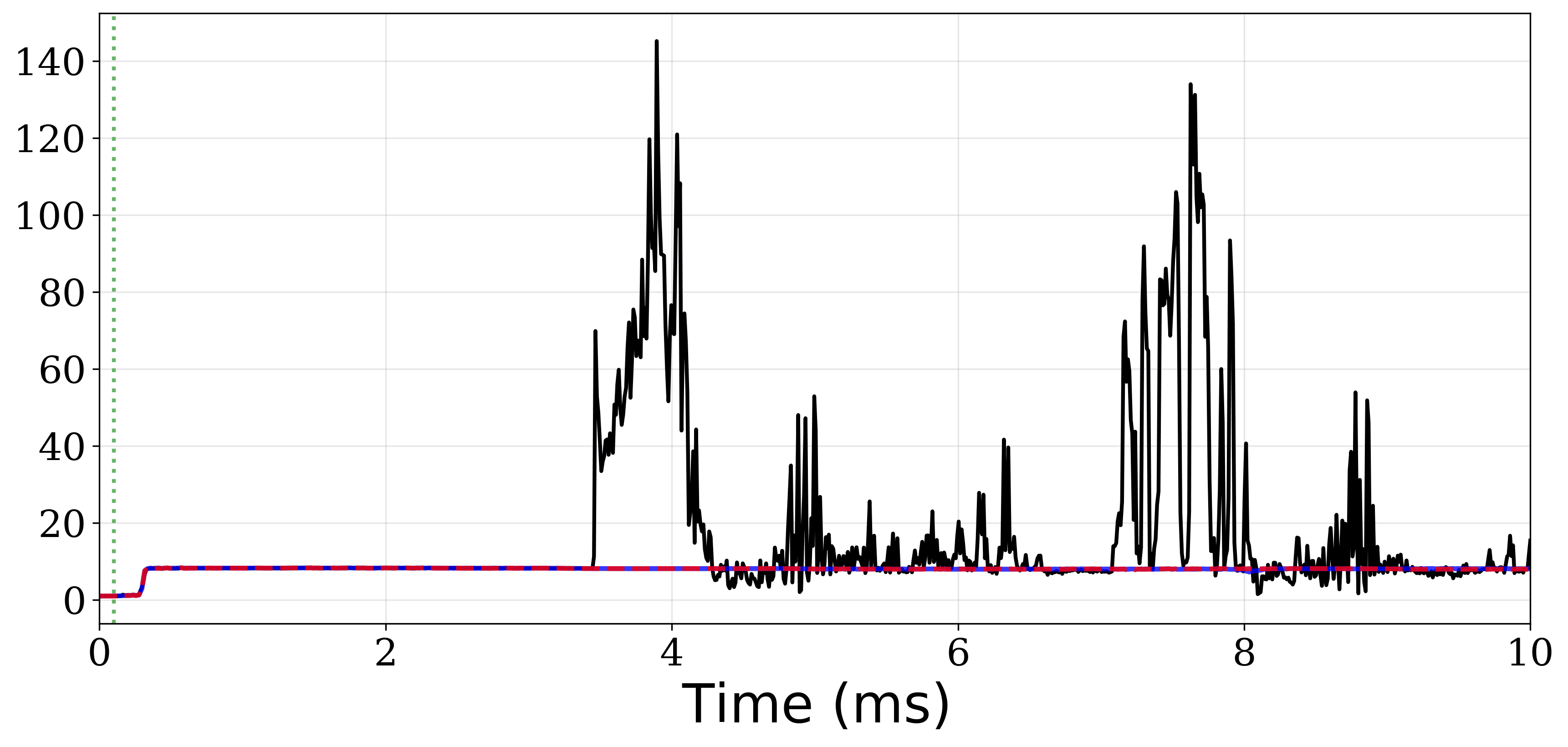}
\includegraphics[scale=0.19, clip=true]{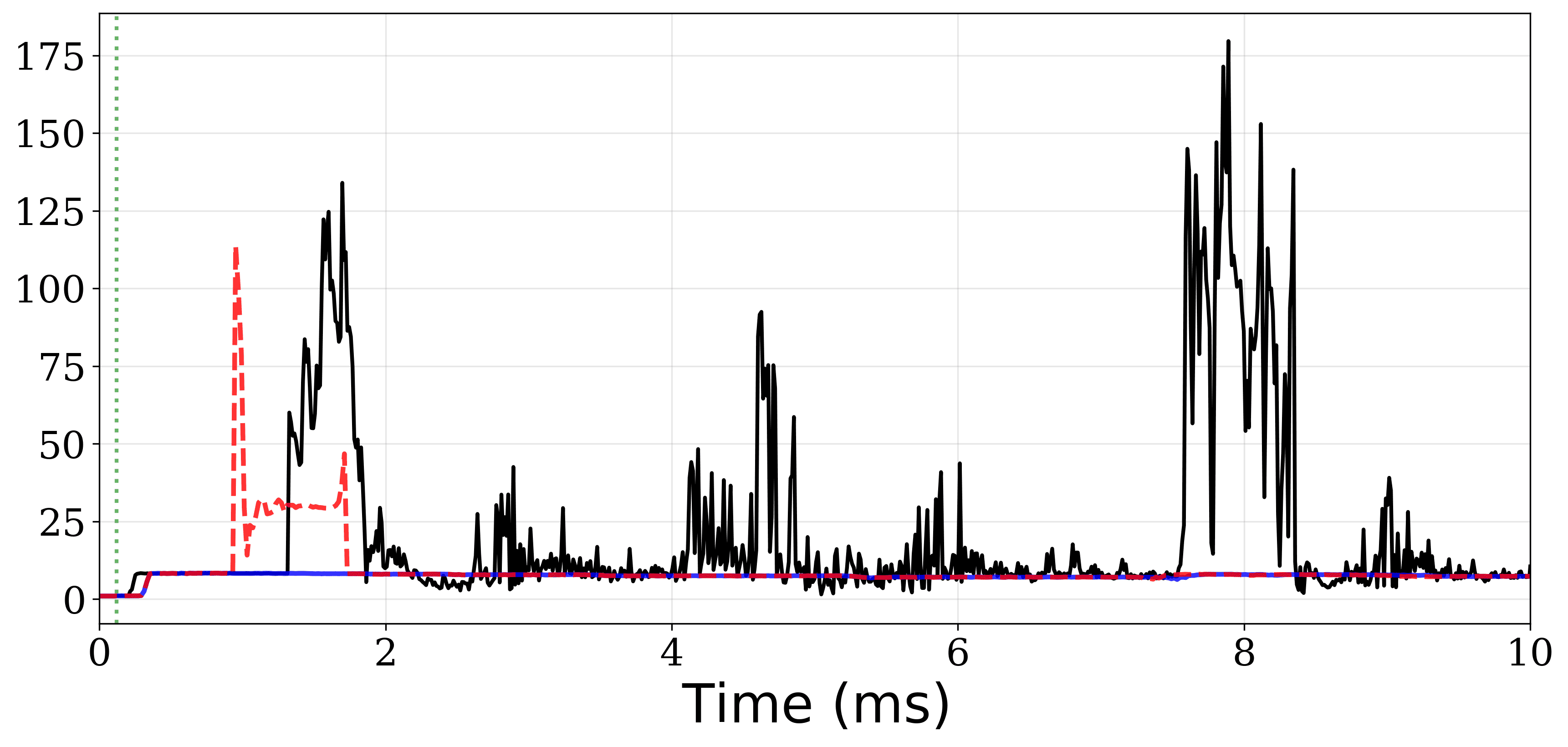}
\includegraphics[scale=0.22, clip=true]{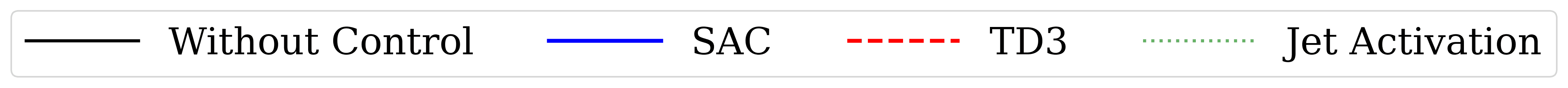}
\caption{Time evolution of normalized pressure ($p/p_{\infty}$) at monitoring points 1 (top row) and 2 (bottom row) for throttling ratio - TR30, TR34 and TR40. The plots compare the uncontrolled baseline (black) against active control using SAC (blue) and TD3 (red). The vertical green line shows the jet activation time.}
\label{fig:p1p2_all_TR}
\end{figure}

The TR serves as the primary parameter for controlling the back-pressure within the intake. Physically, a higher TR corresponds to a greater blockage at the exit, which induces higher back-pressure and forces the shock system to move further upstream. In the uncontrolled baseline scenario, this accumulation of back-pressure destabilizes the inlet. As shown in figure \ref{fig:p1p2_all_TR}, the onset of unstart is marked by a strong discontinuity in the pressure value at probe $p_2$, indicating the rapid upstream propagation of the shock train and the occurrence of subsonic spillage at the cowl lip. All three different TR have unstarted and higher TR (e.g., TR40) trigger this spillage significantly earlier than lower ratios (e.g., TR30). Consequently, the pressure at probe $p_1$ rises significantly, which reflects the loss of stable flow. The active control system is expected to counter this behavior by modulating the microjets to maintain the shock train within the isolator, thereby preventing the unstart phenomena. 
\begin{figure}
\centering
\includegraphics[scale=0.225, clip=true, trim= 0cm 0cm 0cm 0cm]{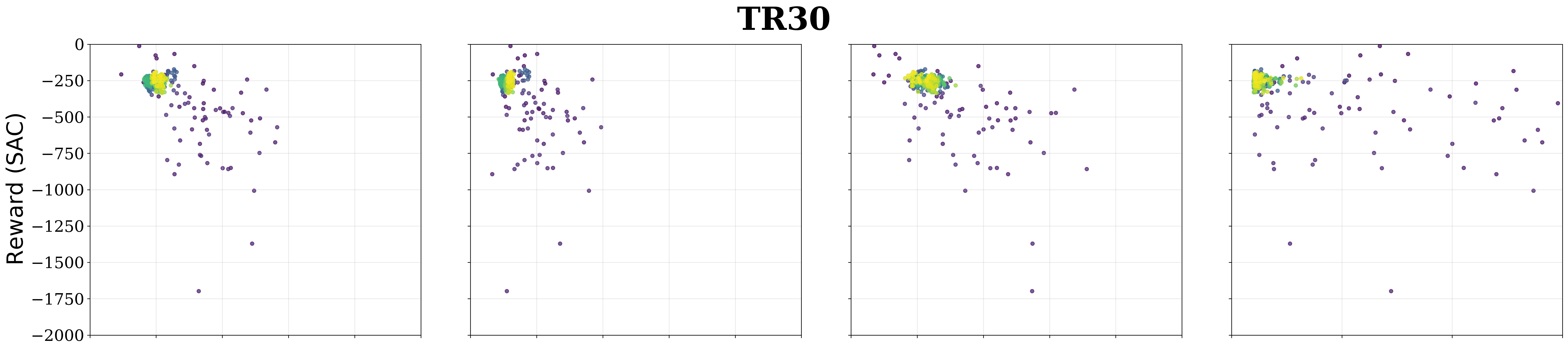}
\includegraphics[scale=0.225, clip=true]{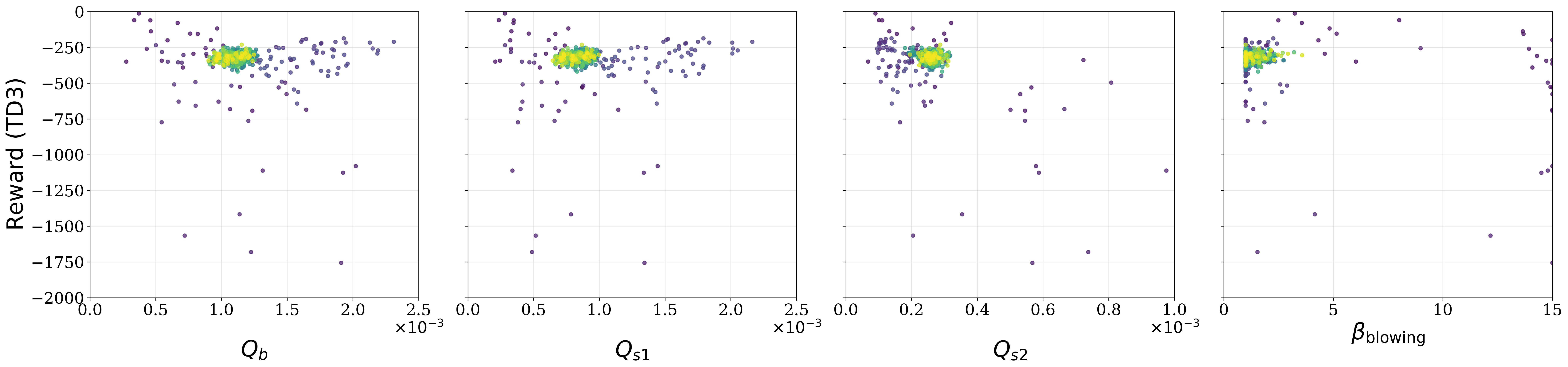}
\includegraphics[scale=0.225, clip=true, trim= 0cm 0cm 0cm 0cm]{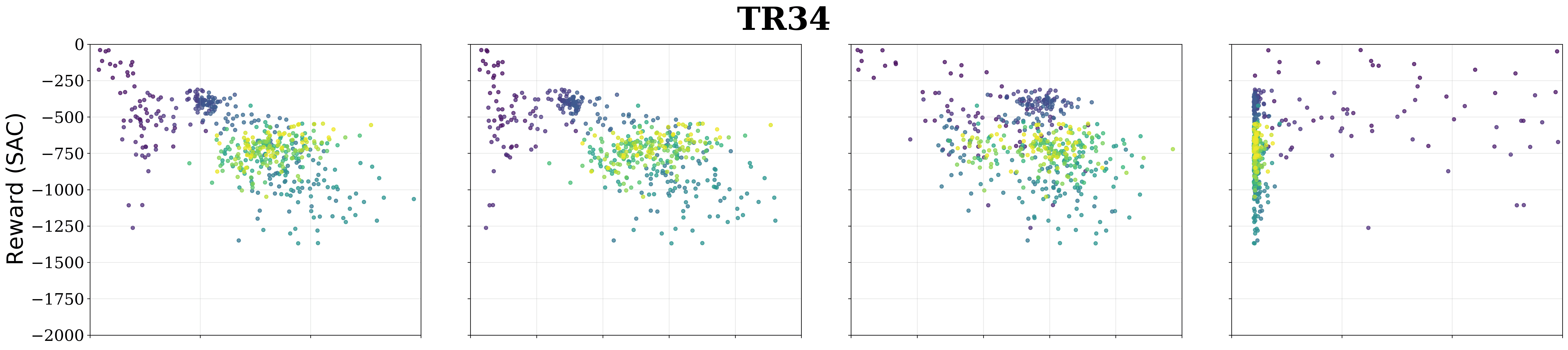}
\includegraphics[scale=0.225, clip=true]{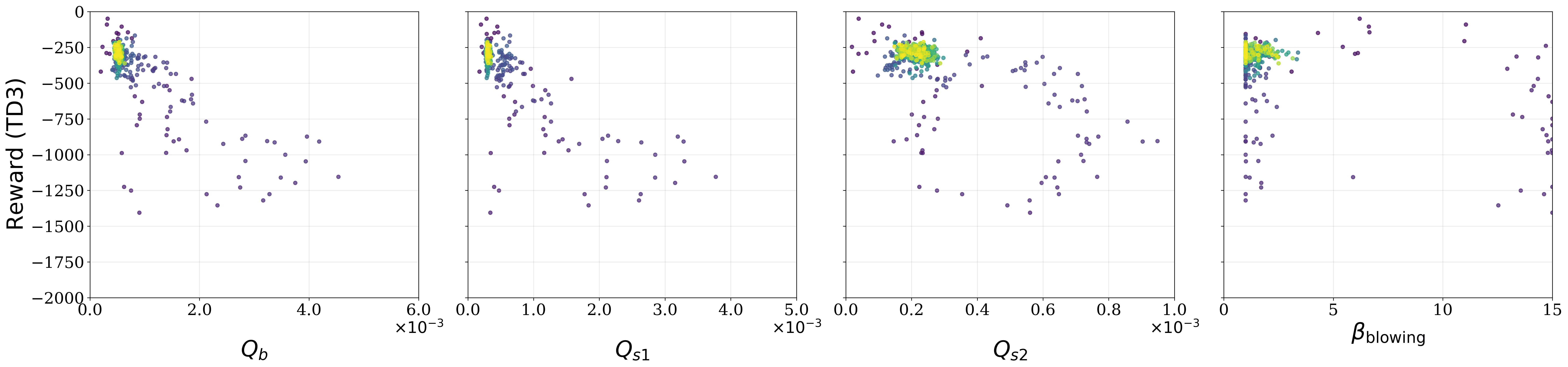}
\includegraphics[scale=0.225, clip=true, trim= 0cm 0cm 0cm 0cm]{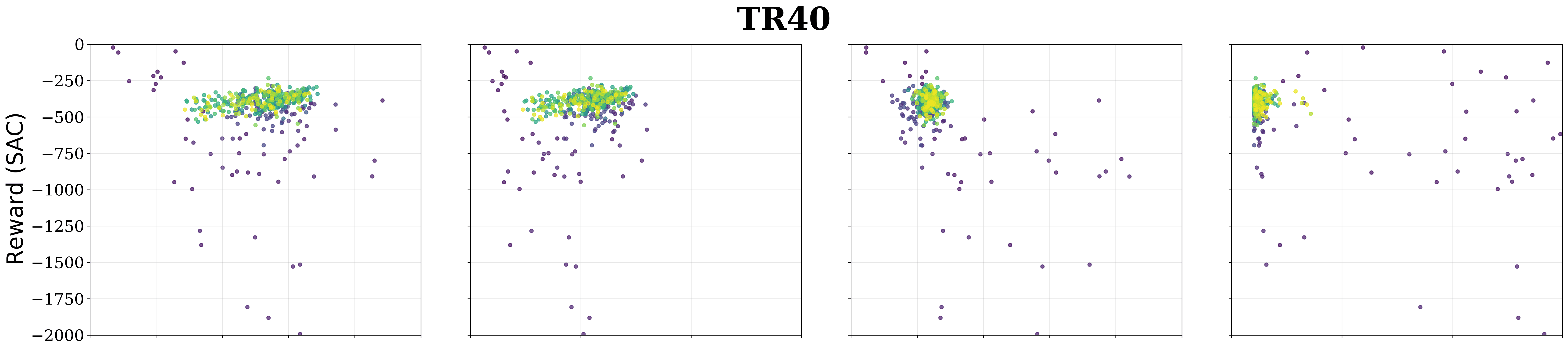}
\includegraphics[scale=0.225, clip=true]{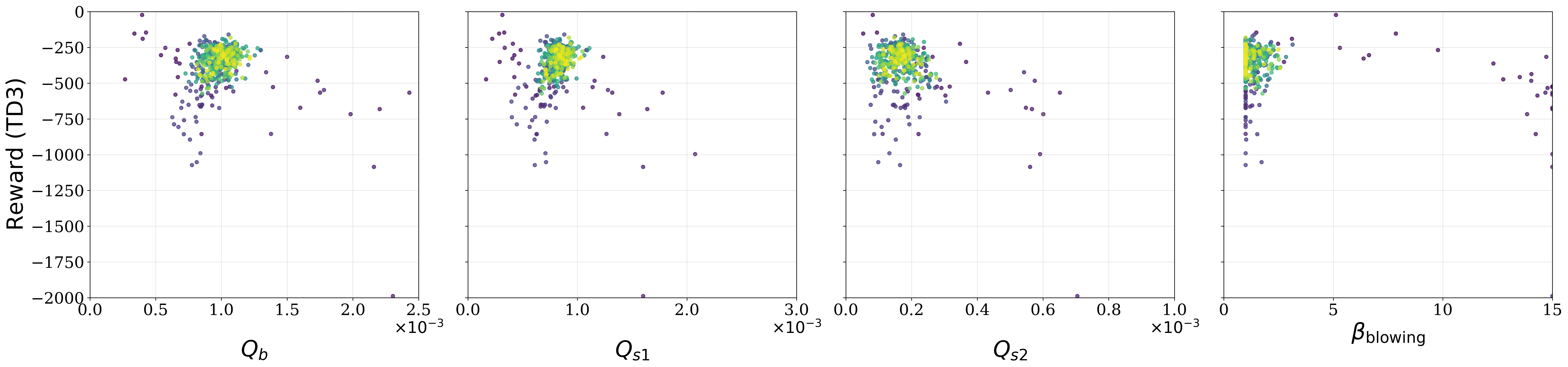}
\includegraphics[scale=0.28, clip=true]{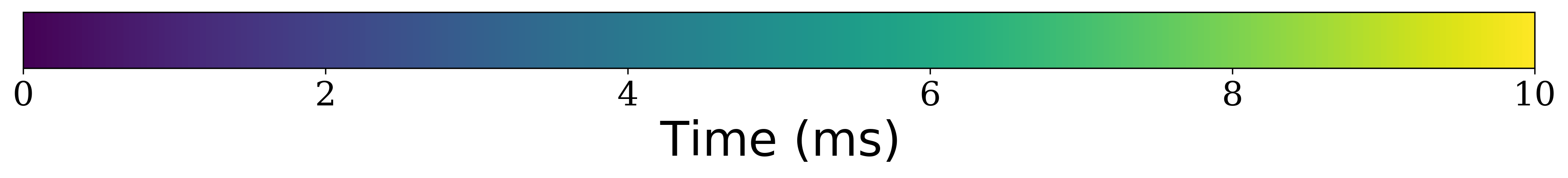}
\caption{Distribution of instantaneous reward against control parameters: blowing mass flux ($Q_b$), suction mass fluxes ($Q_{s1}, Q_{s2}$), and blowing angle ($\beta_{\text{blowing}}$) for throttling ratio - TR30, TR34 and TR50. The top row represents the SAC agent, and the bottom row represents the TD3 agent. The color gradient indicates the physical time.}
\label{fig:reward_all_TR}
\end{figure}
To achieve this control, we use Soft Actor-Critic (SAC) and Twin Delayed Deep Deterministic Policy Gradient (TD3). As discussed earlier, the choice of off-policy methods is driven by the need for sample efficiency. Since high-fidelity CFD simulations are computationally expensive, it is important to maximize the learning utility of every generated data sample stored in the replay buffer. For this analysis, both SAC and TD3 agents were trained independently across three distinct throttling ratios: TR30, TR34, and TR40, to evaluate their performance under varying back-pressure intensity. The time evolution of the normalized wall pressures at probes $p_1$ and $p_2$ provides a direct measure of the intake's stability and the onset of unstart. As established in the baseline analysis (Figure \ref{fig:FigTHRO}), the pressure at $p_2$ acts as the primary indicator for spillage. Under uncontrolled conditions, increasing the TR accelerates the upstream propagation of the shock train. This is manifested as a sharp, discontinuous rise in $p_2/p_{\infty}$, which signifies the passing of the shock over the probe, followed by large-amplitude fluctuations characteristic. The effectiveness of these trained agents is shown in Figure \ref{fig:p1p2_all_TR}, which compares the time evolution of normalized pressures at TR30, TR34 and TR40. While both agents successfully stabilize the flow where the baseline fails, distinct behavioral differences emerge between the algorithms, specifically at TR30 and TR40. The SAC agent shows better stability across all throttling ratios. For each TR30, TR34 and TR40, the agent maintains a control which maintain pressure at $p_1$. The pressure at $p_2$ remains flat and bounded near the baseline levels, which indicates that the shock train is firmly within the isolator without any upstream movement. In contrast, the TD3 agent shows a \textit{early spillage} at TR30 and TR40. This is marked by a transient pressure spike at $p_2$ (red dashed line) shortly after activation, reaching values of approximately $60 p_{\infty}$ at TR30 and exceeding $100 p_{\infty}$ at TR40. These spikes indicate that the shock system briefly moved upstream, which triggered unstart. However, the policy quickly recovered and restarted, forcing the shock train back into the isolator. These dynamics are also visible in the upstream pressure probe, $p_1/p_{\infty}$ (top row).

To explain the physical mechanisms explored by the agents, Fig.~\ref{fig:reward_all_TR} shows the distribution of instantaneous rewards as a function of the four primary control parameters, blowing mass flux ($Q_b$), suction mass fluxes ($Q_{s1}$ and $Q_{s2}$), and blowing angle ($\beta_{\mathrm{blowing}}$), over the physical time interval from 0 to 10~ms.
 Since the reward function is designed to penalize pressure variations from the baseline state (without any throttling or back-pressure), the instantaneous reward serves as a direct measure for the back-pressure intensity within the isolator. The scatter plots show distinct differences in the learning dynamics of the two algorithms. The SAC agent (top rows) shows a broad, dispersed distribution of action-reward pairs, particularly in the early phases of the simulation (purple/blue points). This behavior is characteristic of SAC’s maximum entropy formulation, which actively encourages exploration of the parameter space to prevent premature convergence to suboptimal policies. In contrast, the TD3 agent (bottom rows) displays tightly concentrated clusters of action points, which reflects its deterministic nature and tendency to converge to a specific control strategy with minimal ongoing exploration. The action-reward distribution shows that for a fixed high reward, which corresponds to maintaining a stable back-pressure, the SAC agent takes a range of actions rather than converging to a single point. This observation indicates that to sustain the shock train system, the agent must dynamically adjust its blowing and suction mass fluxes ($Q_b, Q_{s1}, Q_{s2}$) within specific, effective intervals. Regardless of the policy or throttling ratio, the optimal policies consistently converge towards small blowing angles ($\beta_{blowing}$), typically tangential to the wall. This preference suggests that the most effective control mechanism involves energizing the boundary layer through tangential momentum addition rather than blowing at an high angle. Large blowing angles would cause flow separation, induce strong, near-normal shocks that could further destabilize the shock train, leading to the very unstart event the agent aims to prevent.

\begin{figure}[htpb]

\centerline{
\includegraphics[scale=0.24]{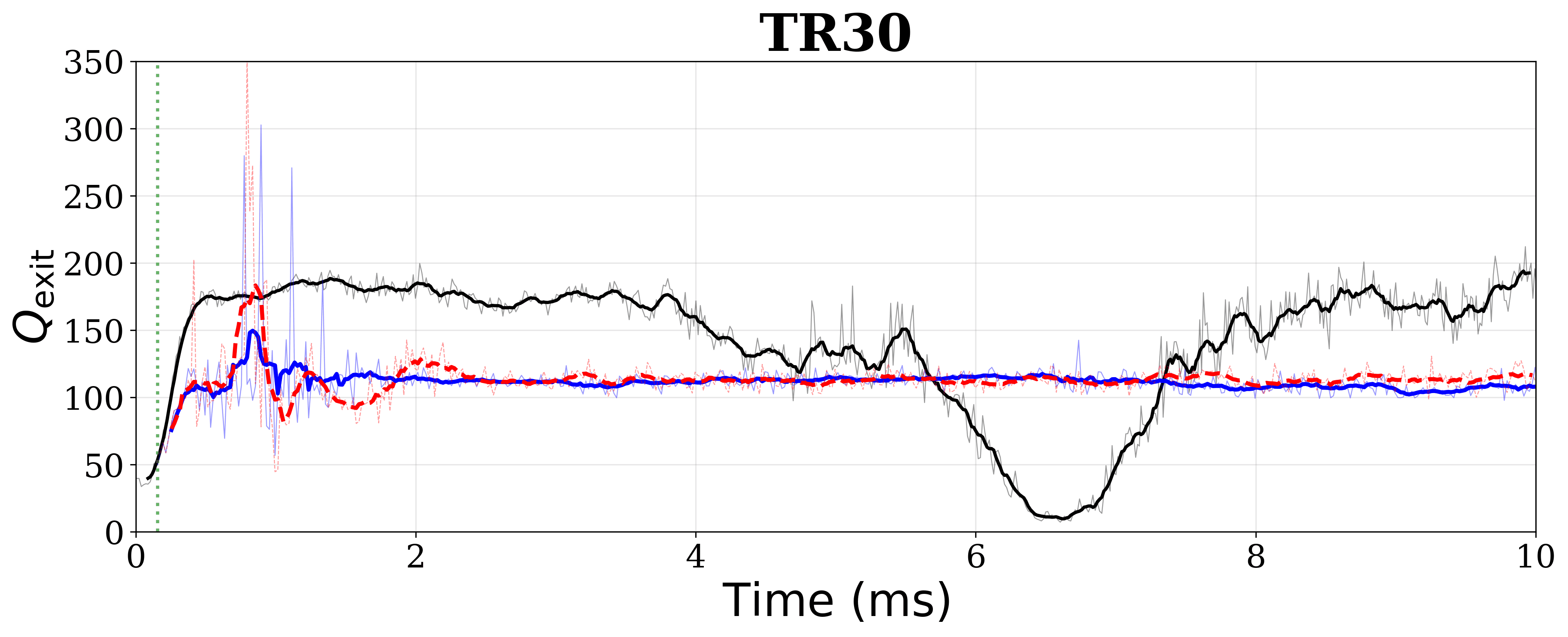}\hspace{1em}
\includegraphics[scale=0.24]{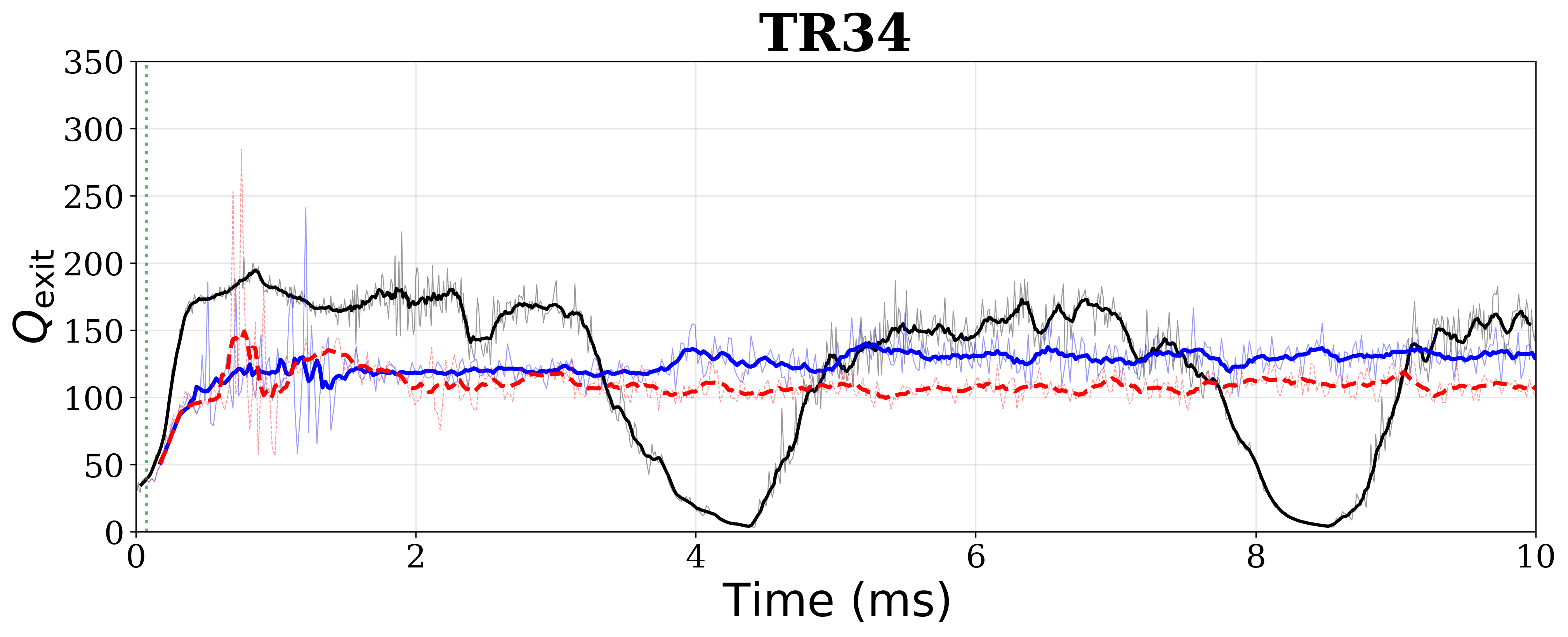}
}
\centerline{
\includegraphics[scale=0.24]{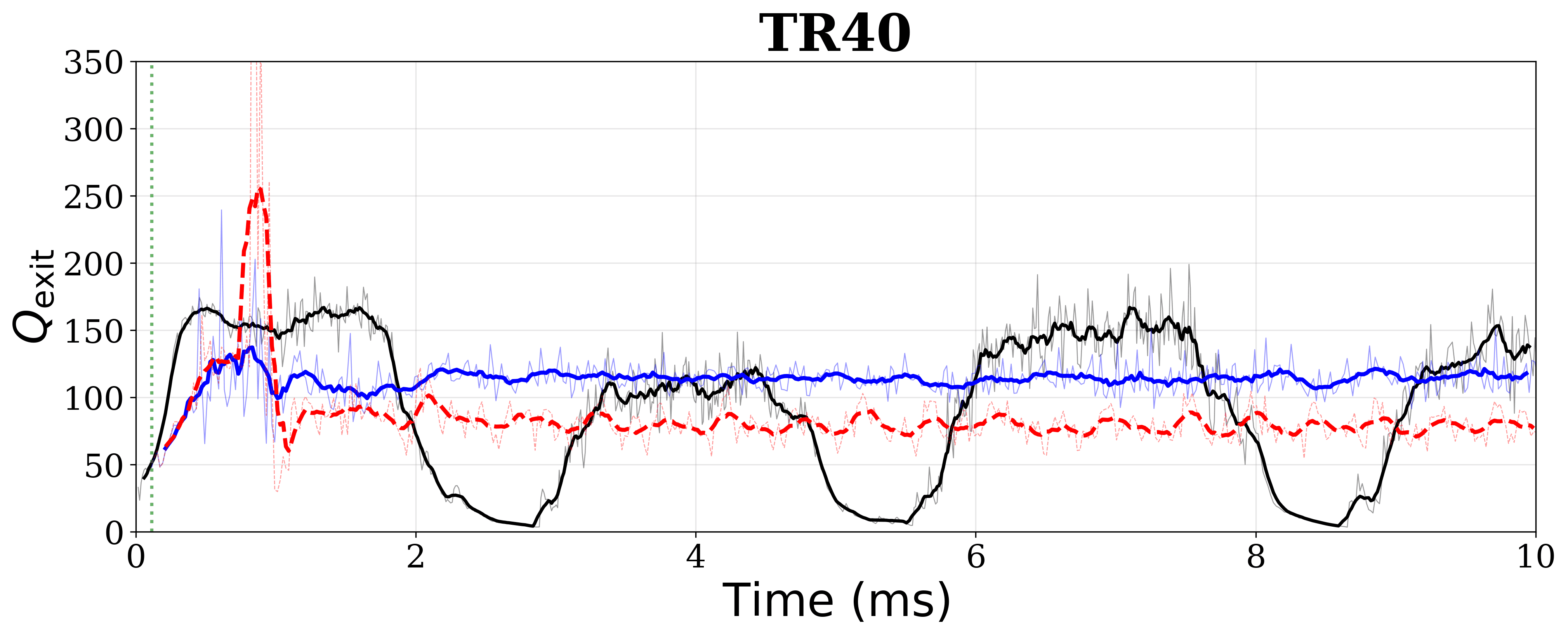}
}
\centerline{
\includegraphics[scale=0.20]{figures/legend_pressure_massflow.png}
}
\caption{Evolution of the mass flow rate at the isolator exit ($Q_{\text{exit}}$) for throttling ratios - TR30, TR34 and TR40. Raw data is shown with low opacity, overlaid by a 10-point moving average (high opacity).}
\label{fig:mass_all_TR}

\end{figure}

To further validate the intake's operational status, we examine the time evolution of the mass flow rate calculated at the isolator exit, denoted as $Q_{exit}$. This metric serves as a definitive indicator of the flow choking status; a steady, high mass flow signifies that the inlet is \textit{started} and effectively capturing the freestream air, whereas a significant drop indicates \textit{unstart}, where the shock train has been expelled, causing massive upstream spillage and loss of capture capability. Figure \ref{fig:mass_all_TR} shows the evolution of $Q_{exit}$ for TR30, TR34, and TR40. In the absence of control (black), the mass flow rate suffers degradation. For example, at TR34, $Q_{exit}$ drops significantly after the unstart event (approximately $t=3$ ms), which confirms that the blockage has forced the flow to spill around the intake cowl. At higher throttling ratios (TR34 and TR40), the baseline flow shows large-amplitude oscillations, where the intake alternates between start and unstart. The deployment of active flow control fundamentally changes this trajectory. Both the SAC (blue) and TD3 (red) agents successfully prevent these oscillations and maintain $Q_{exit}$ at a quasi-steady value throughout the simulation. In summary, suction microjets reduce the core mass flow rate at the isolator exit but delay Kantrowitz unstart by weakening the shock train, reducing boundary-layer blockage, and lowering the effective upstream pressure. This modification of the internal flow allows the inlet to remain started at higher back pressures, enhancing operability without increasing the absolute mass flow. In essence, suction improves stability at the expense of delivered mass flow.

Synthesizing the comparative analysis of the pressure evolution, reward distribution, and exit mass flow rate across all throttling ratios (TR30, TR34, TR40), we can derive the following conclusions regarding the performance and suitability of the two DRL algorithms:

\begin{itemize}
    \item \textbf{Superior Stability and Transient Response (SAC):} The SAC agent shows better stability across all throttling ratios and maintains a consistent shock train with no spillage. In contrast, the TD3 agent triggers early spillage at TR30 and TR40, characterized by sharp pressure spikes at $p_2$ and high-frequency oscillations in the exit mass flow rate ($Q_{exit}$). While TD3 successfully recovered from these instabilities, SAC’s ability to prevent them entirely indicates a more stable policy and more effective control during the critical initial phase of back-pressure buildup.
    \item \textbf{Impact of Policy Nature on Robustness:} The difference in the performance is directly related to the fundamental difference between the algorithms. TD3 learns a deterministic policy, which tends to exploit high-reward actions very soon but can result in \textit{brittle} policies that struggle to adapt to the fast, non-linear dynamics of the shock motion, which triggers spillage. Conversely, SAC learns a stochastic policy regularized by entropy, which encourages broad exploration of the state-action space. As seen in Figure \ref{fig:reward_all_TR}, this allowed SAC to experience and learn from a broader range of flow conditions during training, resulting in a more generalized control policy that is resilient to sudden flow disturbances.
\end{itemize}
Consequently, the SAC agent is identified as the optimal choice for inference and deployment. Its extensive exploration during the training phase ensures that it has encountered a diverse range of back-pressure conditions before converging to an optimal policy. This broader experience makes SAC significantly more robust to unseen variations and less likely to fail when subjected to aggressive disturbances.

\subsection{Zero-shot generalization with and without noisy measurements}
Zero-shot (0-shot) generalization denotes the ability of a trained model to operate effectively under previously unseen conditions without any additional training or parameter adaptation. In the context of DRL, this implies that a control policy learned from a restricted set of environments can be directly deployed in new flow regimes, geometries or operating conditions while retaining satisfactory performance. For AFC, zero-shot generalization is of central importance because the generation of accurate training data; whether through direct numerical simulations, large-eddy simulations or laboratory experiments, is computationally and experimentally expensive. The ability of a DRL-based controller to generalize in a zero-shot manner across flow conditions is therefore a key requirement for real-world deployment. Strong zero-shot performance indicates that the agent has learned control strategies that exploit dominant flow physics, rather than overfitting to specific realizations of the training environment. This property is particularly desirable in AFC, where robust manipulation of nonlinear and multiscale dynamics is required across a range of operating points.
As a result, improving zero-shot generalization has become a major research direction aimed at closing the gap between numerical demonstrations of DRL-based flow control and reliable application in experimental and industrial settings.

In this section, we assess the zero-shot generalization capability of the trained flow controller. A policy learned exclusively under the TR40 condition is directly deployed, without any gradient updates or access to additional training data, to actively control flows at TR30 and TR50. Note that TR30 exhibits a lower back pressure than TR40, whereas TR50 shows a substantially higher back pressure than TR40. This evaluation probes the robustness and transferability of the learned control strategy across distinct flow regimes, thereby testing whether the controller can mitigate unstart through a generalized physical mechanism (microjets) rather than relying on condition-specific mitigation strategies. This evaluation is performed under three distinct sensing conditions: 

\begin{enumerate}
    \item \textbf{Clean sensor data:} The agent receives exact pressure readings from the high-fidelity CFD solver.

    \item \textbf{5\% noisy sensor data:} A moderate level of measurement uncertainty is introduced.

    \item \textbf{10\% noisy sensor data:} A high level of measurement uncertainty is introduced.
\end{enumerate}

The inclusion of noisy sensor data is essential for assessing the practical viability of the DRL controller, as pressure measurements are inherently subject to uncertainty. To emulate these conditions, synthetic noise is injected into the pressure signals during the inference phase. The noisy sensor signal, $p_{\mathrm{noisy}}$, is obtained from the true pressure signal, $p_{\mathrm{original}}$, computed by the high-fidelity CFD solver, as
\begin{equation}
p_{\mathrm{noisy}} = p_{\mathrm{original}} \left( 1 + \frac{\delta}{100}\,\mathcal{N}(0,1) \right),
\end{equation}
where $\delta$ denotes the noise intensity (5 or 10\%), and $\mathcal{N}(0,1)$ is a random variable drawn from a standard normal distribution with zero mean and unit variance. By systematically increasing the noise level, we assess whether the learned control policy is governed by dominant flow physics rather than overfitting to precise, noise-free numerical values.

\begin{figure}
\centering
\includegraphics[scale=0.29, clip=true]{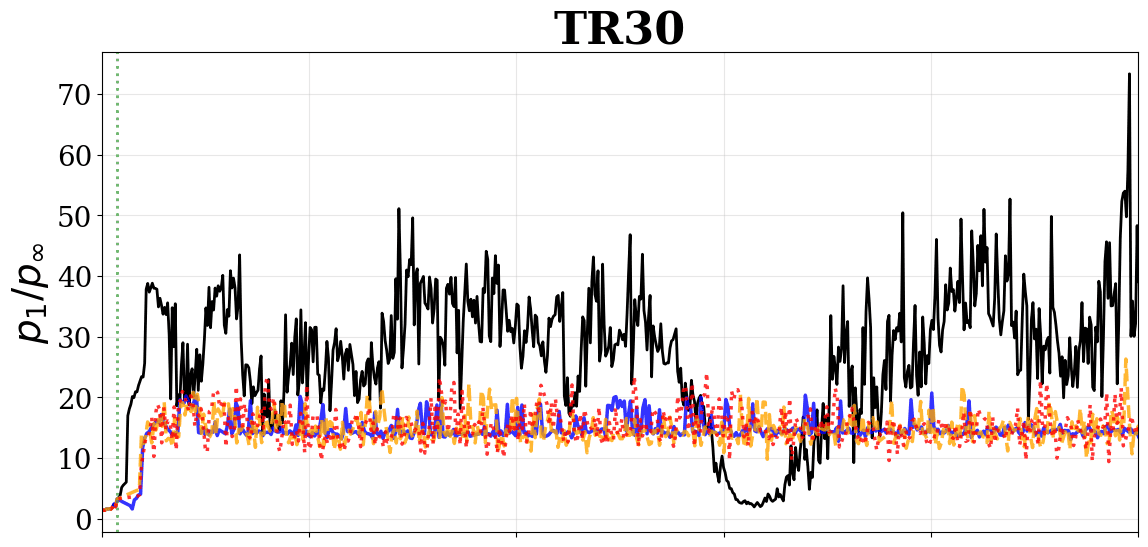}
\includegraphics[scale=0.29, clip=true]{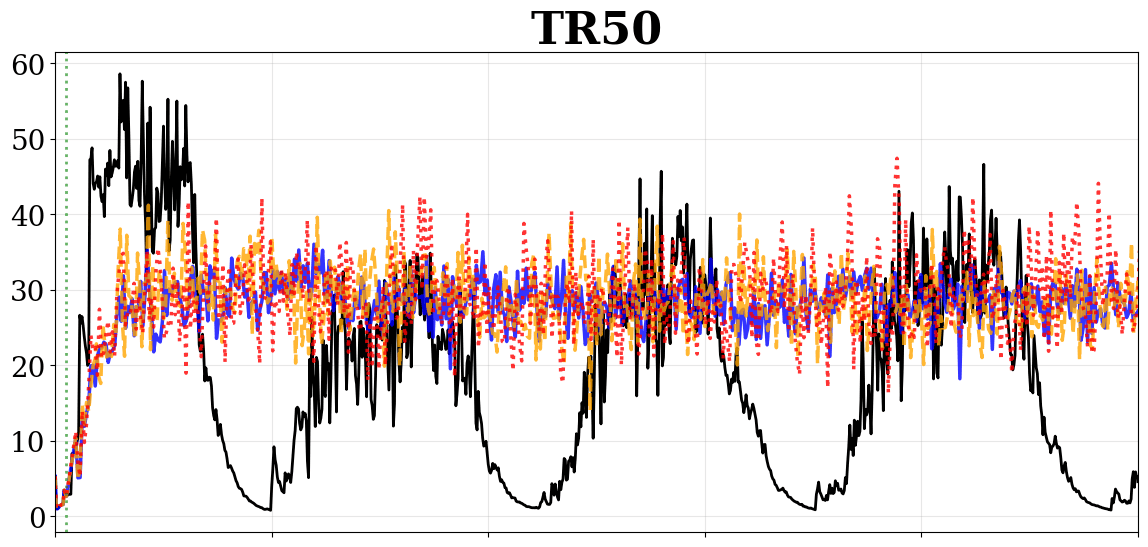}
\includegraphics[scale=0.3, clip=true]{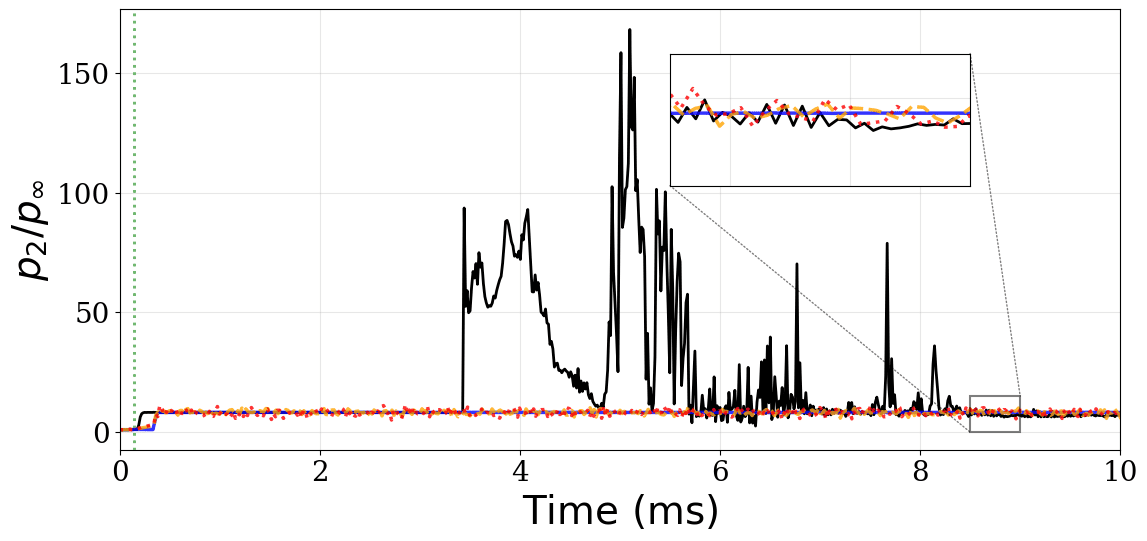}
\includegraphics[scale=0.3, clip=true]{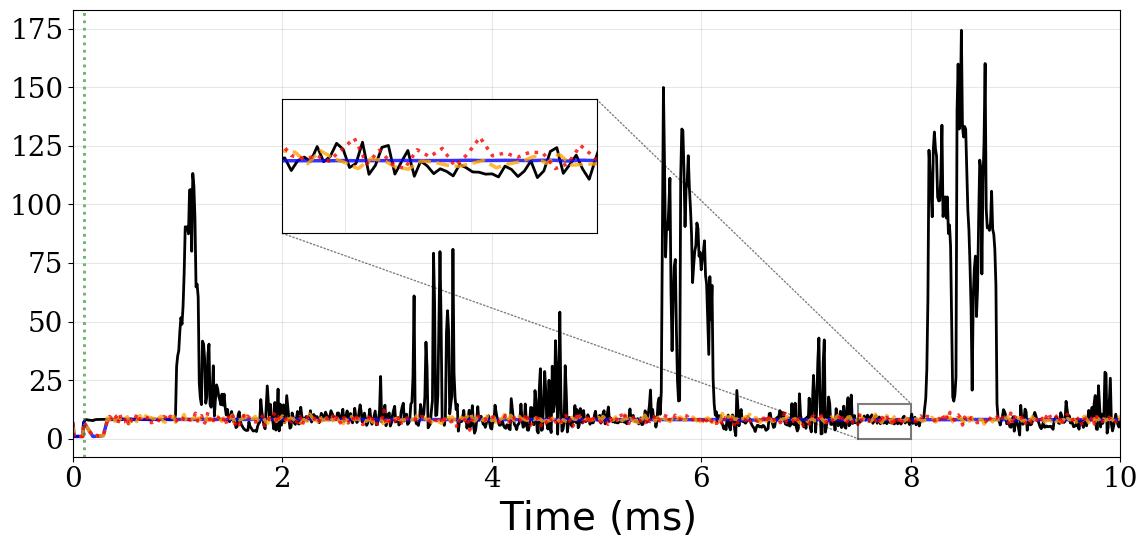}
\includegraphics[scale=0.3, clip=true]{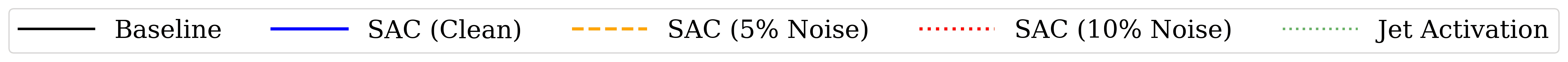}
\caption{Zero-shot pressure evolution ($p_1/p_{\infty}, p_2/p_{\infty}$) for TR30 and TR50 under varying sensor noise levels (0\%, 5\%, 10\%).}
\label{fig:p1p2_infer_0shot}
\end{figure}
\begin{figure}[htpb]
\centering
\includegraphics[scale=0.23, clip=true]{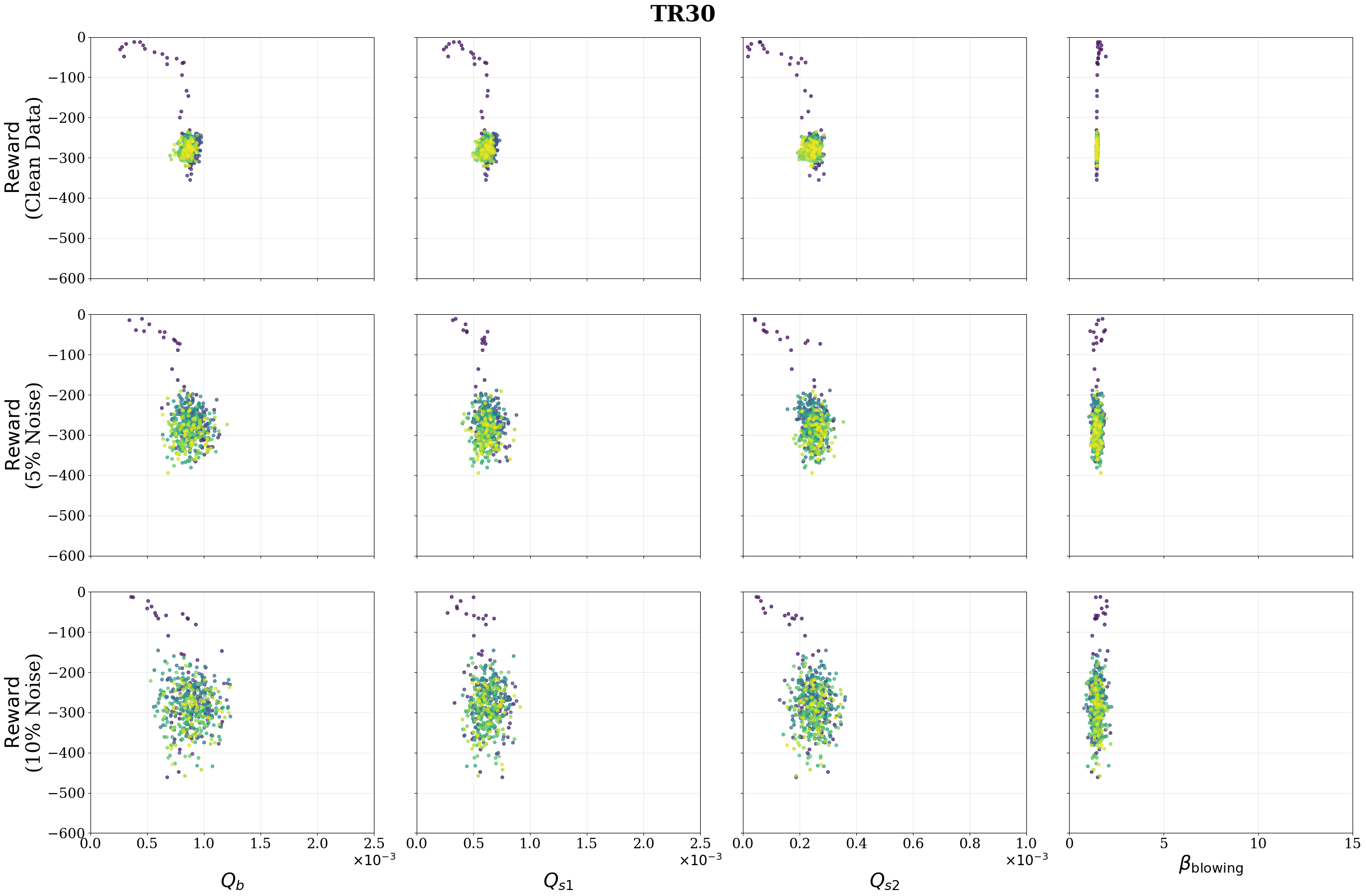}
\includegraphics[scale=0.23, clip=true]{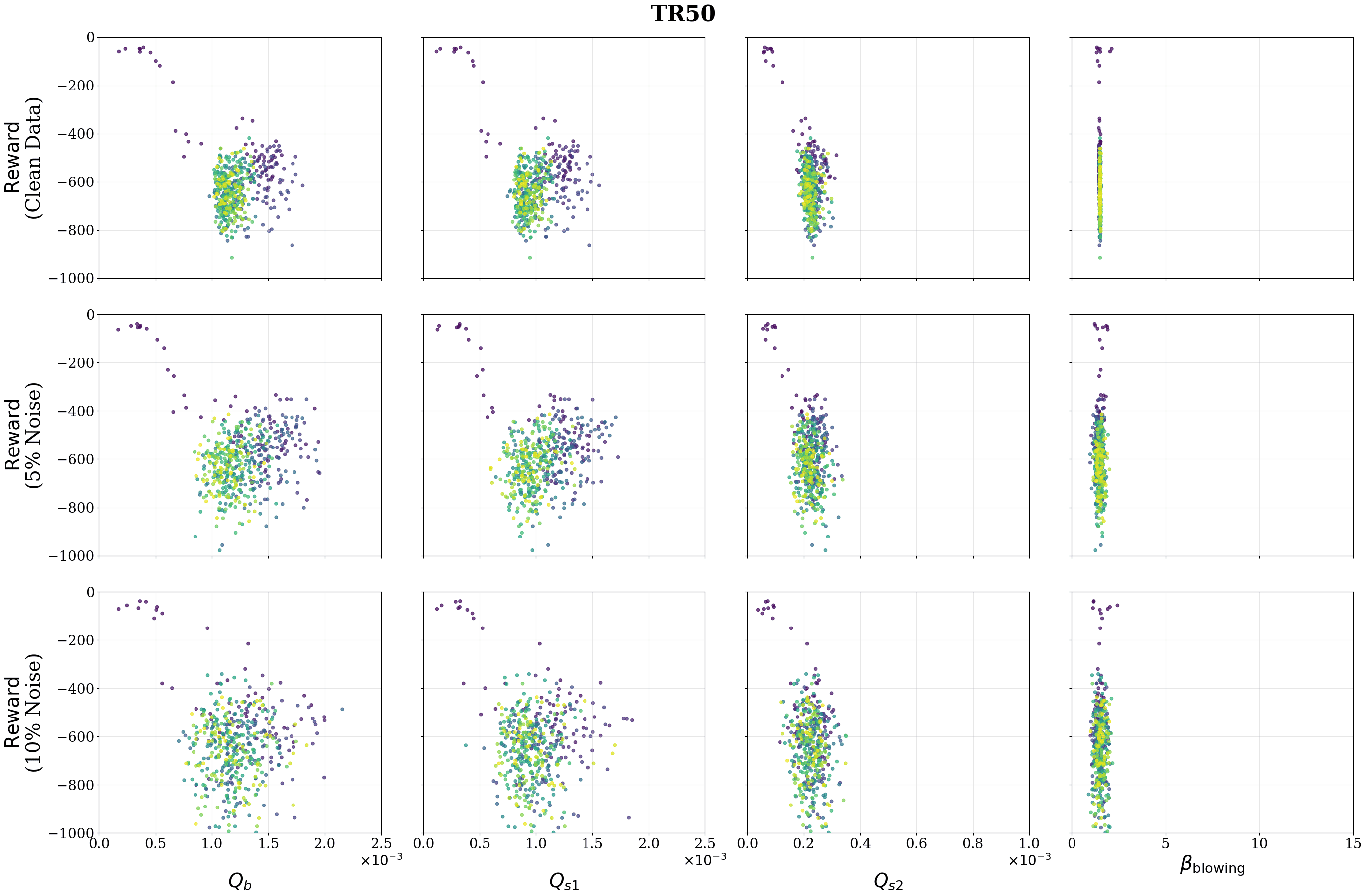}
\includegraphics[scale=0.24, clip=true]{figures/colorbar_reward.png}
\caption{Zero-shot Generalization: Distribution of instantaneous reward against control parameters: blowing mass flux ($Q_b$), suction mass fluxes ($Q_{s1}, Q_{s2}$), and blowing angle ($\beta_{\text{blowing}}$) for throttling ratio - TR30 and TR50. The top row represents the clean sensor readings, the middle row represent 5\% noise and the bottom row represents 10\% noise. The color gradient indicates the physical time.}
\label{fig:reward_infer_0shot}
\end{figure}
\begin{figure}[htpb]
\centering
\includegraphics[scale=0.25, clip=true]{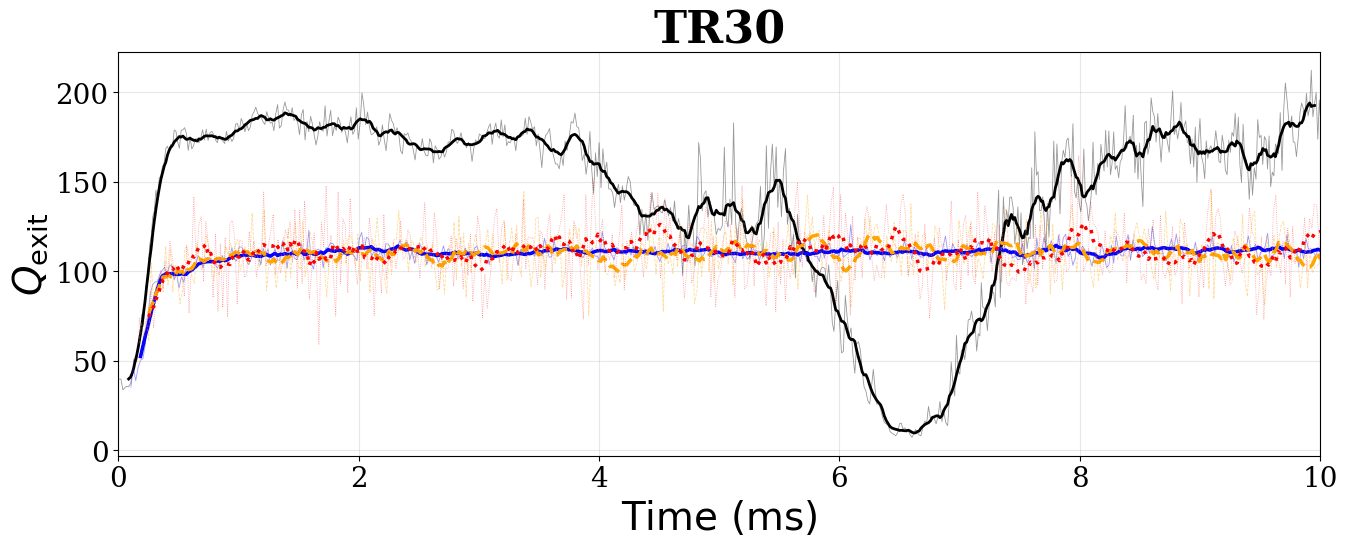}
\includegraphics[scale=0.25, clip=true]{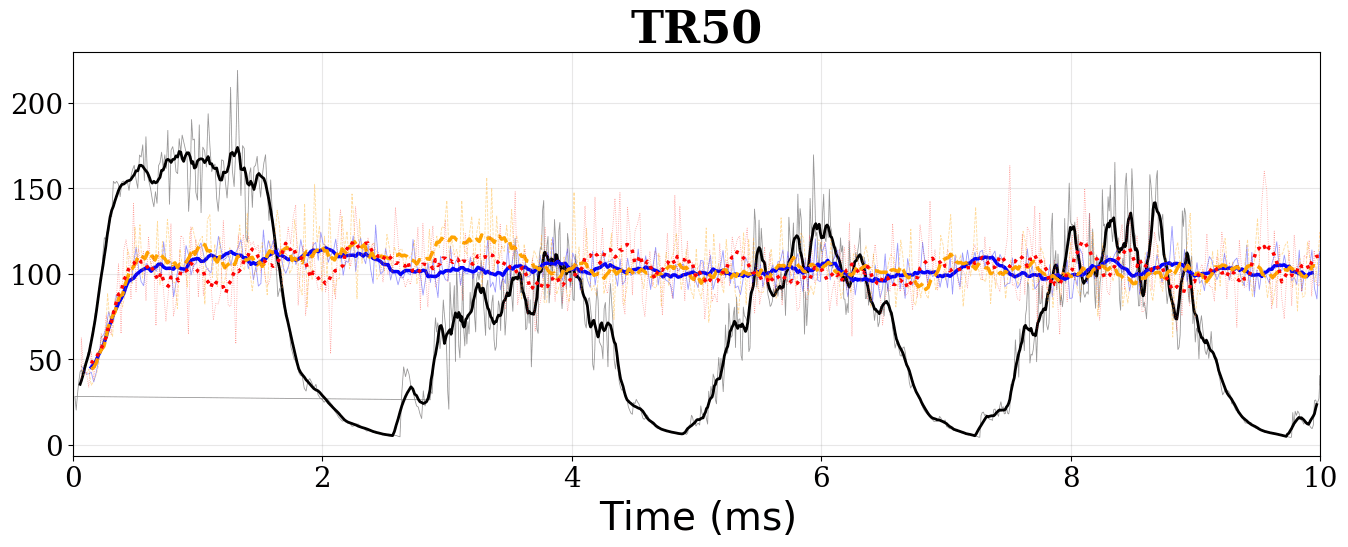}
\includegraphics[scale=0.26, clip=true]{figures/legend_inference.png}

\caption{Zero-shot Generalization: Evolution of exit mass flow rate ($Q_{exit}$) under varying sensor noise for TR30 and TR50.}
\label{fig:mass_infer_0shot}
\end{figure}
The zero-shot generalization capability of the policy, which is trained in TR40, is first evaluated under clean sensor data (0\% noise). As shown in the blue in Figure \ref{fig:p1p2_infer_0shot}, the controller successfully prevents unstart in both the lower back-pressure case (TR30) and the higher back-pressure case (TR50). In the TR50, where the uncontrolled baseline (black line) suffers an immediate and catastrophic unstart (even before 1 ms), the agent effectively anchors the shock system, maintaining the downstream pressure $p_2$ at a stable value. This successful deployment on unseen operating points confirms that the model has not merely memorized the specific actions for TR40, but has instead learned a generalized physical strategy for shock stabilization that adapts dynamically to varying back-pressure intensities.
However, the introduction of sensor noise shows the sensitivity of the control to measurement uncertainty. As the noise level increases to 5\% (orange) and 10\% (red), the pressure at $p_1$ and $p_2$ exhibits growing oscillations. This behavior is most pronounced in the TR50 case with 10\% noise, where the upstream pressure $p_1$ fluctuates violently, and even the typically stable downstream probe $p_2$ begins to show unsteadiness. Despite these significant oscillations, it is important to note that the controller does not fail; the pressure remains well below the unstart threshold defined by the baseline, which demonstrates that the agent retains sufficient control ability to prevent spillage even when the shock position cannot be held perfectly steady. The impact of this sensor noise on global flow stability is further examined by the exit mass flow rate ($Q_{exit}$), shown in Figure \ref{fig:mass_infer_0shot}. For clean sensor data (blue), the agent maintains a highly steady mass flow in both TR30 and TR50 cases. But, consistent with the pressure fluctuations observed in Figure \ref{fig:p1p2_infer_0shot}, the introduction of 5\% and 10\% noise induces large-amplitude oscillations in $Q_{exit}$. These fluctuations are particularly severe in the high-back-pressure TR50 case with 10\% noise. This degradation in stability is explained by the action-reward distributions in Figure \ref{fig:reward_infer_0shot}. With clean data (top row), the agent’s actions form tight, high-confidence clusters, indicating precise control. As measurement uncertainty is introduced (middle and bottom rows), these clusters disperse significantly, which reflects a widening of the action space. The noise forces the agent to constantly adjust its blowing and suction rates in response to stochastic sensor artifacts, which directly cause the high-frequency pressure oscillations observed in Figure \ref{fig:p1p2_infer_0shot}. Nevertheless, the persistence of instantaneous rewards across all noise levels confirms that the policy remains robust. While the 10\% noise prevents the agent from holding the shock train fixed in time, the most important objective, which is preventing unstart and maintaining internal flow, is successfully achieved. This demonstrates the controller’s robustness to previously unseen and noisy conditions.

\subsection{Optimal Locations for Pressure Sensors}
The objective of this analysis is to identify the most important pressure sensor locations within the isolator to accurately represent the flow field using a minimal number of measurements. The dataset used for optimal sensor estimation consists of pressure readings from previously used 100 sensor locations collected from TR40’s pressure data, where there is no control. 
\begin{figure}[htpb]
\centering
\includegraphics[scale=0.5, clip=true]{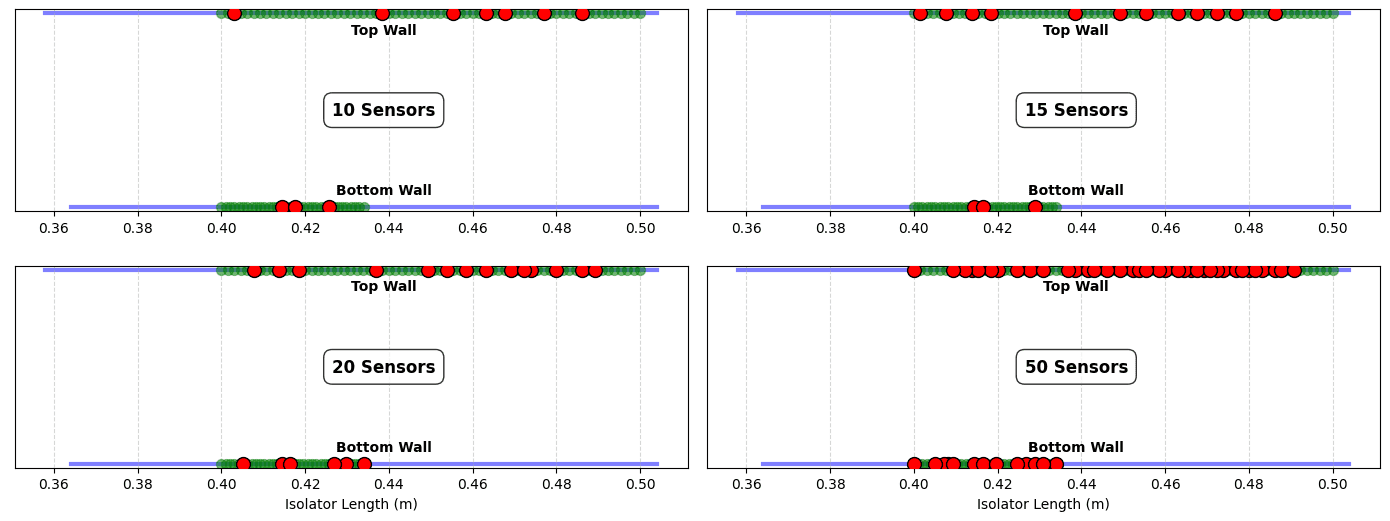}
\caption{Optimal sensor locations (red circles) for different numbers of sensors, selected from all 100 sensor points that have been used previously (green circles). The sensors are positioned along the isolator top and bottom walls (shown in blue color line) for varying optimal sensor counts (10, 15, 20, and 50).}
\label{fig:optimal_sensors}
\end{figure}

The optimal sensor locations are determined using a data-driven approach based on Singular Value Decomposition (SVD) followed by QR factorization with column pivoting \cite{jagtap2022deepL}. First, a data matrix $\mathbf{X}$ is constructed where rows represent the 100 spatial points and columns represent time snapshots of the pressure history. Subsequently, the data matrix is decomposed using SVD to extract the dominant spatial features, or modes, of the system, represented by the equation:
\begin{equation}
\mathbf{X} = \mathbf{U}\mathbf{\Sigma}\mathbf{V}^T
\end{equation}
Here, $\mathbf{U}$ contains the eigenbasis modes, which represent the dominant spatial modes of the system associated with the largest singular values and thus capture the majority of the flow’s energy. To determine which sensor locations best resolve these dominant modes, the first $r$ columns of $\mathbf{U}$, denoted as $\mathbf{\Psi}_r$, are extracted. A QR factorization with column pivoting is then performed on the transpose of these modes:
\begin{equation}
\mathbf{\Psi}_r^T \mathbf{P}^T = \mathbf{Q}\mathbf{R}
\end{equation}
This operation yields a permutation vector $\mathbf{P}$, which reorders the sensor indices based on their linear independence and contribution to the reconstruction. The first $r$ indices in the permutation vector $\mathbf{P}$ correspond to the optimal sensor locations that minimize reconstruction error and maximize the determinant of the observation matrix.  The resulting optimal configurations are visualized in Figure \ref{fig:optimal_sensors}. The figure shows the spatial distribution of the optimal sensors (red) relative to the full set of 100 spatial points (green) along the top and bottom walls of the isolator (shown by the blue color line) for varying sensor counts ($r = 10, 15, 20, \text{and } 50$). Note that on the bottom wall, the original sensor locations lie upstream of the suction jets, which limits the available space for additional sensors. As a result, more sensor placement freedom is available on the top wall. Also, the optimal sensor locations are observed to coalesce near the high-pressure region close to the isolator exit, where the dominant flow dynamics occur.

\subsubsection{Zero-shot generalization under noisy measurements with optimal sensors}
Following the previous analysis, the configuration with 15 optimal sensors is selected to train a SAC agent for the TR40 case. This specific selection was made to evaluate whether a sparse set of 15 discrete pressure measurements provides a sufficiently rich state representation to capture the essential flow dynamics. By successfully training the agent with this reduced input, we aim to demonstrate that 15 sensors are adequate for representing the state of the isolator for effective control. To validate the feasibility of the optimal sensor placement strategy, a new SAC agent was trained for the TR40 condition using only the 15 optimal pressure sensors as state input. It is important to note that this reduction in sensor count necessitates a modification to the reward interpretation. Since the reward function computes the norm of the pressure error vector, reducing the vector length from 100 to 15 automatically lowers the absolute magnitude of the computed reward for an equivalent physical state. Therefore, the negative reward values observed here, approaching the maximum of zero, partly reflect a scaling artifact of the sparse formulation, but also indicate the increased difficulty of controlling flow under limited observability. 

\begin{figure}[htpb]
\centering
\includegraphics[scale=0.30, clip=true]{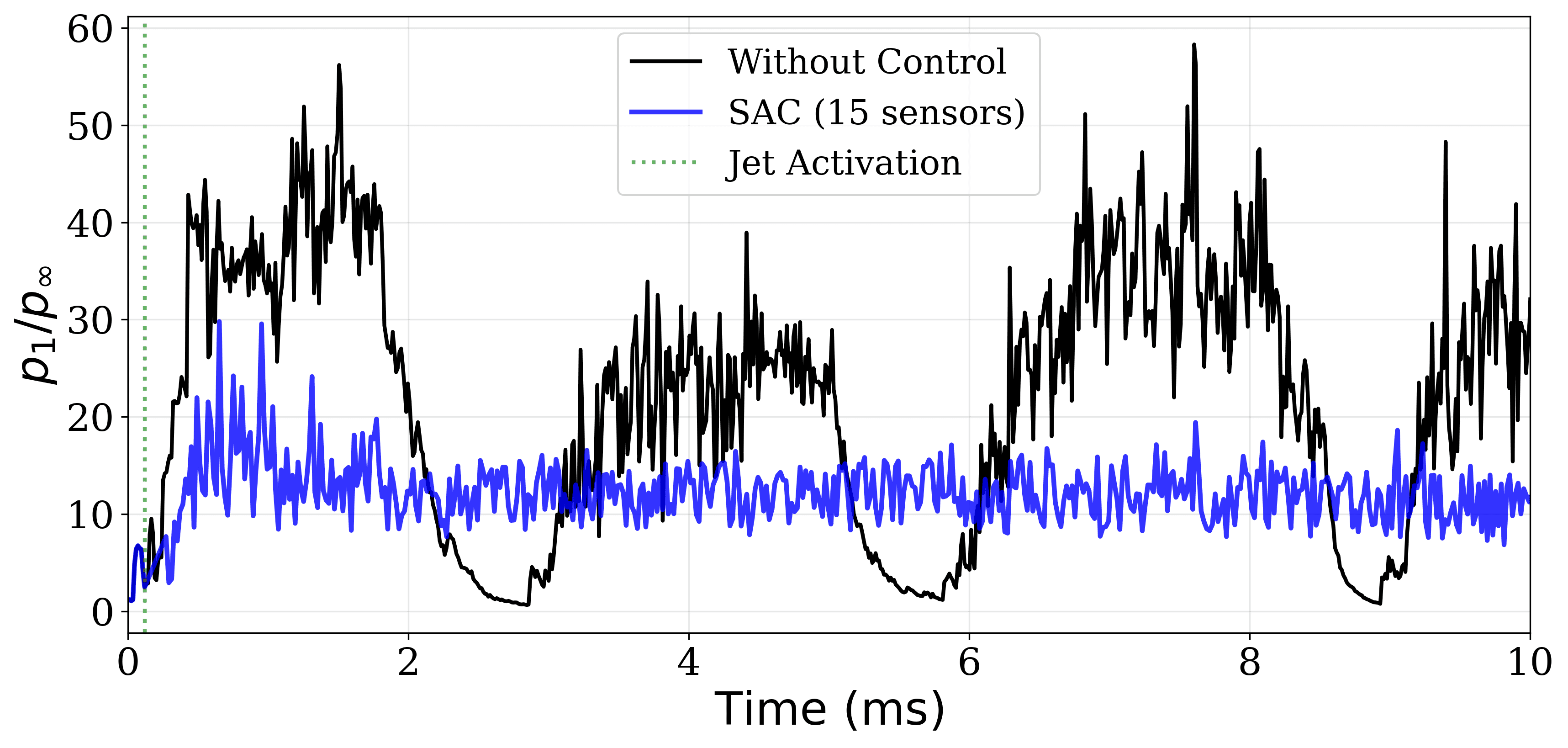}
\includegraphics[scale=0.30, clip=true]{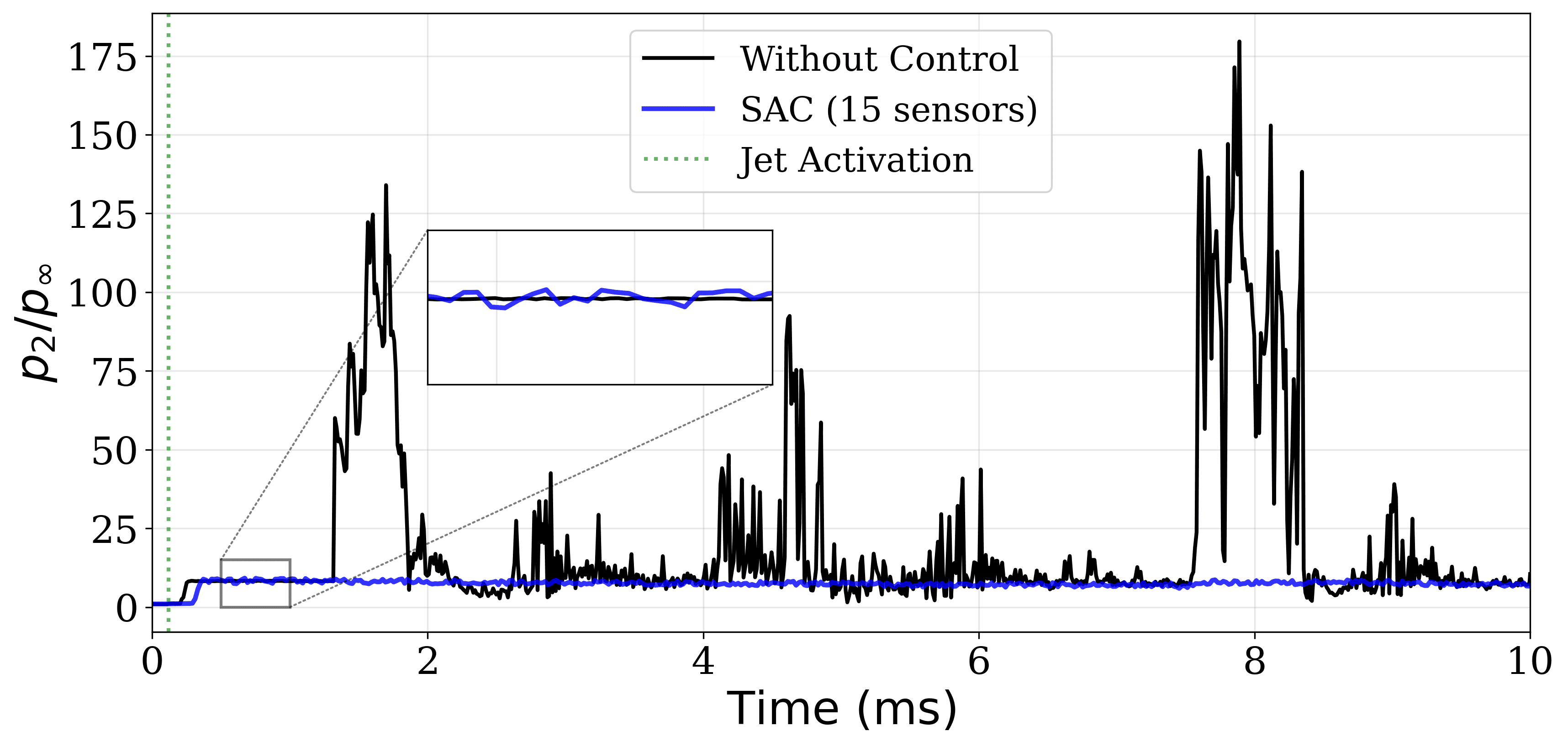}
\caption{Time evolution of normalized pressure ($p/p_{\infty}$) at monitoring points 1 (left) and 2 (right) for throttling ratio TR40 with 15 optimal sensors.}
\label{fig:15sensor_p1-p2}
\end{figure}

\begin{figure}[htpb]
\centering
\includegraphics[scale=0.26, clip=true]{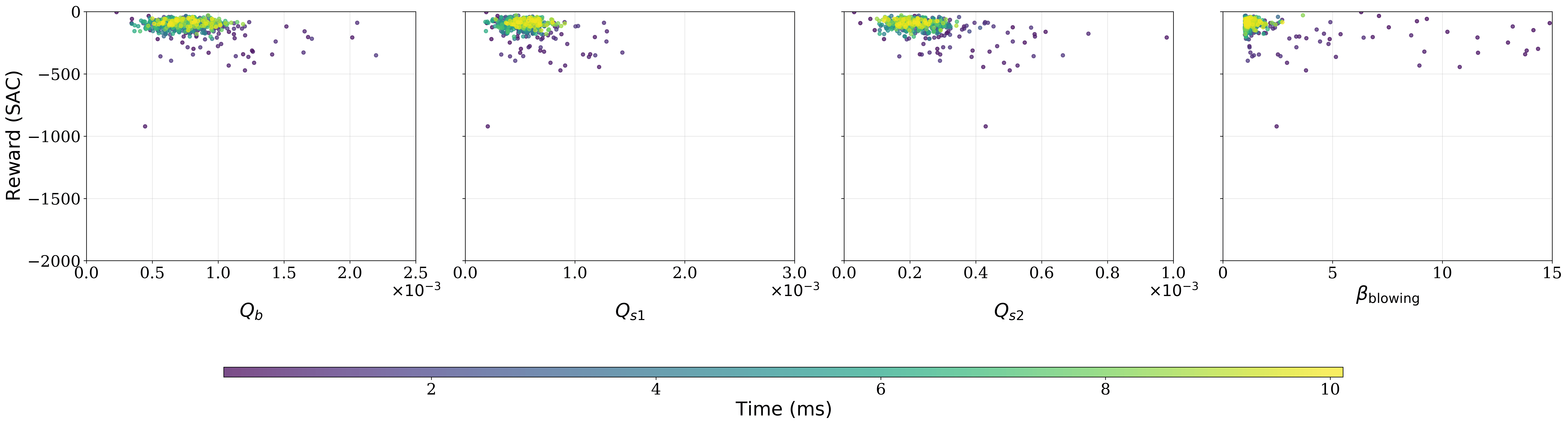}
\caption{Distribution of instantaneous reward during training against control parameters: blowing mass flux ($Q_b$), suction mass fluxes ($Q_{s1}, Q_{s2}$), and blowing angle ($\beta_{\text{blowing}}$) for throttling ratio TR40 with 15 optimal sensors.}
\label{fig:15sensor_reward}
\end{figure}

\begin{figure}[htpb]
\centering
\includegraphics[scale=0.30, clip=true]{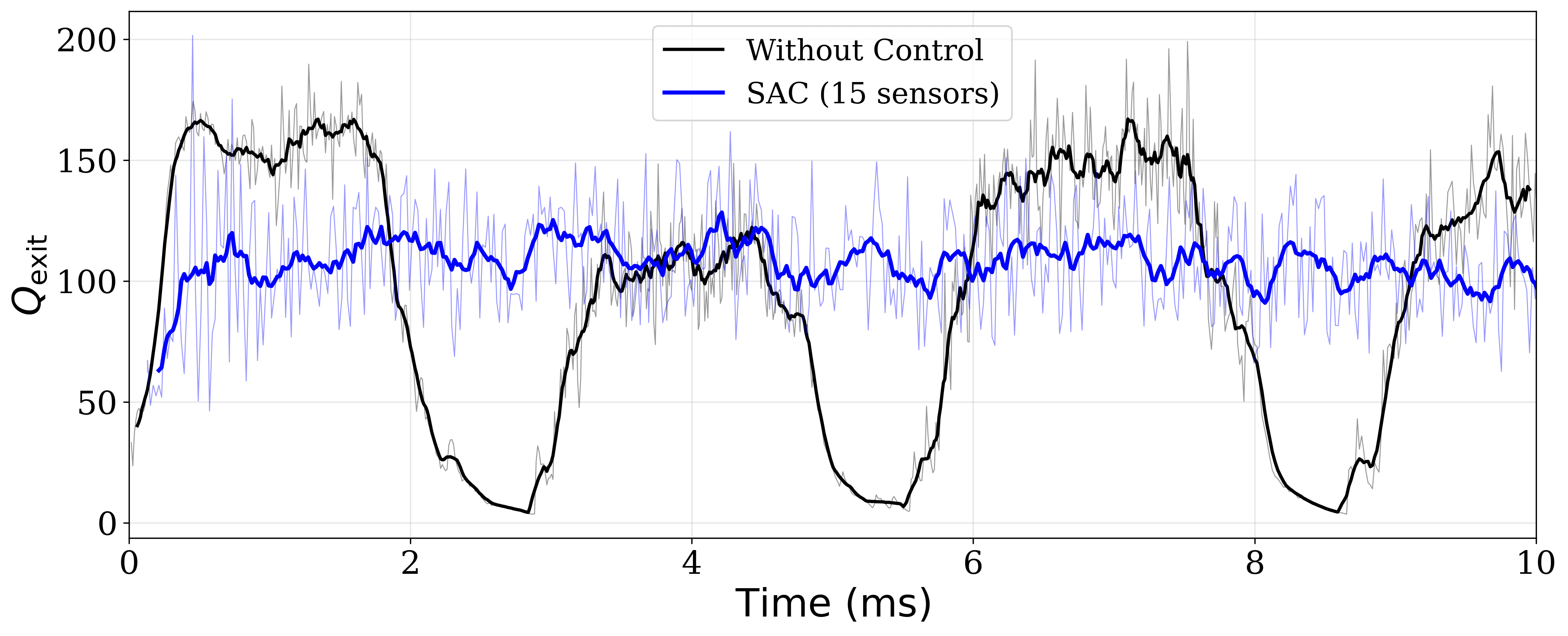}
\caption{Evolution of the mass flow rate at the isolator exit ($Q_{\text{exit}}$) for throttling ratio TR40 with 15 optimal sensors.}
\label{fig:15sensor_massflow}
\end{figure}

The control performance of this sparse-sensing agent is shown in Figure \ref{fig:15sensor_p1-p2}. The pressure at $p_2$ confirms that the agent successfully prevents the unstart observed in the baseline (black). However, the pressure at $p_1$ exhibits significantly high-frequency oscillations compared to the 100-sensor cases. This unsteadiness suggests that while 15 sensors are sufficient to detect the global shock position, the reduced spatial resolution can not capture the fine-scale shock train pattern. This behavior is validated by the action-space distribution shown in Figure \ref{fig:15sensor_reward}. Unlike the convergence observed with 100 sensors, the policy for the 15-sensor case converges to a wider cluster of actions. This increased variance indicates that the reduced state observation leads to higher uncertainty in the optimal control policy, causing the agent to constantly oscillate its actuation parameter values ($Q_b, Q_{s1}, Q_{s2}$) to maintain stability. These continuous, rapid fluctuations of the microjets directly cause the pressure oscillations observed at $p_1$. Despite these oscillations, the primary objective is established, which is preventing unstart. The exit mass flow rates without and with SAC (15 sensors) are plotted in Figure \ref{fig:15sensor_massflow}. While the exit mass flow $Q_{\mathrm{exit}}$ exhibits significant fluctuations, similar to the $p_1$ pressure dynamics, the SAC maintains the exit mass flow. In contrast, the baseline case suffers a substantial loss of mass capture due to unstart. These results confirm that 15 optimally placed sensors provide a sufficiently rich state representation to prevent unstart.

\begin{figure}[htpb]
\centering
\includegraphics[scale=0.3, clip=true]{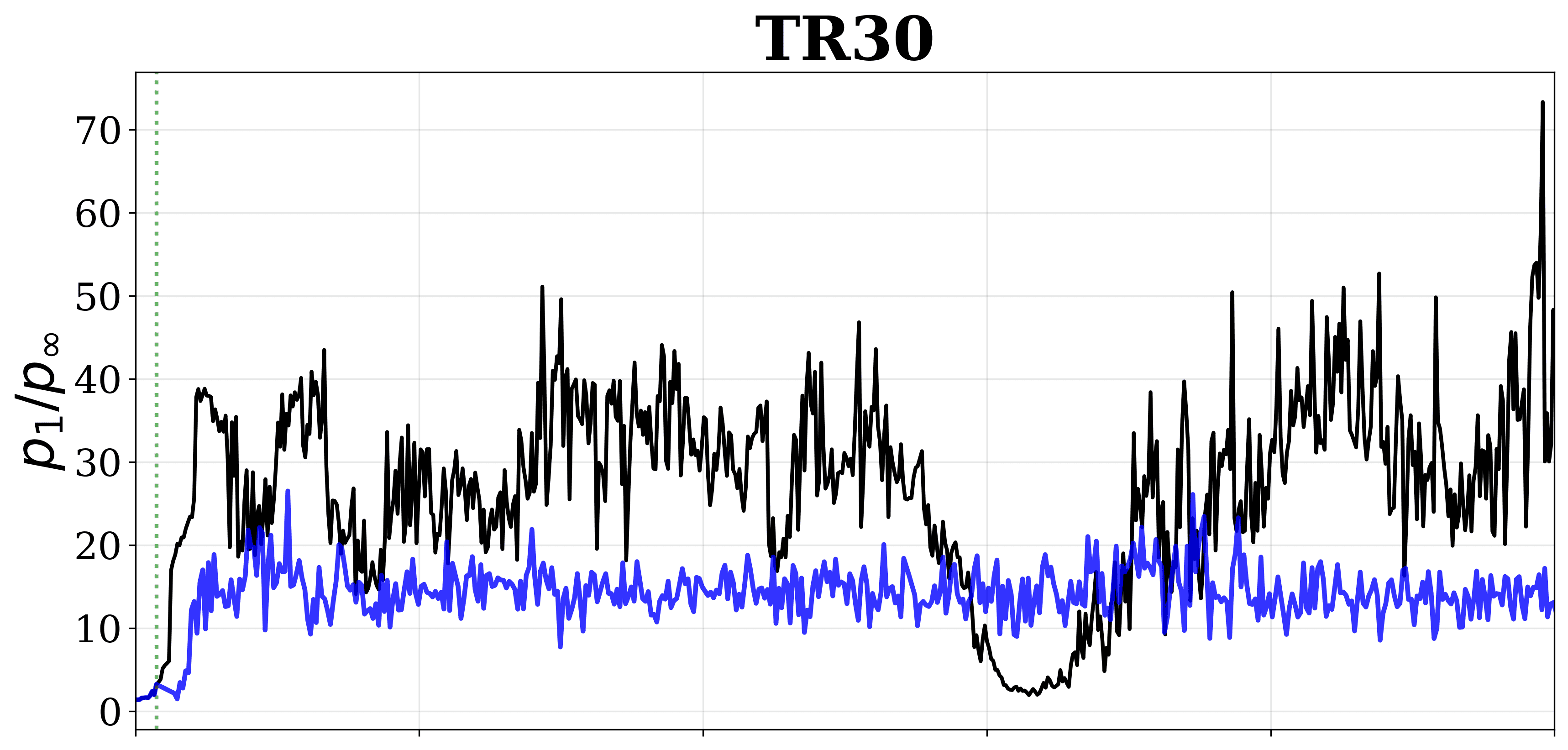}
\includegraphics[scale=0.3, clip=true]{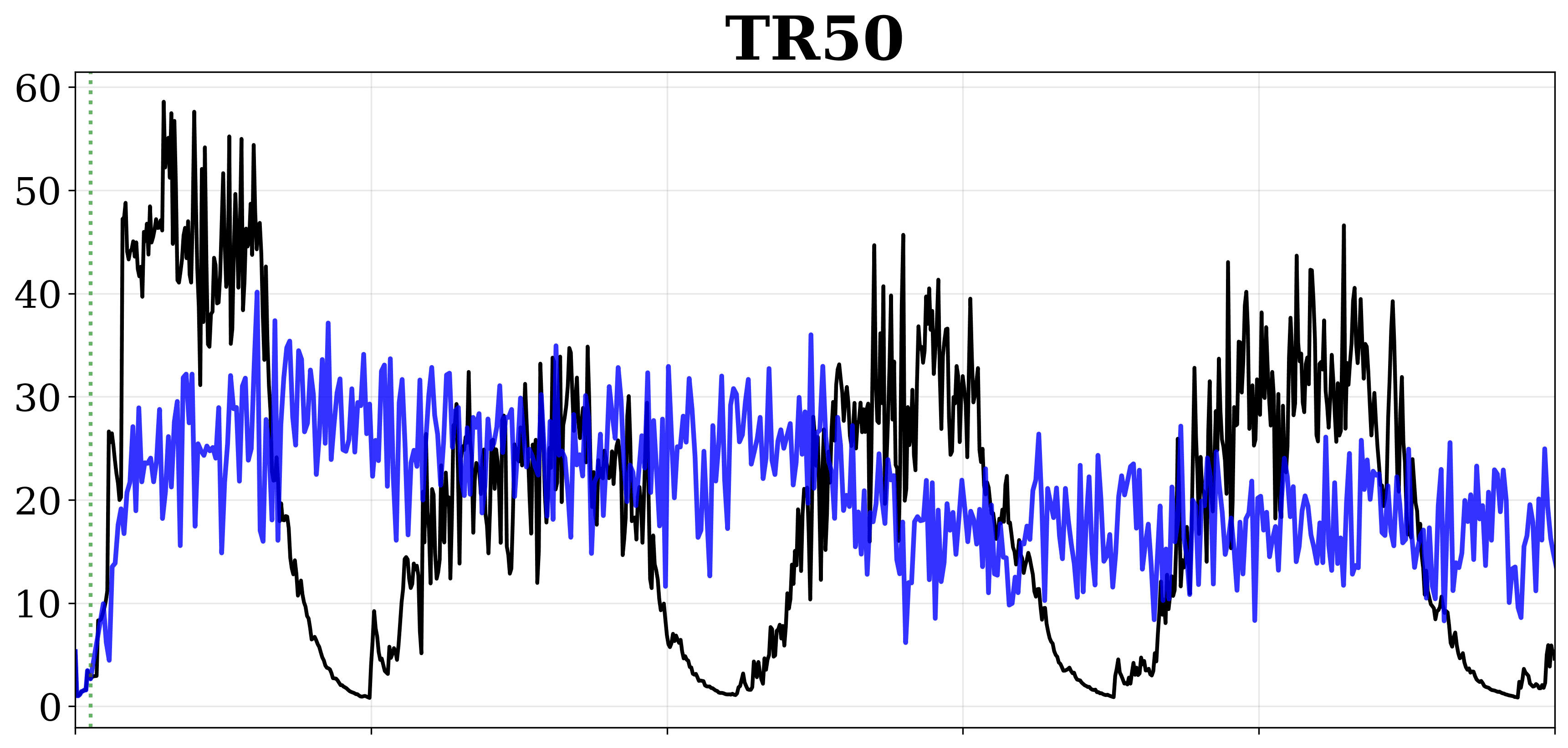}
\includegraphics[scale=0.3, clip=true]{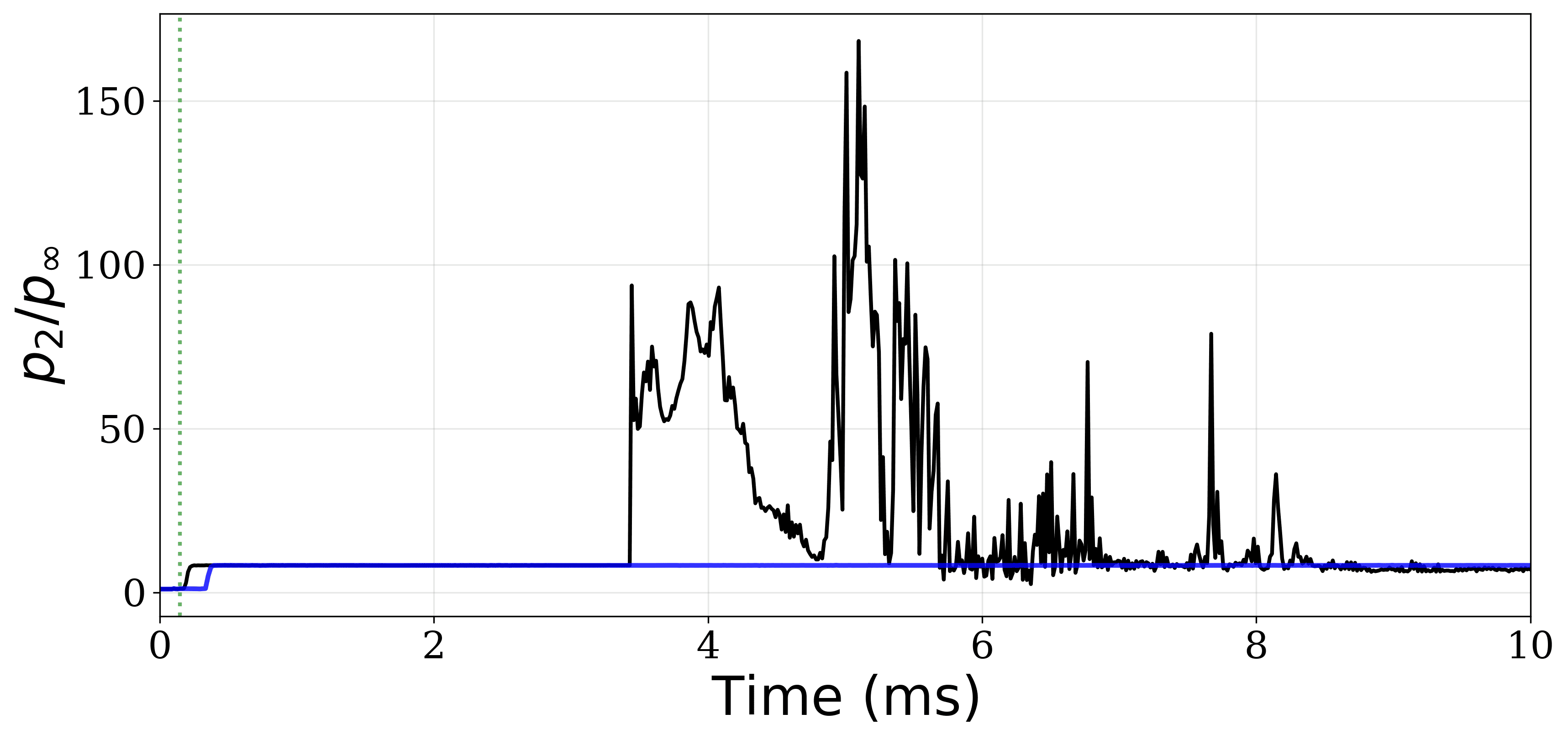}
\includegraphics[scale=0.3, clip=true]{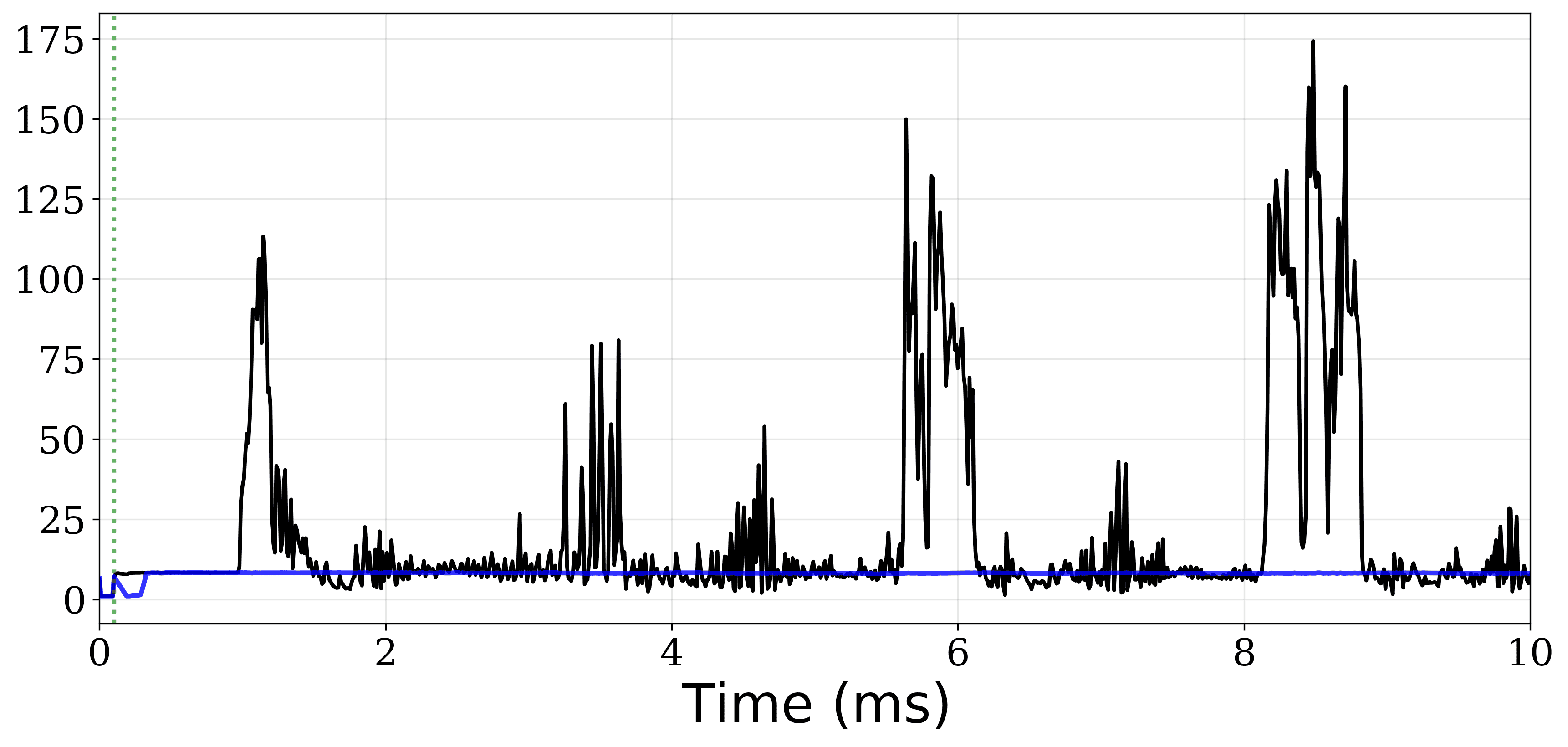}
\includegraphics[scale=0.3, clip=true]{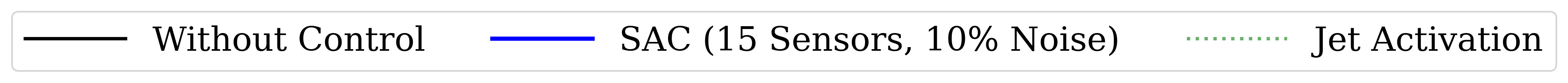}
\caption{Zero-shot Generalization: Pressure evolution ($p_1/p_{\infty}, p_2/p_{\infty}$) for TR30 and TR50 with 15 optimal sensors under 10\% noise level.}
\label{fig:p1p2_infer_0shot_15sensors}
\end{figure}

To this end, the robustness of the sparse-sensing policy is evaluated under the most challenging conditions: zero-shot deployment to TR30 and TR50 environments subject to 10\% sensor noise. This test assesses whether the reduced 15-sensor state representation retains sufficient information to generalize across operating points even when corrupted by severe measurement uncertainty. The results of this rigorous evaluation are presented in Figure \ref{fig:p1p2_infer_0shot_15sensors}, which displays the pressure evolution for TR30 and TR50 under 10\% noise. In the TR50 case, the upstream pressure $p_1$ shows significant oscillations (blue), similar to the behavior observed in the 100-sensor configuration (Figure \ref{fig:p1p2_infer_0shot}). These fluctuations confirm that the combination of high back-pressure and severe measurement noise forces the agent into a highly reactive control loop. However, the critical indicator of success is the downstream pressure $p_2$. Despite the intensive upstream backpressure, the $p_2$ remains flat and bounded near the baseline level, and there are no signs of the sharp pressure discontinuity or unstart. This confirms that even with a sparse, noisy state representation, the agent can push back the shock train within the isolator and prevent the unstart.
\begin{figure}
\centering
\includegraphics[scale=0.23, clip=true]{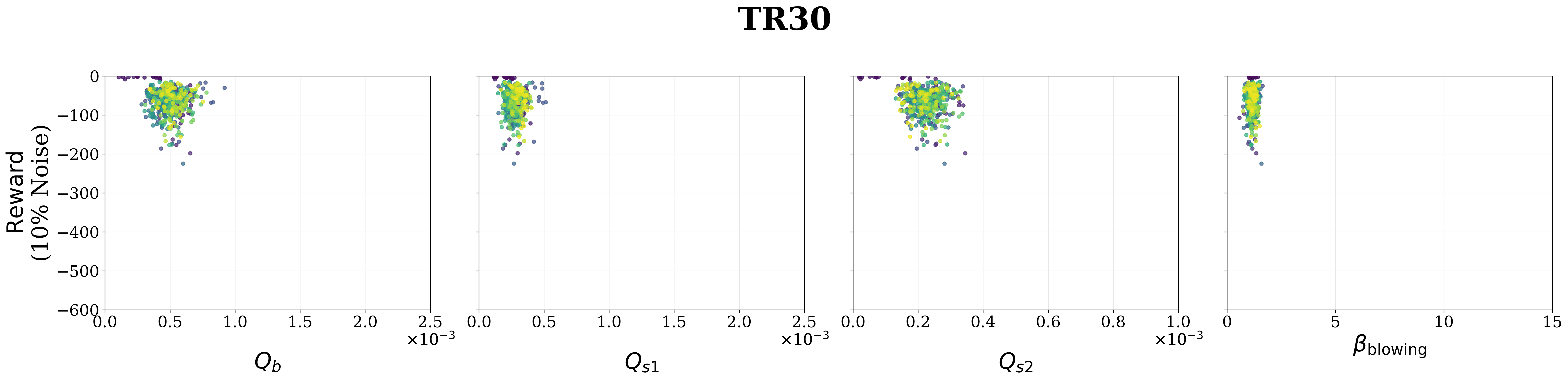}
\includegraphics[scale=0.23, clip=true]{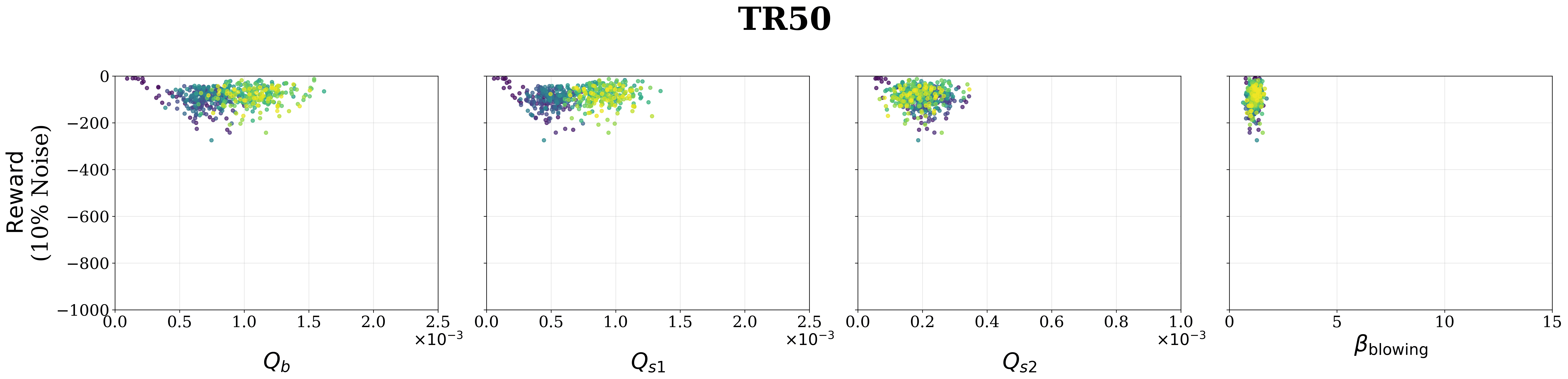}
\includegraphics[scale=0.24, clip=true]{figures/colorbar_reward.png}
\caption{Zero-shot Generalization: Distribution of instantaneous reward against control parameters: blowing mass flux ($Q_b$), suction mass fluxes ($Q_{s1}, Q_{s2}$), and blowing angle ($\beta_{\text{blowing}}$) for throttling ratio - TR30 and TR50 with 15 optimal sensors under 10\% noise level. The color gradient indicates the physical time.}
\label{fig:reward_infer_0shot_15sensors}
\end{figure}
\begin{figure}[htpb]
\centering
\includegraphics[scale=0.25, clip=true]{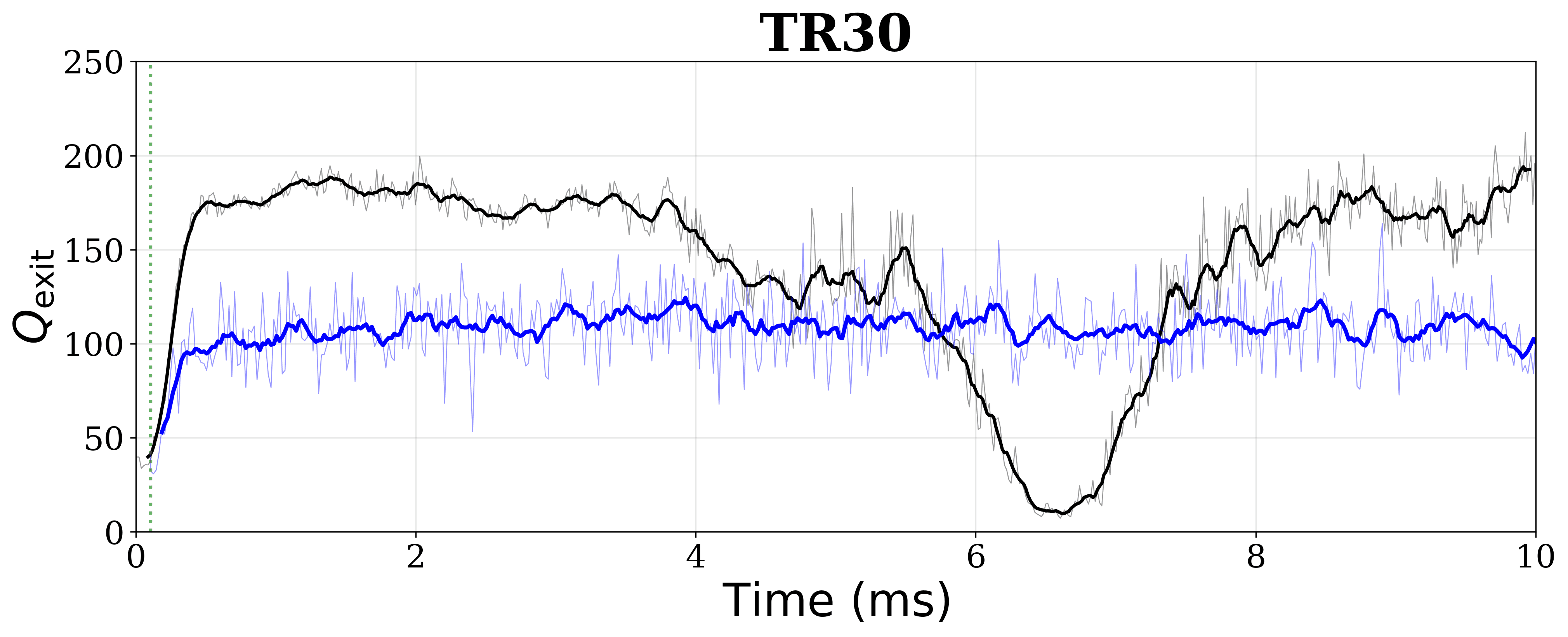}
\includegraphics[scale=0.25, clip=true]{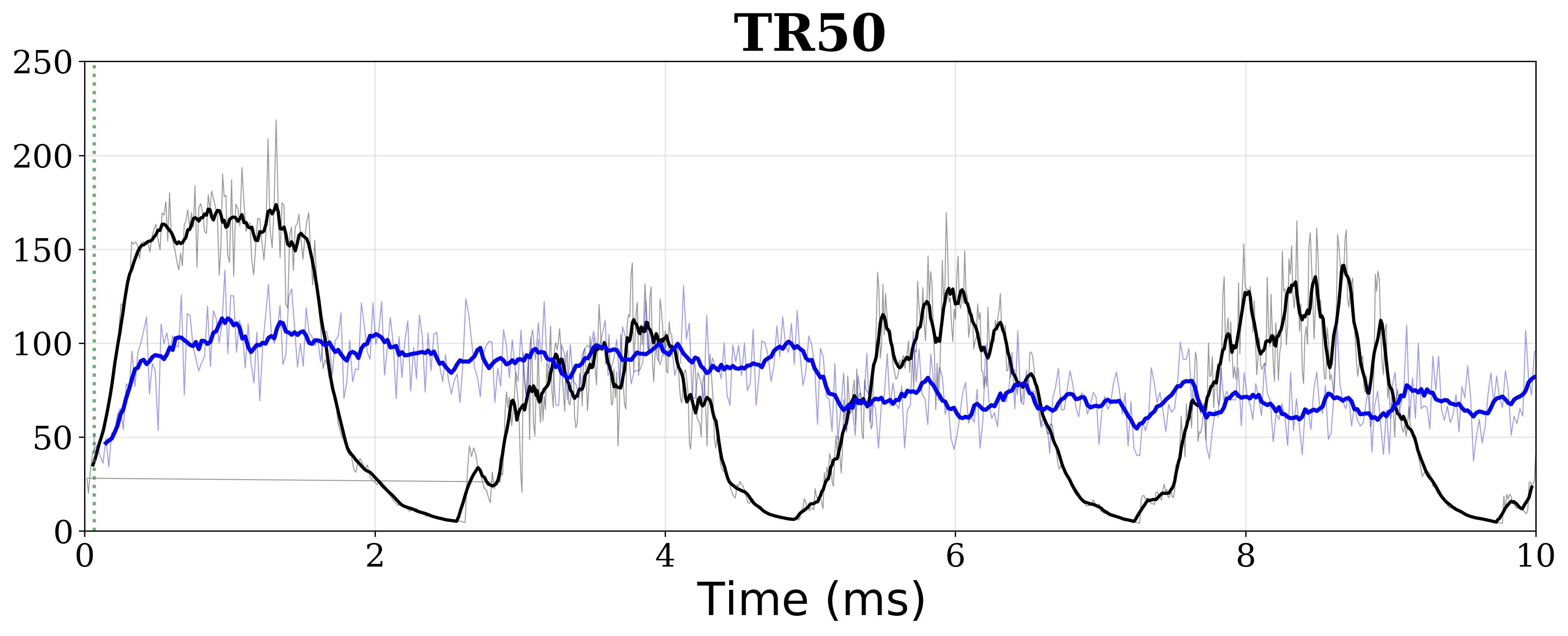}
\includegraphics[scale=0.26, clip=true]{figures/legend_inference_15sensors.png}
\caption{Zero-shot Generalization: Evolution of exit mass flow rate ($Q_{exit}$) with 15 optimal sensors under 10\% noise level for TR30 and TR50.}
\label{fig:mass_infer_0shot_15sensors}
\end{figure}
The action-reward distribution in Figure \ref{fig:reward_infer_0shot_15sensors} explains the underlying mechanism for this behavior. Interestingly, the action clusters appear less sparse and more concentrated than those observed in the 100-sensor case with equivalent noise (Figure \ref{fig:reward_infer_0shot}). The variance is reduced because stochasticity is confined to fewer input channels, so injecting noise into only 15 channels rather than 100 introduces less overall noise into the network and yields more consistent action selection. For the lower back-pressure case (TR30), the agent maintains lower jet mass fluxes. In contrast, for the high back-pressure TR50 case, the high-reward clusters shift towards higher mass flow rates ($Q_b, Q_{s1}$). This indicates that the policy trained in TR40 condition correctly identifies the stronger upstream shock propagation inherent to TR50 and increases the actuations to counter it. The exit mass flow rate ($Q_{exit}$; Fig. \ref{fig:mass_infer_0shot_15sensors}) follows the same trend as the action space and, because the actions are less scattered, shows substantially reduced oscillations compared with the 100-sensor case. Appendix \ref{appF} presents Mach-number and Q-criterion plots at different time snapshots.

\subsection{Zero-Shot Generalization to Unseen Reynolds Numbers}
Zero-shot generalization to unseen Reynolds numbers is important for hypersonic intakes because operational conditions span a wide and continuously varying range, and robust control must be maintained despite changes in flight speed, altitude, and atmospheric density that alter shock–boundary-layer interactions, flow separation, and mass capture without requiring retraining. In this section, we demonstrate the zero-shot generalization of the controller trained on the TR40 case at a Reynolds number of $5 \times 10^6$ to two previously unseen Reynolds numbers, $10 \times 10^6$ and $15 \times 10^6$, for the TR50 configuration with 10\% noise, highlighting its ability to maintain stable and consistent control performance without additional training.
\begin{figure}[htpb]
\centering
\includegraphics[scale=0.3, clip=true]{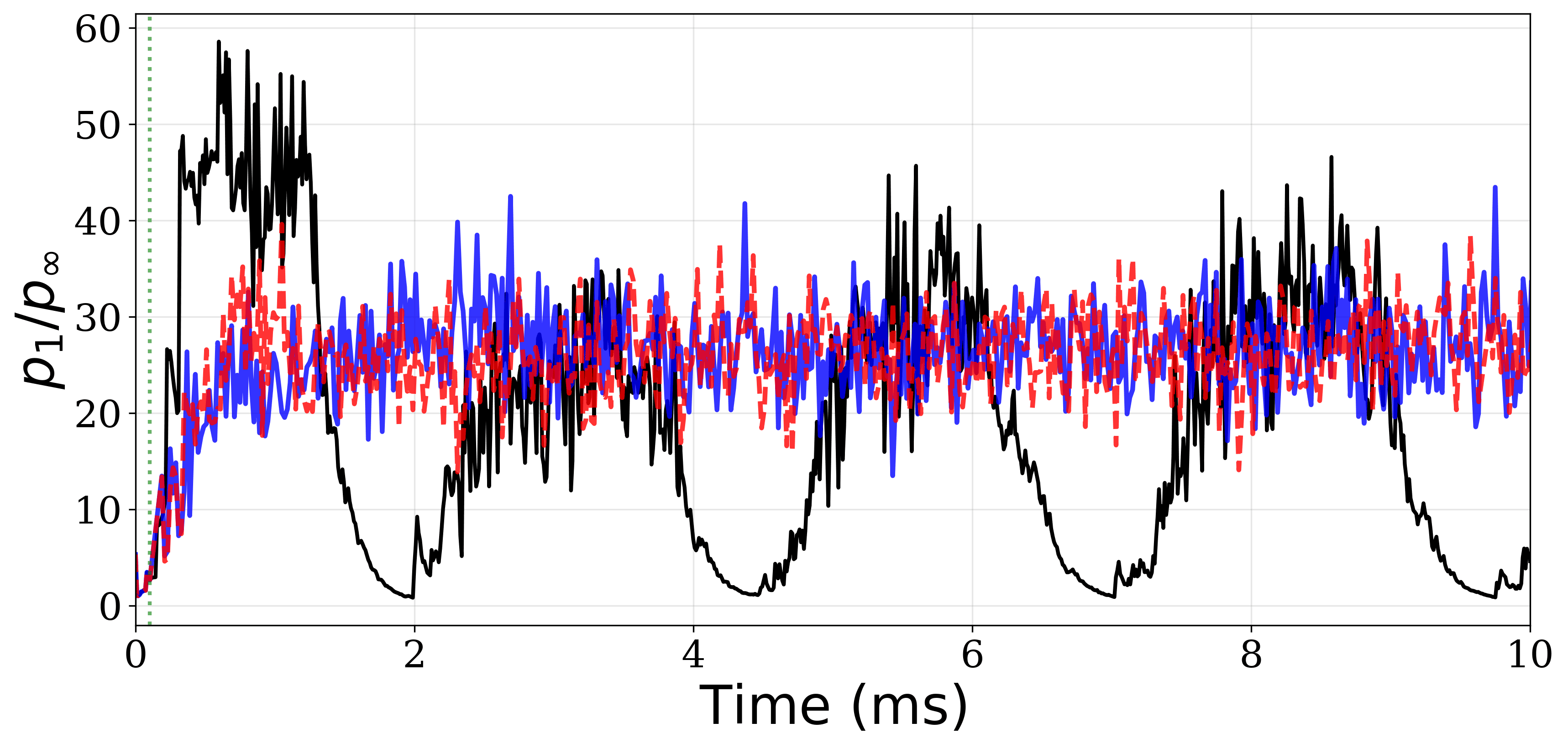}
\includegraphics[scale=0.3, clip=true]{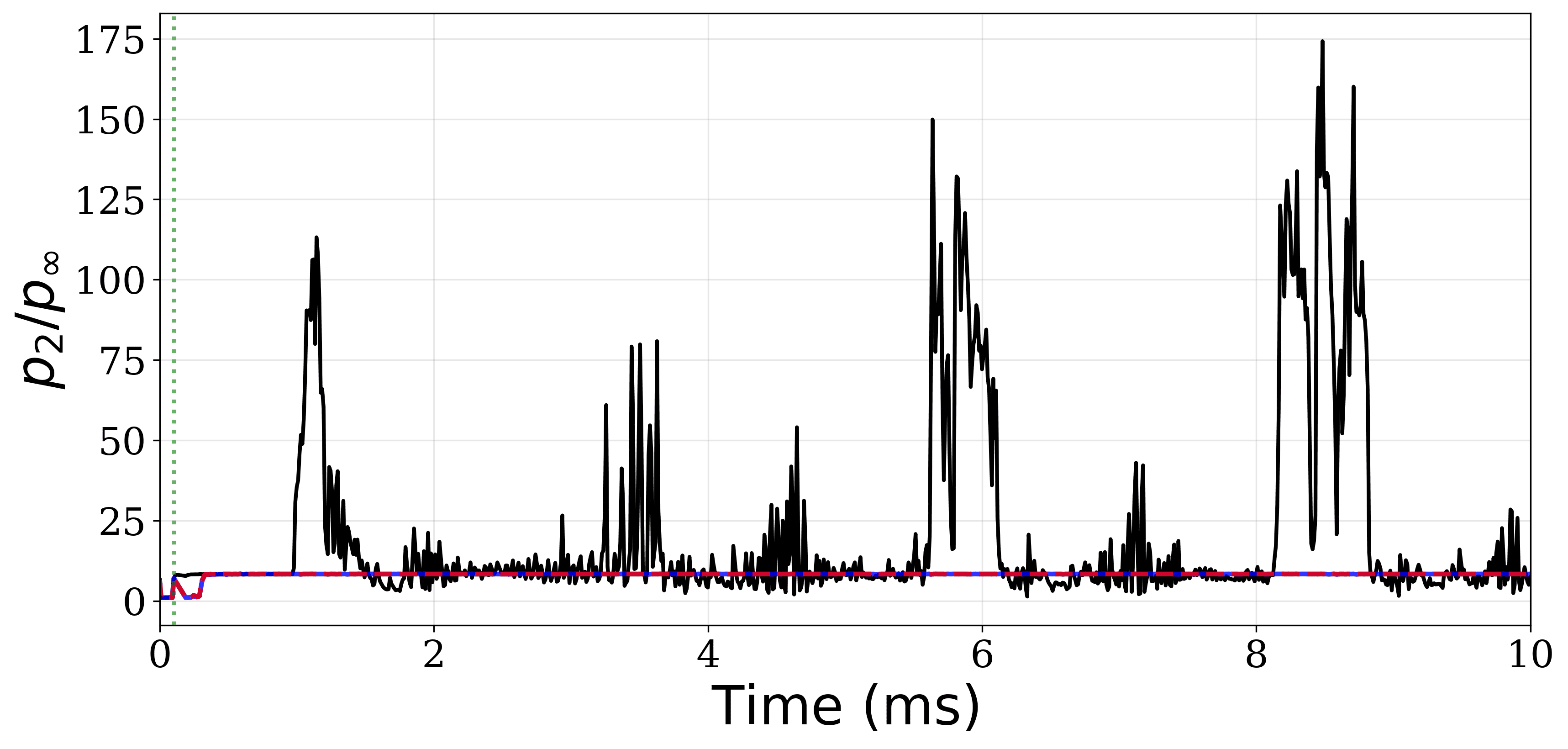}
\includegraphics[scale=0.3, clip=true]{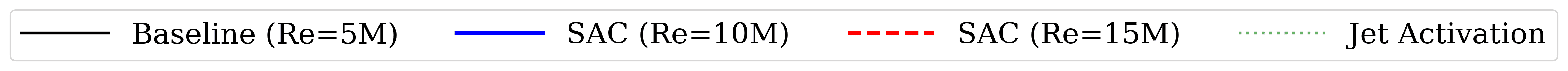}
\caption{Zero-shot Generalization: Pressure evolution ($p_1/p_{\infty}, p_2/p_{\infty}$) for $Re=10\times10^{6}$ and $Re=15\times10^{6}$ with 15 optimal sensors under 10\% noise level in TR50.}
\label{fig:Re_p1p2_infer_0shot_15sensors}
\end{figure}

\begin{figure}
\centering
\includegraphics[scale=0.23, clip=true]{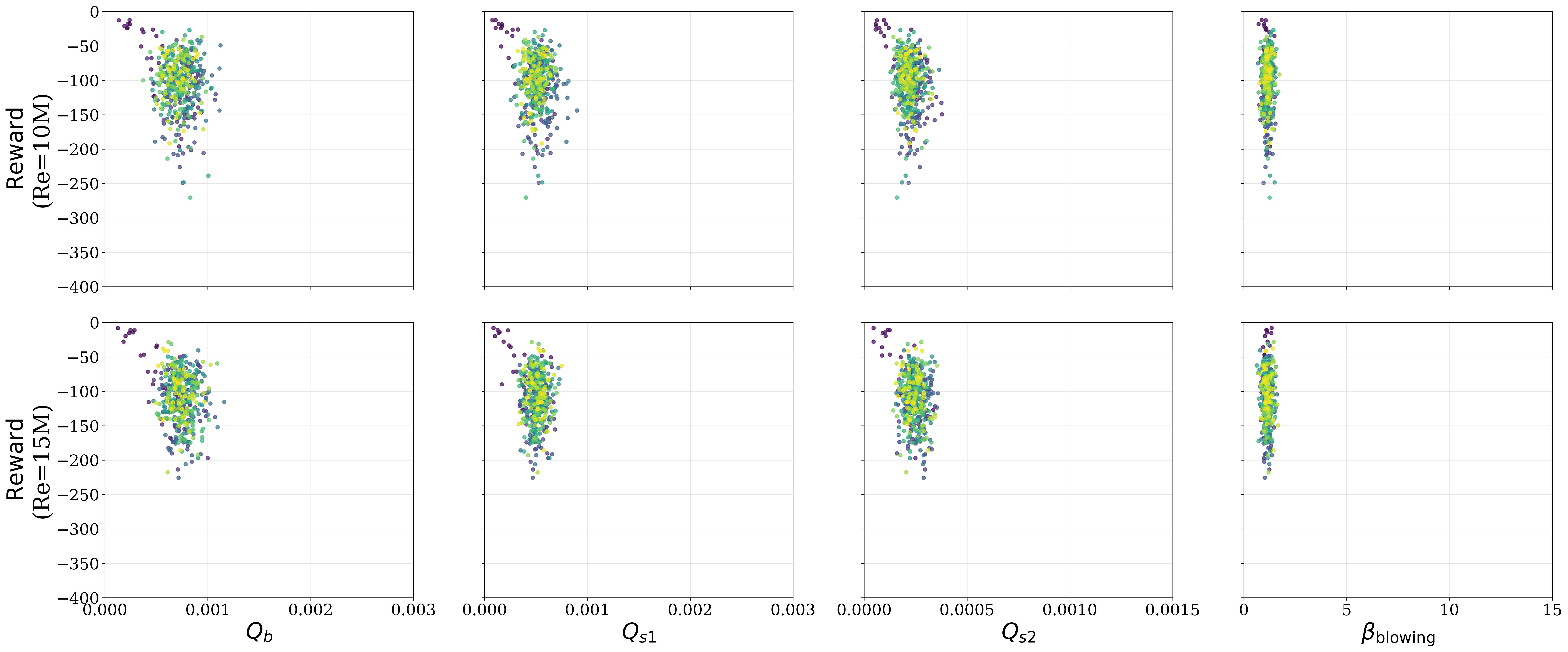}
\includegraphics[scale=0.24, clip=true]{figures/colorbar_reward.png}
\caption{Zero-shot Generalization: Distribution of instantaneous reward against control parameters: blowing mass flux ($Q_b$), suction mass fluxes ($Q_{s1}, Q_{s2}$), and blowing angle ($\beta_{\text{blowing}}$) for $Re=10\times10^{6}$ and $Re=15\times10^{6}$ with 15 optimal sensors under 10\% noise level in TR50. The color gradient indicates the physical time.}
\label{fig:Re_reward_infer_0shot_15sensors}
\end{figure}

Figure \ref{fig:Re_p1p2_infer_0shot_15sensors} shows the time evolution of the normalized wall pressures during the inference. The downstream pressure, $p_{1}/p_{\infty}$, shows strong high-frequency oscillations for both $Re = 10 \times 10^{6}$ (blue) and $Re = 15 \times 10^{6}$ (red). These fluctuations are larger than those observed during training, primarily due to the combined effects of 10\% sensor noise and the increased turbulence at higher Reynolds numbers. The upstream pressure, $p_{2}/p_{\infty}$, is the key indicator of inlet stability. In the uncontrolled baseline case (black), a rapid pressure rise at $p_{2}$ indicates an immediate unstart. In contrast, the controlled cases maintain $p_{2}$ at a low, steady level throughout the simulation. Despite noisy measurements and aggressive flow conditions, the controller successfully anchors the shock train within the isolator, preventing the terminal shock from propagating past the downstream sensing location.
To explain how the agent maintains stability in this new flow regime, Figure \ref{fig:Re_reward_infer_0shot_15sensors} shows the distribution of control actions during inference. The scatter plots indicate that the agent behaves consistently despite the unseen conditions, with actions for both Reynolds numbers forming tight, well-defined clusters. In particular, the policy repeatedly selects blowing angles ($\beta_{\text{blowing}}$) close to near-wall tangential actuation, while adjusting the blowing ($Q_b$) and suction ($Q_{s1}, Q_{s2}$) mass fluxes within narrow ranges. The clustering of high-reward outcomes within these regions indicates that the agent is identifying and exploiting a stable control regime. These results demonstrate that the agent successfully generalizes to previously unseen Reynolds numbers. Using a policy trained only at $Re = 5 \times 10^{6}$, the controller adapts to the thinner boundary layers and modified shock--boundary layer interaction scales at $Re = 10 \times 10^{6}$ and $Re = 15 \times 10^{6}$.

\section{Summary and Discussions}
\noindent
\noindent
In this study, we investigate the control of unstart phenomena in a Mach 5 air-breathing hypersonic intake using a deep reinforcement learning (DRL)-based active flow control (AFC) framework. Unstart events, characterized by strong shock–boundary-layer interactions, flow spillage or separation, and large pressure oscillations, pose severe threats to thrust performance, structural integrity, and operational reliability for hypersonic vehicles. Conventional control strategies, including adjoint-based optimization, PID, or model-based controllers, are computationally intractable or fundamentally incapable of maintaining robustness under the highly nonlinear, transient, and uncertain conditions typical of hypersonic flows. Reduced-order or simplified models fail to capture the full range of flow dynamics, making them unsuitable for developing effective control strategies in this context. By leveraging a high-fidelity, high-order computational fluid dynamics (CFD) environment in combination with DRL, we develop model-free, adaptive control policies capable of exploiting the complete flow information to learn complex, non-intuitive actuation strategies. Consequently, benchmarking against classical or reduced-order approaches is not meaningful, and the proposed high-fidelity CFD + DRL framework represents the most capable and practically deployable controller for this challenging hypersonic flow problem. The key findings and contributions of this work can be summarized as follows:

\begin{enumerate}
    \item \textbf{High-fidelity CFD framework for unstart simulation:} We find that the fourth-order spectral Discontinuous Galerkin (DG) scheme, even when coupled with a strong-stability-preserving Runge-Kutta (SSP-RK) method of order (5,4), fails to adequately resolve highly transient, small-scale flow features. Meaningful convergence is observed only from fifth order onward. Accordingly, all results reported here employ a fifth-order spectral DG spatial discretization combined with an SSP-RK scheme of order (5,4) for temporal integration. The in-house solver incorporates conservative adaptive mesh refinement (AMR), enabling accurate resolution of shocks and fine-scale flow structures that are critical for capturing unstart dynamics. Owing to the presence of strong shocks, thin boundary layers, and extreme flow velocities inherent to hypersonic regimes, the governing equations impose a stringent stability constraint on the time step through the Courant-Friedrichs-Lewy (CFL) condition, necessitating very small time-step sizes for numerical stability and accuracy. While computationally demanding, this restriction is essential to faithfully capture the rapid temporal evolution and nonlinear interactions characteristic of high-speed compressible flows. This approach ensures that the control framework is trained and validated on physically accurate flow representations, preserving the highly nonlinear and unsteady behavior associated with hypersonic intakes.

\item \textbf{Deployment of suction and blowing jets:} Suction and blowing jets are applied at constant net mass flow to learn both the optimal jet strength and blowing angle. Suction jets, positioned near the isolator exit, reduce core mass flow but delay \textit{Kantrowitz unstart} by weakening the shock train and lowering upstream pressure, enhancing inlet operability for large back pressures without increasing the absolute mass flow.

    \item \textbf{DRL-based control across varying throttling ratios:} The DRL-based controller was trained over multiple throttling ratios (TRs) representing different back-pressure conditions. This multi-condition training ensures that the learned policy is robust across a wide spectrum of flow regimes, actively controlling unstart by responding adaptively to varying disturbances and pressure fluctuations.

    \item \textbf{Zero-shot generalization to unseen back pressures:} A controller trained exclusively at TR40 was deployed to previously unseen operating conditions, TR30 and TR50, without additional retraining. The successful control of unstart under these conditions demonstrates the ability of the DRL policy to generalize beyond its training envelope, highlighting its potential for real-time deployment in operational hypersonic systems.

    \item \textbf{Robustness under noisy sensor measurements:} The performance of the DRL controller was evaluated with noisy sensor inputs to mimic practical sensing limitations. The controller maintained effective unstart suppression, indicating resilience to measurement uncertainty and reinforcing its suitability for real-world implementation.

    \item \textbf{Optimal sensor placement:} We also identify optimal sensor locations and evaluate controller performance under zero-shot generalization. By relying on a minimal sensor set, the approach ensures practical implementability without compromising control effectiveness, which is critical for onboard hypersonic systems where sensor redundancy is limited.

     \item \textbf{Zero-shot generalization to unseen Reynolds number:} Zero-shot generalization to previously unseen Reynolds numbers is essential for hypersonic intake control, as real operating environments span a broad and continuously varying range of flight conditions. Variations in flight speed, altitude, and atmospheric density modify the Reynolds number, leading to significant changes in shock-boundary-layer interactions, flow separation, and mass capture. A controller that generalizes across Reynolds numbers can therefore maintain robust performance without repeated retraining as conditions evolve. Here, we demonstrate the zero-shot generalization capability of a controller trained on the TR40 configuration at a Reynolds number of $5 \times 10^6$, and evaluate its performance at two previously unseen Reynolds numbers, $10 \times 10^6$ and $15 \times 10^6$, for the TR50 configuration under high back-pressure conditions with 10\% noise. The results show stable and consistent control performance without any additional training.

    \item 
\textbf{Practical Implementation:}
The transition of this DRL framework from numerical simulation to experimental validation depends critically on achieving the required control frequency. In this study, the control policy operates with an update interval of 20 $\mu$s, corresponding to a frequency of 50 kHz. This high-frequency actuation means that the action values for the microjet mass fluxes are updated every 20 $\mu$s, which is effectively a continuous control relative to the millisecond-scale dynamics of the shock train unstart. Simultaneously, the computational inference for mapping the 15-sensor state vector to the 4-dimensional action vector must be executed within the 20 $\mu$s window, which necessitates the use of Field-Programmable Gate Arrays (FPGAs) \cite{brown1992field} or specialized real-time AI accelerators to ensure deterministic, low-latency processing.

\end{enumerate}
An important question that warrants further clarification is why the DRL-based control strategy generalizes effectively across a range of uncertain operating conditions. The strong generalization of the DRL-based controller across uncertain conditions arises from invariance in the underlying flow–control structure. Although changes in Reynolds number and back pressure modify shock strength, separation extent, and time scales, the dominant sensor–actuator relationships and control authority remain unchanged within the investigated regime. By using normalized wall-pressure (with respect to free stream pressure) measurements, operating-point dependence is removed while preserving the spatiotemporal features associated with shock motion and separation dynamics. The agent therefore responds to invariant flow patterns rather than absolute pressure levels. Robustness is further reinforced by injecting noise during training, which discourages reliance on fragile thresholds and promotes feedback based on persistent, physically meaningful features. The resulting zero-shot generalization reflects robust nonlinear feedback control across a family of hypersonic flow conditions, rather than extrapolation to fundamentally new regimes. We also observe that active flow control strategies for hypersonic intake unstart mitigation based on deep reinforcement learning demand comprehensive modeling to advance their \textit{technology readiness level} toward flight demonstration. High-fidelity numerical simulations are essential to capture the strongly nonlinear interactions among flow physics, actuators, sensors, and the learned control policy, providing a controlled environment in which performance, robustness, and stability can be rigorously assessed. Overall, this work demonstrates that DRL-based control strategies, trained within high-fidelity simulation environments and coupled with optimized sensor configurations, can provide robust, real-time mitigation of unstart phenomena in hypersonic intakes. The combination of adaptive intelligence, zero-shot generalization, and noise-resilient operation presents a promising pathway toward reliable and efficient air-breathing hypersonic propulsion, paving the way for future advances in high-speed flight technologies.

Looking forward, a key avenue for advancing the practicality of learning-enabled hypersonic flow control lies in the integration of scientific foundation models (SciFMs) \cite{menon2026scientific} as physics-aware, modal-based surrogates to replace or augment computationally expensive, model-free high-fidelity CFD environments. By leveraging reduced-order representations grounded in dominant flow structures and trained on multi-fidelity simulation and experimental datasets, SciFMs can significantly accelerate policy training, enable rapid uncertainty quantification, and improve interpretability of the learned control mechanisms. Such models offer a promising pathway to bridge the gap between data-driven control and first-principles physics, facilitating scalable optimization across operating envelopes while preserving critical nonlinear dynamics associated with intake unstart. The synergistic coupling of SciFMs with deep reinforcement learning may therefore enable more efficient closed-loop design, support real-time digital twin frameworks, and ultimately accelerate the transition of adaptive hypersonic flow control technologies from simulation to flight-ready systems.

\appendix

\section{Weak formulation of two-dimensional Navier-Stokes equations} \label{appA}
Let the computational domain be defined as $\Omega = \bigcup_{e=1}^{N_e} \Omega_e$, and let $p \ge 0$ represent the polynomial degree of the approximation. We define the discrete \textit{trial} and \textit{test} function spaces using the standard discontinuous piecewise polynomial sets:
\[
\begin{aligned}
\mathcal{V}^h &:= \left\{\, \mathbf{v} \in [L^2(\Omega)]^m \;:\; 
\mathbf{v}|_{\Omega_e} \in [\mathbb{P}^p(\Omega_e)]^m \text{ for each element } \Omega_e \,\right\},\\[6pt]
\mathcal{U}^h &:= \left\{\, \mathbf{w} \in [L^2(\Omega)]^m \;:\;
\mathbf{w}|_{\Omega_e} \in [\mathbb{P}^p(\Omega_e)]^m \text{ for each element } \Omega_e \,\right\}.
\end{aligned}
\]
For this formulation, we set $\mathcal{U}^h = \mathcal{V}^h$. Here, $m$ represents the number of conserved variables, and $\mathbb{P}^p(\Omega_e)$ denotes the space of polynomials with a total degree of at most $p$ on the element $\Omega_e$. The approximate solution $\mathbf{U}^h$ is expanded using trial basis functions as
\[
 \mathbf{U}(\mathbf{x},t) \approx \mathbf{U}^h(\mathbf{x},t) = \sum_{i=1}^{N_p} \mathbf{U}_i^h(t)\, \Phi_i^h(\mathbf{x}),
\]
and we employ a corresponding \textit{group flux formulation} given by
\[
 \mathbf{F}(\mathbf{x},t) \approx  \mathbf{F}^h(\mathbf{x},t) = \sum_{i=1}^{N_p} \mathbf{F}_i^h(t)\, \Phi_i^h(\mathbf{x}); 
\quad 
 \mathbf{G}(\mathbf{x},t) \approx \mathbf{G}^h(\mathbf{x},t) = \sum_{i=1}^{N_p} \mathbf{G}_i^h(t)\, \Phi_i^h(\mathbf{x}),
\]
where $\Phi_i^h(\mathbf{x})$ represent the local basis functions that span $\mathcal{U}^h|_{\Omega_e}$, while $\mathbf{U}_i^h(t)$, $\mathbf{F}_i^h(t)$, and $\mathbf{G}_i^h(t)$ denote the time-dependent coefficients (degrees of freedom) for the solution and fluxes, respectively.

To derive the weak formulation, we multiply the conservation law (equation \eqref{NSE}) by a test function $\mathbf{v}^h \in \mathcal{V}^h$ and integrate over the element $\Omega_e$:
\begin{equation}
\int_{\Omega_e} \mathbf{v}^h \frac{\partial \mathbf{U}^h}{\partial t} \, d\Omega 
+ \int_{\Omega_e} \mathbf{v}^h \left(
\frac{\partial \mathbf{F}_{\text{inv}}^h}{\partial x} 
+ \frac{\partial \mathbf{G}_{\text{inv}}^h}{\partial y}
\right) d\Omega
= \int_{\Omega_e} \mathbf{v}^h \left(
\frac{\partial \mathbf{F}_v^h}{\partial x} 
+ \frac{\partial \mathbf{G}_v^h}{\partial y}
\right) d\Omega.
\end{equation}

Applying integration by parts to the spatial flux terms leads to
\begin{multline}
\int_{\Omega_e} \mathbf{v}^h \frac{\partial \mathbf{U}^h}{\partial t} \, d\Omega 
- \int_{\Omega_e} \left(
\frac{\partial \mathbf{v}^h}{\partial x}\mathbf{F}_{\text{inv}}^h 
+ \frac{\partial \mathbf{v}^h}{\partial y}\mathbf{G}_{\text{inv}}^h
\right) d\Omega
+ \int_{\partial\Omega_e} \mathbf{v}^h 
\left(
\mathbf{F}_{\text{inv}}^h n_x 
+ \mathbf{G}_{\text{inv}}^h n_y
\right) dS \\
= -\int_{\Omega_e} \left(
\frac{\partial \mathbf{v}^h}{\partial x}\mathbf{F}_v^h 
+ \frac{\partial \mathbf{v}^h}{\partial y}\mathbf{G}_v^h
\right) d\Omega
+ \int_{\partial\Omega_e} \mathbf{v}^h 
\left(
\mathbf{F}_v^h n_x 
+ \mathbf{G}_v^h n_y
\right) dS.
\end{multline}

In this context, $\partial\Omega_e$ represents the boundary of the element, and $\hat{\mathbf{n}} = (n_x, n_y)^T$ denotes the outward-pointing unit normal vector. Because the solution is allowed to be discontinuous across element interfaces, the physical boundary fluxes are replaced by uniquely defined numerical fluxes:
\begin{multline}
\int_{\Omega_e} \mathbf{v}^h \frac{\partial \mathbf{U}^h}{\partial t} \, d\Omega 
= \int_{\Omega_e} \left(
\frac{\partial \mathbf{v}^h}{\partial x}\mathbf{F}_{\text{inv}}^h 
+ \frac{\partial \mathbf{v}^h}{\partial y}\mathbf{G}_{\text{inv}}^h
\right) d\Omega
- \int_{\partial\Omega_e} \mathbf{v}^h 
\left(
\mathbf{F}_{\text{inv}}^{h,*} n_x 
+ \mathbf{G}_{\text{inv}}^{h,*} n_y
\right) dS \\[2mm]
- \int_{\Omega_e} \left(
\frac{\partial \mathbf{v}^h}{\partial x}\mathbf{F}_v^h 
+ \frac{\partial \mathbf{v}^h}{\partial y}\mathbf{G}_v^h
\right) d\Omega
+ \int_{\partial\Omega_e} \mathbf{v}^h 
\left(
\mathbf{F}_v^{h,*} n_x 
+ \mathbf{G}_v^{h,*} n_y
\right) dS.
\end{multline}

The numerical fluxes $\mathbf{F}_{\text{inv}}^{h,*}$, $\mathbf{G}_{\text{inv}}^{h,*}$, $\mathbf{F}_v^{h,*}$, and $\mathbf{G}_v^{h,*}$ are computed based on the interior state $\mathbf{U}^{h-}$ and exterior state $\mathbf{U}^{h+}$, thereby coupling neighboring elements and resolving discontinuities at the interfaces.

Finally, the complete \textit{global system} is assembled by summing the contributions from all elements:
\begin{align}
\sum_{e=1}^{N_e} 
\Bigg[ 
& \int_{\Omega_e} \mathbf{v}^h \frac{\partial \mathbf{U}^h}{\partial t} \, d\Omega
- \int_{\Omega_e} \Big(
\frac{\partial \mathbf{v}^h}{\partial x} \mathbf{F}_{\text{inv}}^h 
+ \frac{\partial \mathbf{v}^h}{\partial y} \mathbf{G}_{\text{inv}}^h
\Big) d\Omega
+ \int_{\partial\Omega_e} \mathbf{v}^h 
\Big(
\mathbf{F}_{\text{inv}}^{h,*} n_x 
+ \mathbf{G}_{\text{inv}}^{h,*} n_y
\Big) dS \notag\\
& + \int_{\Omega_e} \Big(
\frac{\partial \mathbf{v}^h}{\partial x} \mathbf{F}_v^h 
+ \frac{\partial \mathbf{v}^h}{\partial y} \mathbf{G}_v^h
\Big) d\Omega
- \int_{\partial\Omega_e} \mathbf{v}^h 
\Big(
\mathbf{F}_v^{h,*} n_x 
+ \mathbf{G}_v^{h,*} n_y
\Big) dS
\Bigg] = 0.
\end{align}

The DG weak formulation relies on numerical fluxes to resolve discontinuities across element boundaries and to compute boundary integrals.

\vspace{0.2cm}
\noindent\textbf{Inviscid Flux:} To address solution jumps at element interfaces, the Lax-Friedrichs (Rusanov) flux \cite{edwards2006dominant} is employed
\[
\mathbf{F}_{\text{inv}}^{h,*} = \frac{1}{2} \left[ \mathbf{F}^h(\mathbf{U}^{h-}) + \mathbf{F}^h(\mathbf{U}^{h+}) \right] - \frac{\lambda}{2} (\mathbf{U}^{h+} - \mathbf{U}^{h-}),
\]
where $\mathbf{U}^{h-}$ and $\mathbf{U}^{h+}$ represent the interior and exterior states, respectively, and $\lambda$ denotes the maximum wave speed at the interface. The first term provides an average of the fluxes, while the second introduces dissipation proportional to the solution jump, thereby stabilizing the scheme near discontinuities. The flux $\mathbf{G}_{\text{inv}}^{h,*}$ is calculated analogously.

\vspace{0.2cm}
\noindent\textbf{Viscous Flux:} The numerical viscous fluxes at element interfaces, $\mathbf{F}_v^{h,*}$ and $\mathbf{G}_v^{h,*}$, are evaluated using the BR1 scheme \cite{bassi1997high}. Within the DG framework, this requires the introduction of an auxiliary gradient variable \(\mathbf{S}^h = \nabla \mathbf{U}^h\). This allows the interfacial viscous flux to be expressed as a function of both the solution and its gradient:
\[
\mathbf{F}_v^{h,*} = \frac{1}{2} \left[ \mathbf{F}_v(\mathbf{U}^{h-}, \mathbf{S}^{h-}) + \mathbf{F}_v(\mathbf{U}^{h+}, \mathbf{S}^{h+}) \right].
\]
Similarly, $\mathbf{G}_v^{h,*}$ is computed.

The auxiliary gradient \(\mathbf{S}^h\) is determined locally for each element via the DG gradient operator:
\[
\int_{\Omega_e} \phi_j \, \mathbf{S}^h \, d\Omega 
= - \int_{\Omega_e} \nabla \phi_j \, \mathbf{U}^h \, d\Omega 
+ \int_{\partial \Omega_e} \phi_j \, \hat{\mathbf{U}} \, \hat{\mathbf{n}} \, dS,
\]
where \(\phi_j\) are the test functions and \(\hat{\mathbf{U}}\) is the numerical trace of \(\mathbf{U}^h\) on the element boundary, typically defined as the interior value or an average with neighboring elements. This formulation ensures the computed gradient remains consistent with the DG framework, allowing viscous fluxes to incorporate solution gradients while preserving local conservation and stability across non-conforming interfaces.

The Spectral DG formulation is realized by utilizing high-order polynomial basis functions, with interpolation and flux evaluations collocated at Gauss–Lobatto–Legendre (GLL) nodes within each element. By adopting Lagrange polynomials defined at these nodes and applying Gauss–Lobatto quadrature for the weak form integrals (Hesthaven \& Warburton \cite{hesthaven2008nodal}), the method guarantees exact integration at the collocation points. This strategy maintains the local conservation and geometric flexibility of standard DG while achieving spectral (exponential) convergence for smooth solutions, effectively synthesizing the advantages of both spectral and discontinuous Galerkin methods.

\section{SSP-RK Time Discretization} \label{appB}

The semi-discrete DG formulation results a system of ordinary differential equations (ODEs) evolving in time:
\begin{equation}
\frac{d\mathbf{U}^h}{dt} = \mathcal{L}(\mathbf{U}^h, t)
\end{equation}
where $\mathbf{U}^h$ denotes the vector of discrete solution values, and $\mathcal{L}$ encapsulates the spatial discretization operator, incorporating both volume and surface flux contributions. To advance this system, we apply the general form of an $s$-stage SSPRK method (\cite{spiteri2002new}) to the semi-discrete DG equations:
\begin{align}
\mathbf{U}^{h,(0)} &= \mathbf{U}^{h,n}, \\
\mathbf{U}^{h,(i)} &= \sum_{k=0}^{i-1} \left[\alpha_{ik} \mathbf{U}^{h,(k)} + \Delta t \, \beta_{ik} \, \mathcal{L}(\mathbf{U}^{h,(k)})\right], \quad i = 1, 2, \ldots, s, \\
\mathbf{U}^{h,n+1} &= \mathbf{U}^{h,(s)},
\end{align}
where $\mathbf{U}^{h,(i)}$ represents the discrete solution at the intermediate stage $i$. Consistency and stability are ensured by requiring the coefficients to satisfy $\alpha_{ik} \geq 0$, $\beta_{ik} \geq 0$, and $\sum_{k=0}^{i-1} \alpha_{ik} = 1$.

For the specific SSPRK54 scheme, the Courant-Friedrichs-Lewy (CFL) coefficient is given by $\tilde{c} = \min_{i,k} \frac{\alpha_{ik}}{\beta_{ik}} \approx 1.508$. This coefficient dictates the maximum allowable time step for the convective contributions:
\begin{equation}
\Delta t_C \leq \tilde{c} \, \frac{\Delta x_{min}}{(2p+1)\lambda_{\max}},
\end{equation}
where $\Delta x_{min}$ represents the characteristic element size and $\lambda_{\max}$ denotes the maximum wave speed within the domain (as determined in the surface flux computations). 
Similarly, the stability constraint associated with the diffusion or viscous terms is defined as
\begin{equation}
\Delta t_D \leq \tilde{d} \, \frac{\Delta x^2_{min}}{(2p+1)^2\nu_{\max}},
\end{equation}
where $\tilde{d} \approx 0.5$, and $\nu_{\max}$ corresponds to the larger of the kinematic viscosity and the thermal diffusivity.

Consequently, the global time step $\Delta t$ is selected as the minimum of the convective (hyperbolic) and diffusive (parabolic) constraints: 
$\Delta t = \text{min} (\Delta t_C,\Delta t_D) $.
The SSPRK54 method ensures fourth-order temporal accuracy while maintaining strong stability properties, which is important for accurately resolving shocks encountered in compressible transonic flows.

\section{Velocity Profile and Mass Balance for Microjets} \label{appC}
\begin{figure}[htpb]
    \centering 
    \includegraphics[scale=0.8]{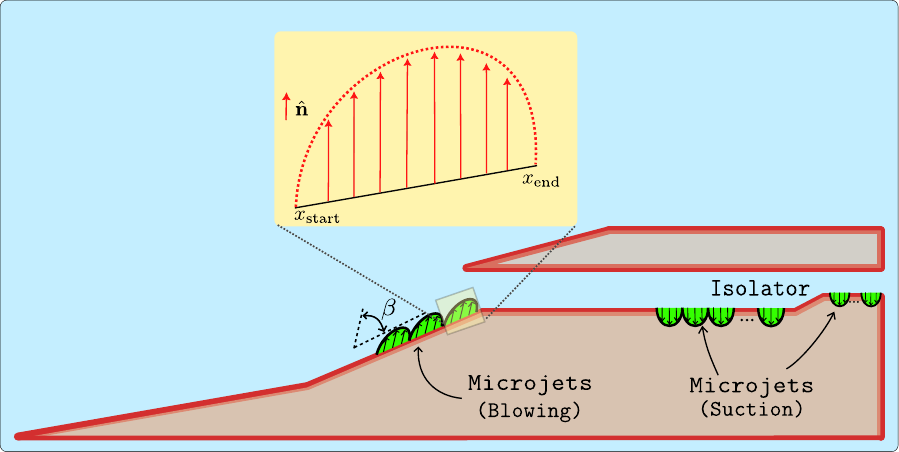}
    \caption{Schematic of the hypersonic inlet control surface showing blowing and suction microjet arrays. The inset details the flattened sinusoidal velocity profile used to ensure continuity at the wall boundaries.}
\label{fig:jet_profile}
\end{figure}

\noindent \textbf{Jet Velocity Profile:} The jet velocity distribution follows a flattened sinusoidal profile across each jet width to ensure smooth flow transitions at the jet boundaries while maximizing the actuation area (see Figure \ref{fig:jet_profile}). Let \( x \) denote the coordinates along the inlet surface. 
For a jet spanning from \( x_{\text{start}} \) to \( x_{\text{end}} \), the local jet velocity magnitude is defined as:
\begin{equation}
    \|\mathbf{u}_{\text{jet}}(x)\| = \lambda \cdot u_{\infty} \cdot \sin\left(\pi \frac{x - x_{\text{start}}}{x_{\text{end}} - x_{\text{start}}}\right)^{\phi},
\end{equation}
where \( \lambda \) is a continuous velocity factor determined by the RL policy, \( u_{\infty} \) is the freestream velocity magnitude, and the exponent \( \phi \approx 0.1 \) creates a flattened distribution that decays smoothly to zero at the edges.

To account for the inlet geometry, the local velocity vector $\mathbf{u}_{\text{local}}(x)$ is defined in the global frame by the product of the magnitude and a unit direction vector $\hat{\mathbf{n}}$:
\begin{equation}
    \mathbf{u}_{\text{local}}(x) = \|\mathbf{u}_{\text{jet}}(x)\| \cdot \hat{\mathbf{n}},
\end{equation}
where $\hat{\mathbf{n}}$ incorporates the local surface slope $\alpha$ and is given by:
\begin{equation}
    \hat{\mathbf{n}} = 
    \begin{bmatrix}
        \cos(\theta_{\text{jet}}) \\
        \sin(\theta_{\text{jet}})
    \end{bmatrix},
\end{equation}
with $\theta_{\text{jet}}$ representing the effective jet angle. For the \textit{blowing jets} located on the second compression ramp ($\alpha = 21^\circ$), the angle is learnable and directed outward (positive sign relative to the surface normal):
\[
    \theta_{\text{jet, blow}} = \alpha + \beta,
\]
where $\beta$ is a continuous action output by the RL agent representing the blowing angle relative to the surface tangent.
For the \textit{suction jets} located on the isolator floor and step ($\alpha = 0^\circ$), the flow is directed inward (negative sign), perpendicular to the wall.
\[
    \theta_{\text{jet, suction}} = -90^\circ.
\]
This formulation ensures that blowing jets inject momentum with a learnable angle relative to the ramp and the suction jets remove mass normal to the flow direction. The resulting $\mathbf{u}_{\text{local}}$ is used to modify the boundary conditions to incorporate jet effects in the flow.

\vspace{0.2cm}
\noindent \textbf{Mass Balance:} Synthetic microjets and other AFC devices are often implemented in incompressible solvers by prescribing a time-varying velocity at the jet orifice. However, extending these methods to compressible flows introduces a subtle but critical challenge: prescribing the same velocity does not guarantee the same mass flux. For compressible flows, the mass flux depends on both the local density and the velocity:
\begin{equation}
\dot{m} = \int_{A_{jet}} \rho\, (\mathbf{u} \cdot \mathbf{n})\, dA_{jet}.
\end{equation}

To fully define the boundary state of a synthetic jet in a compressible flow solver, three primitive variables must be specified: density, pressure, and the velocity vector. While the DRL agent dynamically learns the optimal velocity vector to manipulate the flow field, the thermodynamic variables must be defined consistently to satisfy the physics of the system, particularly mass balance.

 Let $\dot{\hat{m}}$ denote the discrete approximation of $\dot{m}$, in which the integral is replaced by a finite summation.
\[
\dot{\hat{m}}_{\text{blow}} = \dot{\hat{m}}_{\text{suction}} \Rightarrow  \rho_b A_b v_b = \rho_{s1} A_{s1} v_{s1} + \rho_{s2} A_{s2} v_{s2}.
\]
For brevity, we removed summation sign. The subscripts $b$ and $s$ denote the blowing and suction jets, respectively.

\vspace{0.2cm}
\noindent \textbf{Area Constraints:} The number of blowing jets on the $\Gamma_1$ wall is 3, while the numbers of suction jets on the $\Gamma_2$ and $\Gamma_3$ walls inside the isolator are 6 and 3, respectively.
\[
A_{s1} = 2 A_b, \quad A_{s2} = A_b.
\]
The values of $\rho_{s1}$ and $\rho_{s2}$ are provided by the CFD solver, while the velocities of all microjets are determined by the DRL agent. 

Momentum transfer dominates jet authority: only jets injecting sufficient momentum relative to the primary flow can influence separation, instabilities, or flow direction, whereas low-momentum jets are quickly entrained and convected downstream. Energy considerations complete the picture, as the injected kinetic energy is redistributed, converted to turbulence, and dissipated, dictating the energetic cost and efficiency of control. These processes occur regardless of whether the energy balance is explicitly tracked; neglecting it does not eliminate energy transfer, but merely obscures how actuator power is partitioned and dissipated.

\section{TD3 Algorithm} \label{appD}
\begin{algorithm}
\caption{TD3 Algorithm}
\begin{algorithmic}[1]
\STATE Initialize the actor network $\pi_\theta$, the critic networks $Q_{\phi_1}, Q_{\phi_2}$, and their corresponding target networks
\STATE Initialize the experience replay buffer $\mathcal{D}$
\WHILE{the training not stopped}
    \STATE Observe the current state $s_t$
    \STATE Select an action with exploration noise: $a_t = \pi_\theta(s_t) + \epsilon$, \text{where } $\epsilon \sim \text{Normal}(0, \sigma)$
    \STATE Execute $a_t$ within the CFD simulation to observe the next state $s_{t+1}$ and reward $r_t$
    \STATE Store the transition tuple $(s_t, a_t, r_t, s_{t+1})$ in the buffer $\mathcal{D}$
    \STATE Sample a random minibatch of transitions from $\mathcal{D}$
    \STATE Compute the smoothed target actions:
        \STATE \quad $\tilde{a}_{t+1} = \pi_{\theta'}(s_{t+1}) + \text{clip}(\text{Normal}(0, \sigma_{\text{target}}), -c, c)$
    \STATE Calculate the target Q-value:
        \STATE \quad $y_t = r_t + \gamma \min(Q_{\phi_1'}(s_{t+1}, \tilde{a}_{t+1}), Q_{\phi_2'}(s_{t+1}, \tilde{a}_{t+1}))$
    \STATE Update the critic networks $Q_{\phi_i}$ by minimizing the Mean Squared Error (MSE) loss
    \IF{current step \% policy\_update\_interval == 0}
        \STATE Update the actor (policy network) via the deterministic policy gradient
        \STATE Perform a soft update on the target networks:
        \STATE \quad For each critic: $\phi_i' \gets \tau \phi_i + (1-\tau) \phi_i'$
        \STATE \quad For the actor: $\theta' \gets \tau \theta + (1-\tau) \theta'$
    \ENDIF
\ENDWHILE
\end{algorithmic}
\end{algorithm}
The TD3 algorithm~\cite{fujimoto2018addressing} is an off-policy actor–critic method tailored for continuous control tasks, extending the foundational DDPG framework~\cite{lillicrap2015continuous} via three critical modifications designed to mitigate overestimation bias and bolster training stability. To address overoptimistic value updates, the method uses clipped double Q-learning, with two concurrent critic networks and deriving the target value from the minimum of their respective Q-estimates. Furthermore, the algorithm implements delayed policy updates, updating the actor network at a lower frequency than the critics to ensure more stable policy learning. Finally, to reduce sensitivity to value spikes and encourage exploration, target policy smoothing is introduced by injecting small, clipped noise into the target actions during critic updates. TD3 significantly enhances sample efficiency and stability within continuous control domains by explicitly mitigating the function approximation error and overestimation bias characteristic of DDPG. However, as an off-policy method, it remains susceptible to bias resulting from distributional shifts between the replay buffer samples and the current policy, which often delay convergence. Consequently, performing multiple independent training runs is recommended to ensure the consistency and reliability of the learned policy performance.

\section{SAC Algorithm} \label{appE}
\begin{algorithm}[H]
\caption{SAC Algorithm}
\begin{algorithmic}[1]
\STATE Initialize the stochastic actor network $\pi_\theta$, the critic networks $Q_{\phi_1}, Q_{\phi_2}$, and their corresponding target networks
\STATE Initialize the entropy temperature $\alpha$ and the target entropy $\bar{\mathcal{H}} = -\dim(\mathcal{A})$ (where $\mathcal{A}$ is the action)
\STATE Initialize the experience replay buffer $\mathcal{D}$
\WHILE{the training not stopped}
    \STATE Observe the current state $s_t$
    \STATE Sample an action from the policy: $a_t \sim \pi_\theta(\cdot | s_t)$
    \STATE Execute $a_t$ within the CFD simulation to observe the next state $s_{t+1}$ and reward $r_t$
    \STATE Store the transition tuple $(s_t, a_t, r_t, s_{t+1})$ in the buffer $\mathcal{D}$
    \STATE Sample a random minibatch of transitions from $\mathcal{D}$
    \STATE Compute the target value components:
        \STATE \quad Sample next actions $\tilde{a}_{t+1} \sim \pi_\theta(\cdot | s_{t+1})$
        \STATE \quad Compute log probability $\log \pi_\theta(\tilde{a}_{t+1} | s_{t+1})$
        \STATE \quad $y_t = r_t + \gamma \left( \min_{j=1,2} Q_{\phi_j'}(s_{t+1}, \tilde{a}_{t+1}) - \alpha \log \pi_\theta(\tilde{a}_{t+1} | s_{t+1}) \right)$
    \STATE Update the critic networks $Q_{\phi_i}$ by minimizing the Mean Squared Error (MSE) loss
    \STATE Update the actor (policy network) by maximizing the expected return and entropy:
        \STATE \quad Minimize $\mathcal{L}_\pi = \mathbb{E}_{s_t \sim \mathcal{D}, \epsilon_t \sim \mathcal{N}} \left[ \alpha \log \pi_\theta(f_\theta(\epsilon_t; s_t)|s_t) - \min_{j=1,2} Q_{\phi_j}(s_t, f_\theta(\epsilon_t; s_t)) \right]$
    \STATE Update the temperature parameter $\alpha$ to match the target entropy:
        \STATE \quad Minimize $\mathcal{L}_\alpha = -\mathbb{E}_{a_t \sim \pi_t} \left[ \alpha (\log \pi_\theta(a_t|s_t) + \bar{\mathcal{H}}) \right]$
    \STATE Perform a soft update on the target networks:
        \STATE \quad For each critic: $\phi_i' \gets \tau \phi_i + (1-\tau) \phi_i'$
\ENDWHILE
\end{algorithmic}
\end{algorithm}
Soft Actor-Critic (SAC) ~\cite{haarnoja2018soft} is an off-policy actor-critic algorithm based on the maximum entropy reinforcement learning framework. Unlike TD3, which learns a deterministic policy, SAC learns a stochastic policy $\pi_\theta$ that maximizes a trade-off between expected return and entropy, a measure of randomness in the policy. This entropy regularization encourages exploration and prevents the policy from collapsing to a suboptimal deterministic behavior too early. The algorithm employs twin Q-networks ($Q_{\phi_1}, Q_{\phi_2}$) to mitigate overestimation bias and uses the reparameterization trick to allow gradients to flow through the stochastic sampling process. Additionally, the temperature parameter $\alpha$, which balances the reward and entropy terms, is automatically tuned during training to maintain a constant target entropy.

\section{Mach Number and Q-Criterion Fields in Various Cases} \label{appF}
\begin{figure}
\centering
\includegraphics[scale=0.9, clip=true]{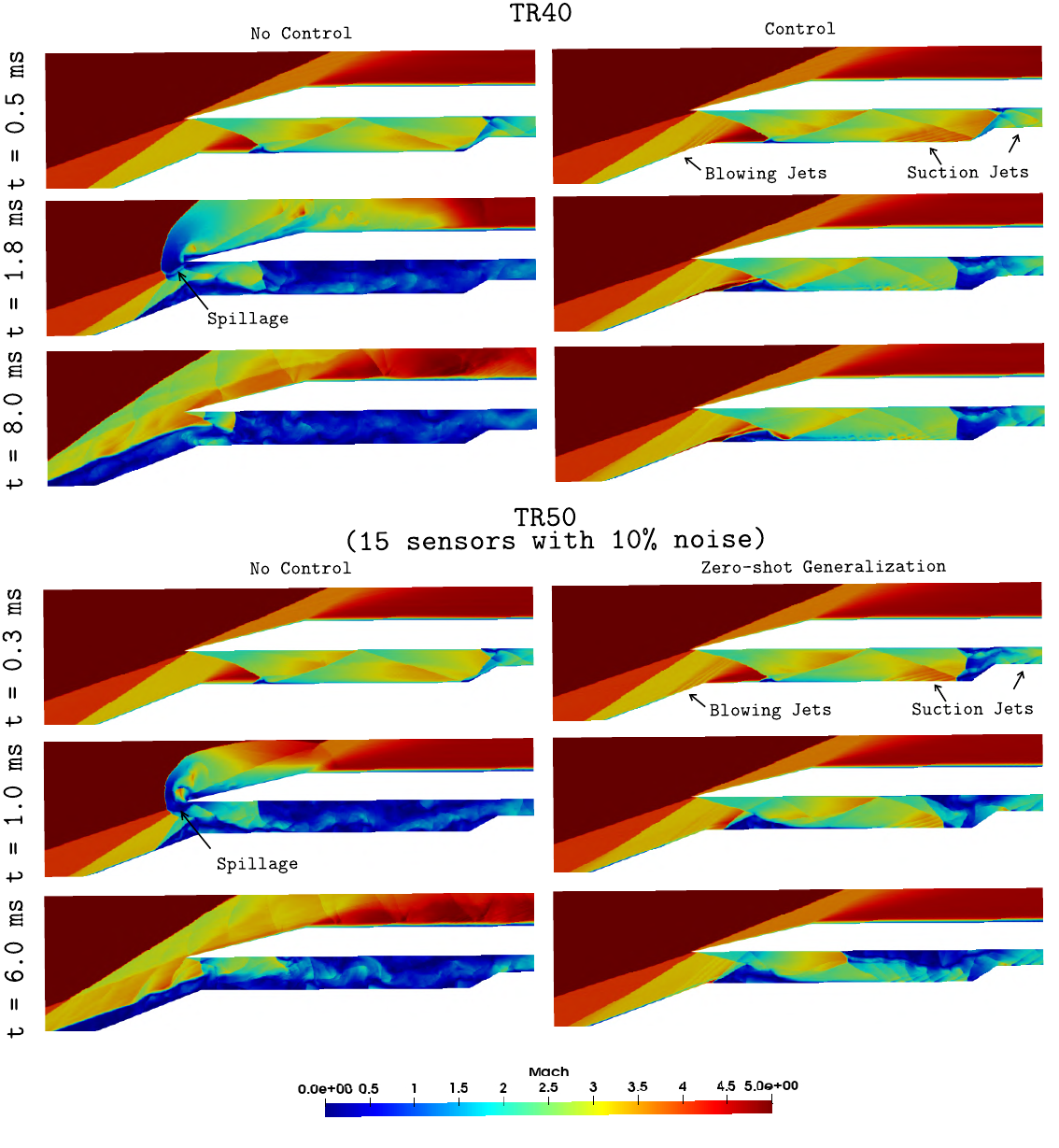}
\caption{TR40 (top): Evolution of Mach number contours with and without control. TR50 (bottom): Uncontrolled case and zero-shot generalization using the controller trained on the TR40 configuration. The evolution of Mach number contours is shown using only 15 optimally placed sensors with 10\% measurement noise.}
\label{fig:appn_mach}
\end{figure}
\begin{figure}
\centering
\includegraphics[scale=0.9, clip=true]{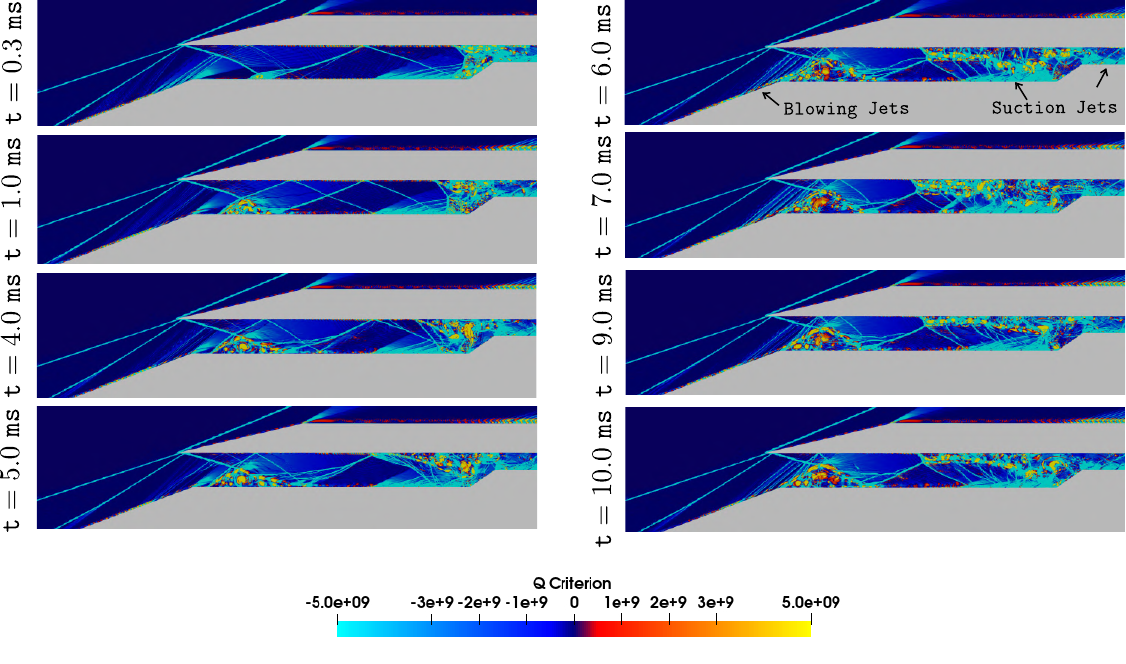}
\caption{Temporal evolution of Q-criterion for the TR50 case with active control and zero-shot generalization (15 sensors, 10\% noise). Vortices generated by the blowing jets are clearly visible after 6~ms.}
\label{fig:appn_qcrit}
\end{figure}
Figure~\ref{fig:appn_mach} presents instantaneous Mach-number contours for uncontrolled and controlled configurations. For the TR40 case (top), the uncontrolled inlet (left column) exhibits shock displacement and flow spillage at approximately 1.8\,ms, whereas the controlled case (right column) remains started. The blowing and suction microjets are clearly visible in the Mach field, where the DRL–based controller stabilizes the shock train within the isolator, preserves a supersonic core, and suppresses spillage. The TR50 case (bottom) compares the uncontrolled configuration (left column) with zero-shot generalization using the TR40-trained controller (right column). In the absence of control, spillage occurs at approximately 1.0\,ms, significantly earlier than in the TR40 case due to the higher imposed back pressure. Despite operating under unseen conditions with 10\% sensor noise and reduced sensing (15 optimal sensors), the controller successfully maintains a started inlet, demonstrating robust zero-shot generalization and effective shock stabilization.

The Q-criterion elucidates coherent vortical structures associated with microjet actuation and their interaction with shock–boundary-layer dynamics. Figure~\ref{fig:appn_qcrit} shows the temporal evolution of the Q-criterion for the TR50 case under active control using zero-shot generalization (15 sensors, 10\% noise). The microjets generate streamwise vortices that energize the boundary layer, suppress separation, and stabilize the shock train, thereby maintaining a started inlet despite the elevated back pressure and measurement noise.

\bibliographystyle{elsarticle-num}
\bibliography{reference}
\end{document}